\documentclass{ieeeaccess}
\usepackage{cite}
\usepackage{amsmath,amssymb,amsfonts}
\usepackage{algorithm,algorithmic}
\usepackage{graphicx}
\usepackage{textcomp}
\usepackage{amsmath}
\usepackage{nccmath}
\usepackage{comment}
\usepackage{soul}
\usepackage{multirow}%
\usepackage[caption=false,font=scriptsize,labelfont=sf,textfont=sf]{subfig}
\usepackage{stfloats}

\def\BibTeX{{\rm B\kern-.05em{\sc i\kern-.025em b}\kern-.08em
    T\kern-.1667em\lower.7ex\hbox{E}\kern-.125emX}}

\usepackage{hyperref}
\usepackage{academicons}
\newcommand{\orcidicon}[1]{\href{https://orcid.org/#1}{\includegraphics[scale=0.01]{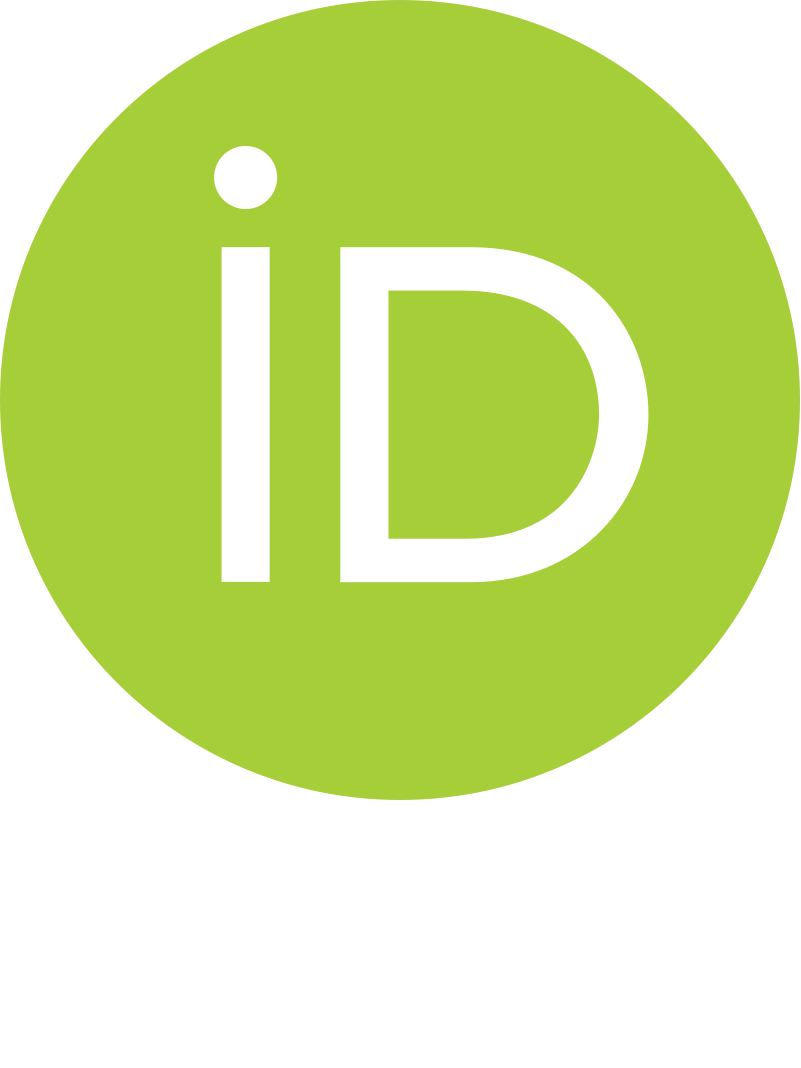}}}

\begin{document}

\history{Received 18 February 2024, accepted 2 May 2024, date of publication 8 May 2024, date of current version 15 May 2024.}
\doi{10.1109/ACCESS.2024.3397732}

\title{ViKi-HyCo: A Hybrid-Control approach for complex car-like maneuvers}
\author{
\uppercase{Edison P. Velasco-Sánchez\,\orcidicon{0000-0003-2837-2001}},
\uppercase{Miguel Ángel Muñoz-Bañón\,\orcidicon{0000-0002-3220-2286}},
\uppercase{Francisco A. Candelas\,\orcidicon{0000-0002-7126-0374}},
\uppercase{Santiago T. Puente\,\orcidicon{0000-0002-6175-600X}}, AND
\uppercase{Fernando Torres\,\orcidicon{0000-0002-6261-9939}},
\IEEEmembership{Senior Member, IEEE}}
\address{Automatics, Robotics, and Computer Vision Group (AUROVA), University of Alicante, 03690 Alicante, Spain.}
\tfootnote{This work was supported in part by the Spanish Government through the research project PID2021-122685OB-I00, the grants for Training of Research Staff PRE2019-088069, from the Government of Spain, and ACIF/2019/088 from the Valencian Community Government and the European Regional Development Fund.}

\markboth
{Edison P. Velasco-Sánchez \headeretal: ViKi-HyCo: A Hybrid-Control approach for complex car-like maneuvers}
{Edison P. Velasco-Sánchez \headeretal: ViKi-HyCo: A Hybrid-Control approach for complex car-like maneuvers}

\corresp{Corresponding author: Edison P. Velasco-Sánchez (e-mail: edison.velasco@ua.es).}

\begin{abstract}
While Visual Servoing is deeply studied to perform simple maneuvers, the literature does not commonly address complex cases where the target is far out of the camera's field of view (FOV) during the maneuver. For this reason, in this paper, we present ViKi-HyCo (Visual Servoing and Kinematic Hybrid-Controller). This approach generates the necessary maneuvers for the complex positioning of a non-holonomic mobile robot in outdoor environments. In this method, we use \hbox{LiDAR-camera} fusion to estimate objects bounding boxes using image and metrics modalities. With the multi-modality nature of our representation, we can automatically obtain a target for a visual servoing controller. At the same time, we also have a metric target, which allows us to hybridize with a kinematic controller. Given this hybridization, we can perform complex maneuvers even when the target is far away from the camera's FOV. The proposed approach does not require an object-tracking algorithm and can be applied to any robotic positioning task where its kinematic model is known. ViKi-HyCo has an error of $\mathbf{0.0428\pm0.0467}$ m in the X-axis and $\mathbf{0.0515\pm0.0323}$ m in the Y-axis at the end of a complete positioning task.
\end{abstract}


\titlepgskip=-15pt

\maketitle
\section{Introduction}
\label{sec1:introduction}
\PARstart{M}{obile} robotics is increasingly challenging precise positioning in unstructured environments, which requires the performance of maneuvers for the robot's motion to achieve an accurate position concerning a specific target in the environment. This task is usually addressed by a visual servoing controller, which is one of the most studied methods in mobile robotics \cite{huang2019visual,nazari2022visual}, where the control actions are obtained by calculating the errors between a target detected in the image plane and its desired position. 

To perform this task, the complexity of algorithms and controllers usually depends on the sensors used for precise positioning, integrating different types of them to recognize their environment and plan the movement toward a target point. Monocular cameras, LiDAR, and RGB-D are the most commonly used environmental perception sensors in mobile robotics \cite{xiao2022motion}. Despite the low cost and easy integration of monocular cameras, the most extended systems using only these sensors present several challenges when trying to recover the metric scale of the environment. A strategy to mitigate this is usually to unify several sensors~\cite{qin2018vins}. To improve visual servoing systems, the authors in \cite{li2021survey} conclude that integration with multiple modules can increase the accuracy and functions of visual systems in robots. Consequently, the integration of sensor fusion and controller management can perform tasks with improved performance.

Visual servoing controllers are based primarily on predefined visual features extracted from an object or scene. Therefore, this type of controller can be abruptly interrupted if it loses these features when the identified object or scene is hidden or leaves the camera's field of view (FOV). Strategies have been developed to mitigate these drawbacks, including the use of homography between the current frame and the keyframe \cite{jia2015switched} or the use of a virtual target-guided fast scanning random tree (RRT) \cite{wang2021virtual}. However, performing a complex maneuver or exchanging the image source for the visual servoing controller is not possible.
 
In addition, another drawbacks of visual servoing controllers is the constant need for the visual characteristics of a target point and prior knowledge of the visual features associated with this desired point. This is why several visual servoing controller techniques rely on a known visual marker \cite{tian2019fog,molnar2020visual,arora2020mobile,qiu2019visual}, and by means of image calibration and marker identification techniques, the position of a target point is known to then establish the best technique of movement towards this point. These investigations focus more on the controller development aspects and leave aside feature extraction and object detection. Therefore, these works have the disadvantage of not operating properly in unstructured environments where there are no predetermined visual markers. For this reason, several robotic motion techniques are based on learning visual servoing that identifies generic characteristic patterns \cite{vivacqua2017self, lagneau2020automatic, ahmadi2020visual,visualNMPC_2024}.

On the other hand, visual servoing controllers based on object recognition and tracking have been developed, where the feature patterns of the targets are obtained by detecting objects with a Neural Network (NN). In \cite{griffin2020video} and \cite{griffin2021depth}, depth estimation of objects is performed by designing a Recurrent NN (DBox) using a generalized representation of bounding boxes and the motion of a camera. Recently, object detection NN based on the YOLO (You Only Look Once) algorithm have been applied for mobile robot visual navigation \cite{bersan2018semantic,dos2019mobile,hu2022object,wu2020optimized}. In \cite{liu2022mgbm} is introduced MGBM-YOLO, an application for visual servoing controller with object detection NN, where the authors propose two YOLOv3 models that are applied to the robotic grasping system of bolster spring based on image-based visual servoing. The MGBM-YOLO visual servoing controller presents a depth estimator depending on the actual area and the desired area of the object's bounding box, therefore, the system must know the object's dimensions beforehand.

The above-reviewed methods are typically applicable in relatively straightforward maneuvers, such as positioning in front of a target. The complexity of these maneuvers can escalate based on the specific robotics application. For example, consider a waste collection scenario using an Ackermann vehicle equipped with a robotic arm positioned at its rear. In this case, the vehicle needs to approach the object (waste) in the front direction and execute precise maneuvers to position itself at the rear for trash pickup using the robotic arm. These complex maneuvers result in the inability to use only traditional controllers. Therefore, combining robotic algorithms and controller approaches is a promising research line for this complex task execution \cite{zhang2019hybrid, abubaker2020mobile, zhang2022new, mondal2022intelligent, fu2022coupling}. In this way, this study researches hybridized controllers to perform complex car-like maneuvers, as in the above example, by exploiting the multi-modality of the LiDAR-camera fusion \cite{munoz2020targetless, paez2023detection}. 

This paper presents ViKi-HyCo, a Hybird-Control approach that combines a Visual Servoing and a kinematic controller for complex maneuvers and the positioning of a car-like robot for waste collection. Using a YOLOv5 NN object detection as a feature detector for the visual servoing controller, the method works in outdoor environments and does not rely on visual markers. In addition, the system includes a spatial target point estimation algorithm with an RGB-D camera for near-sensing tasks or LiDAR-camera fusion for distant detection tasks that are used for a kinematic controller. Our method continuously calculates the four vertices of the desired bounding box of the detected object, so unlike other visual servoing control methods, our algorithm has no prior knowledge of the object dimensions, allowing it to adapt to any object that the NN detects. First, the target point is detected with an object detection NN, and four characteristic points are determined by the detected bounding box. Next, the desired points are calculated using the current object's depth and the desired final depth. The visual servoing controller calculates the camera velocities, which are transformed into robot velocities utilizing the robot's kinematics model. Then, the algorithm determines which controller to use (visual servoing or kinematic) for the robotic system's maneuvers. This algorithm depends on the detection of the object, so the visual servoing controller is used when an object is detected; otherwise, a kinematic controller is used when the target is lost due to image occlusions, failures of the neural object detection network, or if the robot's displacement leaves the detected object outside the camera FOV. Thus, implementing a Hybird-Control solves the problems of a visual servoing controller when performing complex maneuvers by hybridizing with a kinematic controller. 

The main contributions of the paper are the following:
\begin{itemize}

\item A Hybird-Control approach based on a visual servoing and a kinematic controller for complex maneuvers on car-like robots in outdoor environments by exploiting the multi-modality of the LiDAR-camera fusion.
\item The dynamic calculation of the desired bounding box of unknown objects for the visual servoing controller, based on the object bounding box detected by a YOLOv5 NN and the distance estimated by LiDAR-camera fusion for long distances or by an RGB-D camera for short distances.
\item Tests and comparison with another system using a visual servoing controller with object detection with a YOLO NN \cite{liu2022mgbm}.
\end{itemize}

The rest of the paper is organized as follows: In Section \ref{sec:architecture}, we present an overview of complete proposed controller architecture with the subsections: \textit{Robot and Camera Kinematic Model}, the \textit{LiDAR-Camera Fusion}, \textit{Object Localization} and the \textit{Desired Features Estimation} of the detected object. Next, in Section \ref{sec:experiments}, we describe the experimental results of the proposed method using our own real robot BLUE \cite{BLUE2020deeper}, where we compared our method with only the visual servoing controller for forward and backward positioning, and with an algorithm of state-of-the-art. Also, this section shows the run-time of our method. Finally, in Section \ref{sec:cloncusion}, we present the main conclusions obtained from this work and possible future works.

\Figure[ht!](topskip=0pt, botskip=0pt, midskip=0pt)[width=\linewidth]{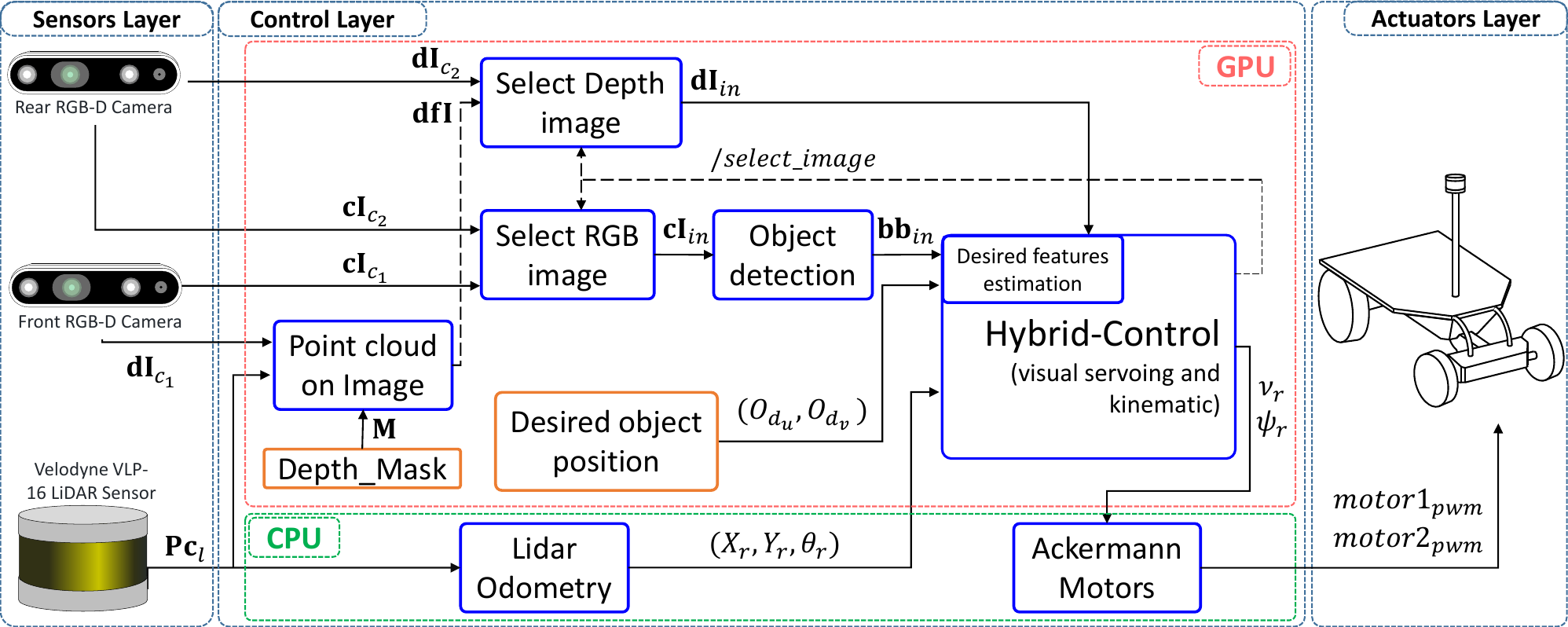}
{Complete ViKi-HyCo pipeline divided into three layers: \textbf{Sensors}: the layer acquiring data from the environment. \textbf{Control}: the sub-processes Point cloud on Image, select depth and RGB image, and object detection take the data from the sensors layer and process them for the Visual Servoing and Kinematic controllers sub-process. The orange boxes are preset parameters; these are the mask for LiDAR and depth camera fusion, and the desired positions of the object in the image plane. The blue boxes are the sub-process in our approach and are detailed in the article. Finally, the LiDAR-based odometry sub-processes for the robot position and the Ackermann Motors sub-process that converts the controller output velocities into velocities to the motors for the \textbf{Actuators} layer.} \label{fig:1_pipeline_ViKi-HyCo}

% \begin{figure*}
% \centerline{\includegraphics[width=\linewidth]{figures/pipeline.pdf}}
% \caption{Complete ViKi-HyCo pipeline divided into three layers: \textbf{Sensors}: the layer acquiring data from the environment. \textbf{Control}: the sub-processes Point cloud on Image, select depth and RGB image, and Object detection take the data from the sensors layer and process them for the Visual Servoing and Kinematic controllers sub-process. The orange boxes are preset parameters, these are the mask for LiDAR and depth camera fusion, and the desired positions of the object in the image plane. The blue boxes are the sub-process in our approach and are detailed in the article. Finally, the LiDAR-based odometry sub-processes for the robot position and the Ackermann Motors sub-process that converts the controller output velocities into velocities to the motors for the \textbf{Actuators} layer.} \label{fig:1_pipeline_ViKi-HyCo}
% \end{figure*}

\section{Proposed Controller Architecture}
\label{sec:architecture}

In this section we define the architecture of the proposed approach, where we start from the control laws of the visual and kinematic servoing controllers. Then, the LiDAR-Camera fusion for the localization of distant objects and the use of an RGB-D camera for close objects are detailed. Next, we describe how to estimate the desired bounding box based on detections by a YOLOv5 NN and object localization. Finally, we define the control law of our ViKi-HyCo hybrid control, which continuously calculates the robot motion velocities and decides which controller to use (kinematic or visual servoing). This approach allows the robot to use a visual servoing controller without interruptions due to occlusions in the image or loss of the detected object, including the change of data input sources for the visual servoing controller, thus achieving complex positioning maneuvers.

\begin{figure}
\centerline{\includegraphics[width=7cm,height=5.25cm]{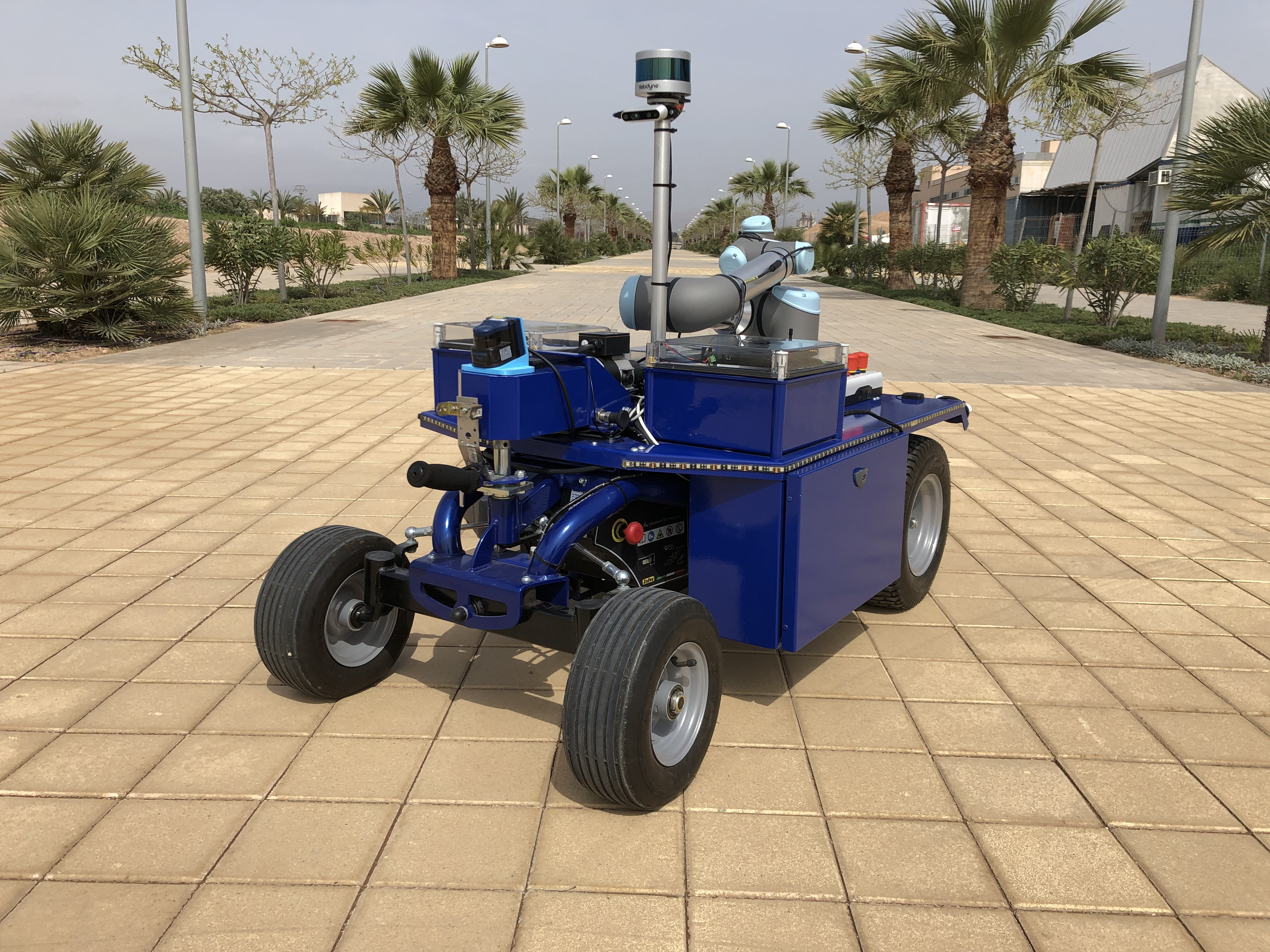}}
\caption{BLUE: roBot for Localization in Unstructured Environments} \label{fig:2_blue_momue}
\end{figure}

\begin{figure}
\centerline{\includegraphics[width=0.5\textwidth]{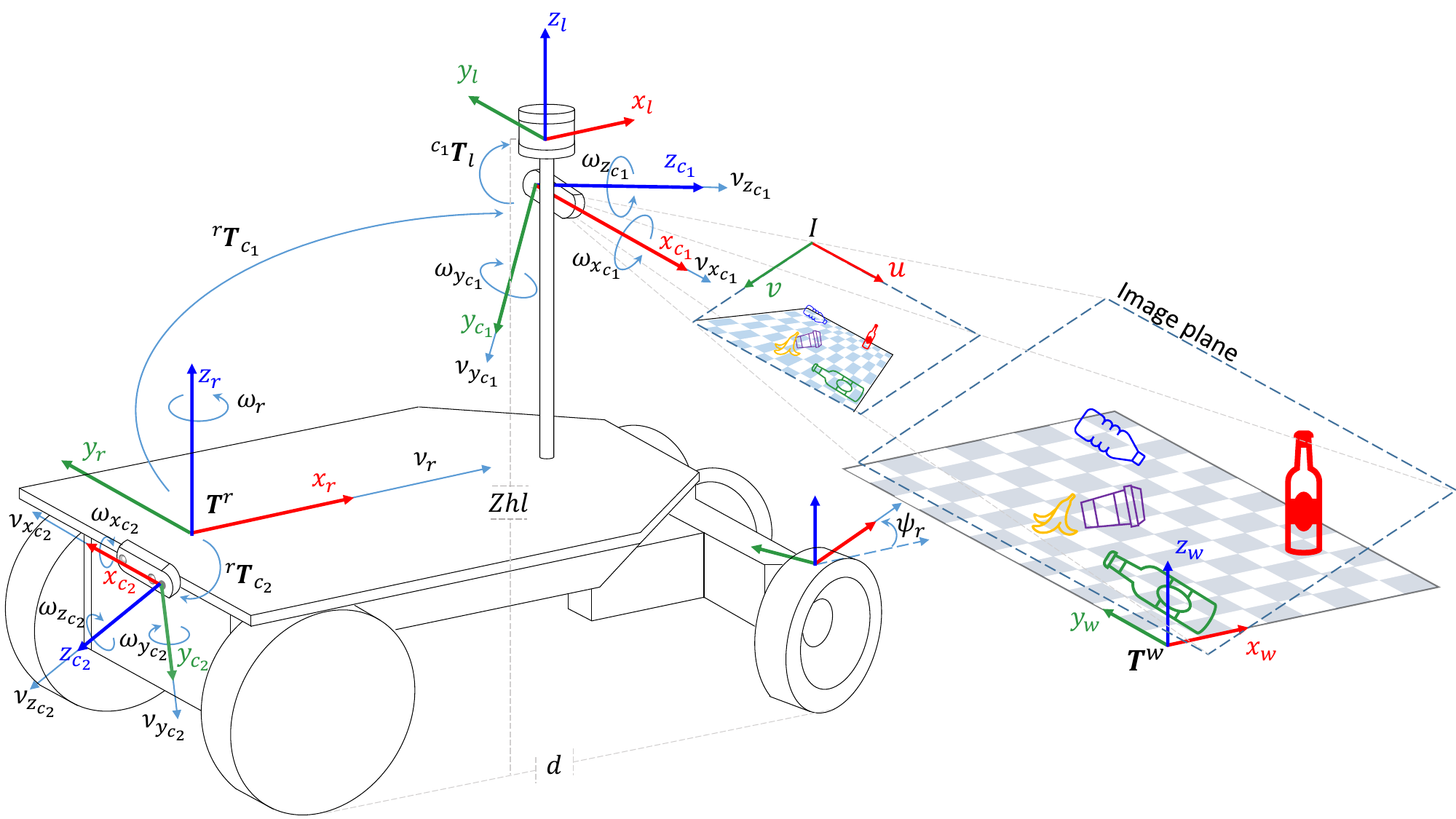}}
\caption{Sensor transformation systems onboard the BLUE robot. The transformation is determined by the letter $\mathbf{T}$ where the upper index is the origin and the lower index is the destination $_{}^{origin}\mathbf{T}_{destination}$.} \label{fig:3_model_system}
\end{figure}

Fig.\ref{fig:1_pipeline_ViKi-HyCo} shows the structure of the proposed system and the sub-processes needed to implement the ViKi-HyCo algorithm. The structure of the system is divided into three layers: Sensors, Control and Actuator layer. The sensor layer takes 2D data from the environment in RGB images, and 3D data from point clouds and depth images of the D channel of the cameras. The Control layer includes the LiDAR odometry process and the ViKi-HyCo controller, which has the following sub-processes: projection of point cloud on image, object detection, visual servoing and kinematic controller. In addition, ViKi-HyCo has a sub-process that depends on object detection and decides the source of information for the visual servoing and kinematic controller calculations. Finally, the velocities are sent from the controller to the robot motors in Actuators layer. To test our approach, the ViKi-HyCo method has been employed in our research platform BLUE shown in the Fig. \ref{fig:2_blue_momue}, which is a non-holonomic robot developed for autonomous navigation in unstructured outdoor environments. The Fig. \ref{fig:3_model_system} shows the location of each of the sensors used in this work. The origin position of each robot sensor is defined by the letter $\mathbf{T}$, where the upper index has the name of the sensor, e.g. $\mathbf{T}^{c_1}$ is the base point of front camera. In addition, the robot's transformation matrices are denoted by the letter $\mathbf{T}$, where an upper index and a lower index indicate the origin and the destination point respectively of the transformation matrix. Thus, the transformation matrix between the robot's base and the front camera is denoted as  ${_{}^{r}\mathbf{T}_{c_1}} $.

\subsection{Robot Kinematic Model}
\label{sec:Robot Kinematic Mode}

\begin{figure}
\centerline{\includegraphics[ width=0.7\linewidth]{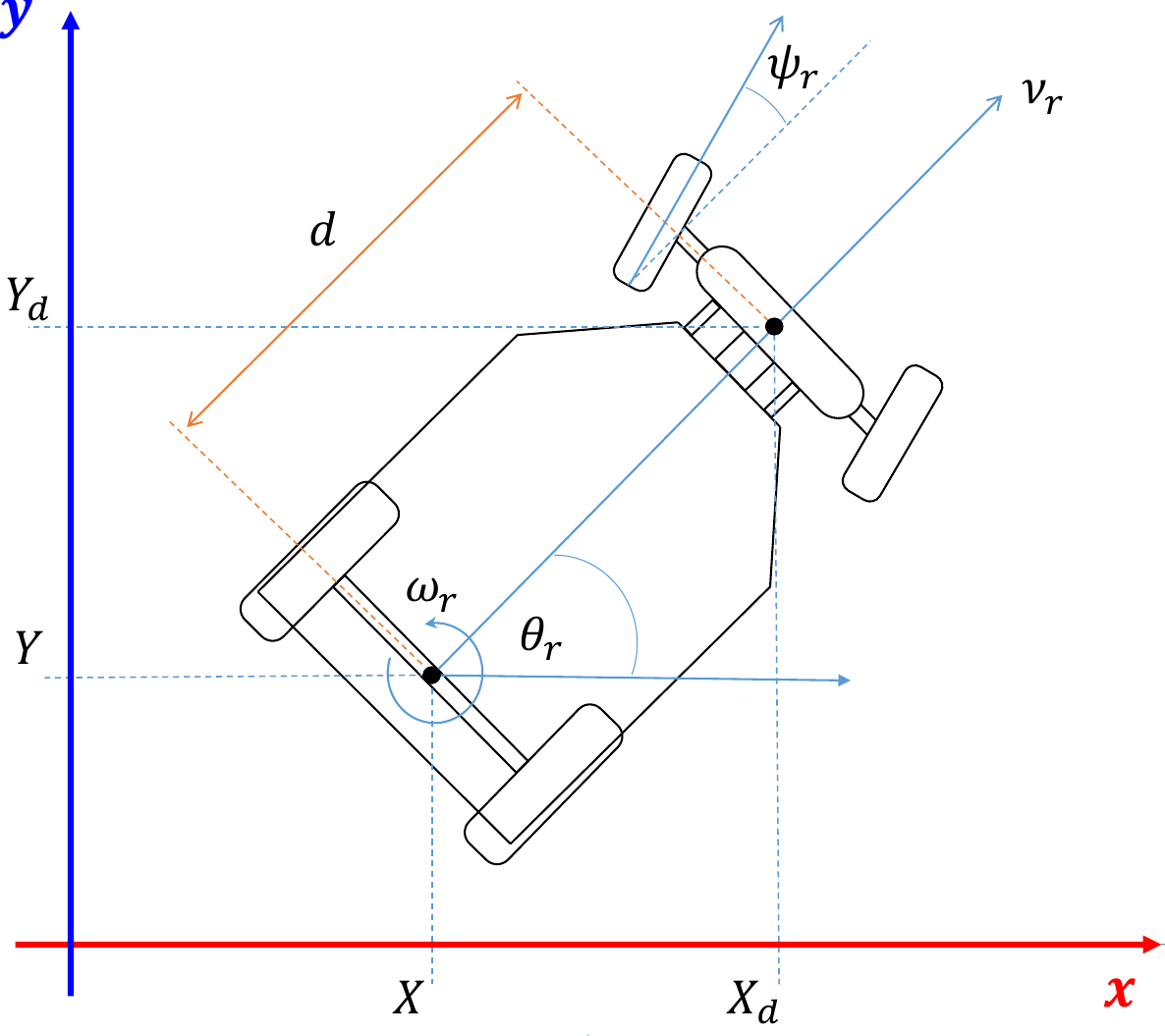}}
\caption{Kinematic model of BLUE robot. } \label{fig:4_kinematic_robot}
\end{figure}

The robotic kinematic model of a non-holonomic robot is widely known in the literature and developed in different applications \cite{corke2011robotics}(in chapter 16.5). In order to generate the control law we use the kinematic model of a car-like robot as shown in Fig \ref{fig:4_kinematic_robot}, where the physical characteristics of the robot are determined. Thus, we have the kinematic modeling of the robot as shown in \eqref{eq:kinematics_blue}: 

\begin{equation}  
    \label{eq:kinematics_blue}
    \mathbf{\dot{h}} = \begin{bmatrix}
    \dot{X}\\ \dot{Y}
    \end{bmatrix}=\begin{bmatrix}
     \cos\theta_r& -d\sin\theta_r\\ 
     \sin\theta_r & d\cos\theta_r\\
    \end{bmatrix} \cdot \begin{bmatrix}
    \nu_r\\ \omega_r
    \end{bmatrix}
\end{equation}

\begin{equation}  
    \label{eq:jacobian_blue}
    \mathbf{J}_b =  \begin{bmatrix}
     \cos\theta_r& -d\sin\theta_r\\ 
     \sin\theta_r & d\cos\theta_r\\
    \end{bmatrix}
\end{equation}

\begin{equation}  
    \label{eq:error_jacobian_vel_blue}
    \mathbf{\dot{h}} = \mathbf{J}_b \cdot \mathbf{V}_r
\end{equation}

Where the velocities necessary to reach a target point are calculated through the position error $\mathbf{\dot{h}} = \mathbf{h} - \mathbf{h_d}  $. Where $\mathbf{h}$ is the current $[X;Y]$ robot's position and $\mathbf{h_d}$ is the desired $[X_d;Y_d]$ position.  Beginning from the Blue robot's kinematics, the Jacobian matrix $\mathbf{J}_b$ \eqref{eq:jacobian_blue} and the velocities $\mathbf{V_r} = [\nu_r;\omega_r]$ are determined. Then feedback law is in \eqref{eq:kinematic_law_control}.

\begin{equation}
   \mathbf{{V}}_r = \mathbf{J}_b^{-1} \cdot (\mathbf{k}_1 \cdot \tanh(\mathbf{\dot{h}}))
   \label{eq:kinematic_law_control}
\end{equation}

The constant $\mathbf{k}_1$ is the positive definite matrices with the controller gains. In addition, the steering angle of the robot is determined  in \eqref{eq:sterring angle}, where $d$ is the distance between the front and rear wheels.

\begin{equation}  
    \label{eq:sterring angle}
    \psi_r = \arctan \begin{pmatrix}
    \frac{d \cdot \omega_r}{ \nu_r}
    \end{pmatrix} 
\end{equation}

The controller velocities are saturated with a tangential function of the error  $\mathbf{\dot{h}}$, thus limiting the velocities to the mechanical characteristics of the robot in the ranges of linear velocity $-0.5<\nu_r<0.5$ m/s and a steering angle $-0.44<\psi_r<0.44$ rad.

The $\theta_r$ orientation and the displacement of the robot in the ${X}$, ${Y}$ axis, shown in \eqref{eq:kinematics_blue}, are determined by a 6DOF LiDAR-based odometry system. This LiDAR odometry system is based on the F-LOAM method \cite{wang2021floam} and adapted to our BLUE robot \cite{velasco2023lilo}. This algorithm minimizes the offset position between a measured point cloud $\mathbf{P_c}$ and a local point cloud map. The positioning of the robot to a desired point is controlled by the LiDAR-based odometry system and the robotic kinematic model.

\subsection{Camera Kinematic Model}
\label{sec:camera_mikematics}
\begin{figure} [t]
\centerline{\includegraphics[width=0.7\linewidth]{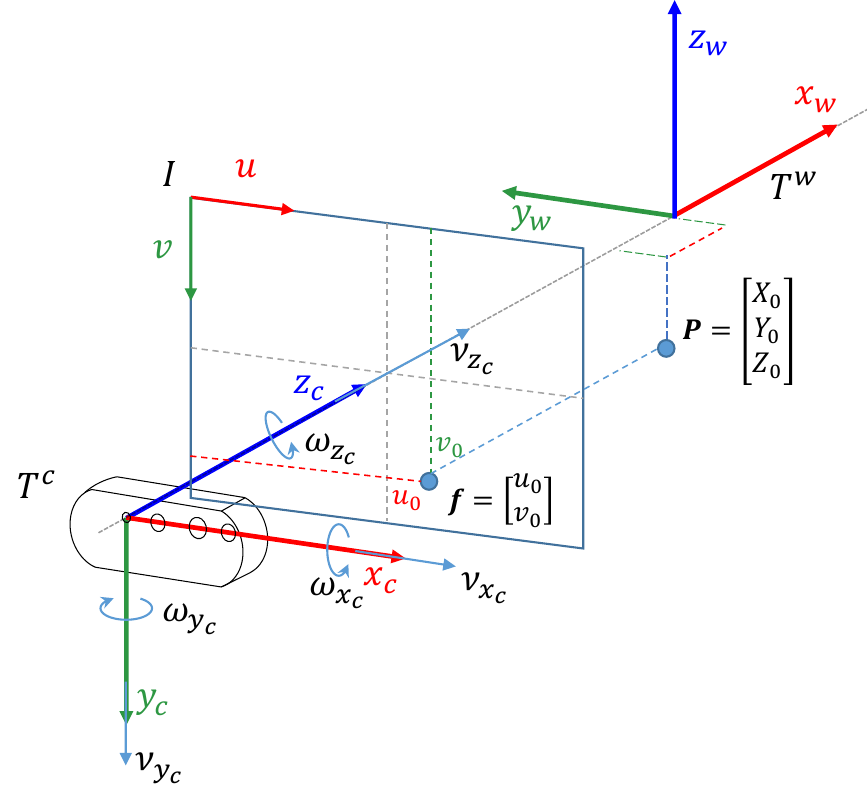}}
\caption{Model of camera system. The point $\mathbf{P}$ corresponds in the image plane to $\mathbf{{f}}$.} \label{fig:5_kinematic_camera}
\end{figure}

Fig. \ref{fig:5_kinematic_camera} shows the kinematic model of the camera we use \cite{vs_paper}. The point $\mathbf{P}=[X_0, Y_0, Z_0]$ corresponds in the image plane to $\mathbf{{f}} = [u_0,v_0]$. Thus, the displacement of a feature point in the image is described by \eqref{eq:kinematics_camera_f0}.

\begin{equation}    
    \label{eq:kinematics_camera_f0}
    \mathbf{\dot{f_0}} = \mathbf{L}_{{f}_{0}} \cdot \mathbf{V}_c
\end{equation}

Where, $\mathbf{L}_{{f}_{0}}$ is the Jacobian matrix and $\mathbf{V}_c$ are the linear and angular velocities $[\nu_{x_c}\, \nu_{y_c}\, \nu_{z_c}\, \omega_{x_c}\, \omega_{y_c}\, \omega_{z_c}]^T$ in each axis-camera . Thus, $\mathbf{L}_{{f}_{0}}$ is defined in \eqref{eq:jacobian_camera}, where $l$ represents the camera focal length.

\begin{equation}
\label{eq:jacobian_camera}
        %\mathbf{\dot{f_0}} = 
       % \begin{bmatrix}
    %    \dot{u_0}\\ \dot{v_0}
    %\end{bmatrix}\!=\!
    \mathbf{L}_{{f}_{0}}\!=\!\begin{bmatrix}
     -\frac{l}{Z_{0}}&\!\!0&\! \frac{u_{0}}{Z_{0}}&\! \frac{u_{0} v_{0}}{l}&\!\!  -(l+\frac{{u_{0}}^{2}}{l})&\!\! v_{0}\\ 
     0&\!\! -\frac{l}{Z_{0}}&\! \frac{v_{0}}{Z_{0}}&\! l+\frac{{v_{0}}^{2}}{l}&\!\! -\frac{u_{0} v_{0}}{l}&\!\! -u_{0}\\
    \end{bmatrix}
\end{equation}

Therefore, for $\mathbf{f}$th feature points the camera kinematic model is represented in \eqref{eq:kinematics_camera_fth}.

\begin{equation}
\label{eq:kinematics_camera_fth}
   \mathbf{\dot{f}} = \mathbf{L}_{{f}} \cdot \mathbf{V}_c 
\end{equation}

Here, the Jacobian matrix of the camera for $n$ features points $\mathbf{L}_{{f}}$ is detailed in \eqref{eq:jacobian_camera_fth}.

\begin{equation}
\label{eq:jacobian_camera_fth}
    \mathbf{L}_{{f}}\!=\!\begin{bmatrix}
     -\frac{l}{Z_{0}}&\!\!0&\! \frac{u_{0}}{Z_{0}}&\! \frac{u_{0} v_{0}}{l}&\!\!  -(l+\frac{{u_{0}}^{2}}{l})&\!\! v_{0}\\ 
     0&\!\! -\frac{l}{Z_{0}}&\! \frac{v_{0}}{Z_{0}}&\! l+\frac{{v_{0}}^{2}}{l}&\!\! -\frac{u_{0} v_{0}}{l}&\!\! -u_{0}\\
\vdots&\vdots&\vdots&\vdots &\vdots &\vdots\\
-\frac{l}{Z_{n}}&\!\!0&\! \frac{u_{n}}{Z_{n}}&\! \frac{u_{n} v_{n}}{l}&\!\!  -(l+\frac{{u_{n}}^{2}}{l})&\!\! v_{n}\\ 
     0&\!\! -\frac{l}{Z_{n}}&\! \frac{v_{n}}{Z_{n}}&\! l+\frac{{v_{n}}^{2}}{l}&\!\! -\frac{u_{n} v_{n}}{l}&\!\! -u_{n}
    \end{bmatrix}
\end{equation}

The control law for a visual servoing controller is shown in equation \eqref{eq:law_control_servoVisual}, where the feature error $\mathbf{\dot{f}}=(\mathbf{f}-\mathbf{f}_d)$, is the displacement of the current $\mathbf{f}$ and desired $\mathbf{f}_d$ features, and the pseudo inverse of the image Jacobian is $(\mathbf{L}_{f})^\dagger$. In addition, we have $\mathbf{\lambda}$ as the positive definite matrix of the control gains.

\begin{equation}
    \mathbf{V_c} = - \mathbf{\lambda} (\mathbf{L}_{f})^\dagger \cdot \mathbf{\dot{f}} 
    \label{eq:law_control_servoVisual}
\end{equation}  
The control law expressed in equation \eqref{eq:law_control_servoVisual}, calculates the motion velocities of the camera to decrease the error between actual and desired visual features, however, to convert these velocities to motion velocities of the robot, it is necessary the physical characteristics of the robot by applying its kinematic model. This will be addressed in the hybrid controller section. 

\subsection{LiDAR and Frontal Camera Fusion}
\label{sec:lidar_camera_fusion}
The fusion of several optical sensors is nowadays clearly helpful for autonomous navigation applications, to identify distances to target points or for mapping \cite{munoz2020targetless}. The use of a LiDAR sensor and RGB cameras requires a first calibration step that has been widely studied in the literature \cite{Zhou_calibration,dhall2017lidar}. For this purpose, there are tools that allow a calibration of the LiDAR sensor and an RGB camera. We calibrate our LiDAR sensors and camera using an application developed in ROS \cite{dhall2017lidar}. This method finds a rigid body transformation to determine the extrinsic parameters of a LiDAR and a camera using 3D-3D point correspondences. In this way, we obtain the transformation matrix $_{}^{c_1}\mathbf{T}_{l}= (_{}^{c_1}\mathbf{R}_{l},  \;_{}^{c_1}\mathbf{t}_{l})$. This transformation matrix with the intrinsic camera parameter matrix $\mathbf{M}_{c_1}$, converts point cloud data $ \mathbf{Pc}_l = [ X_l,  Y_l,  Z_l]$ into image plane projections at points $[\mathbf{u}_l ,\mathbf{v}_l]$, as shown in \eqref{eq:lidar_camera_calibration}. 

\begin{equation}
    \begin{matrix}
        \begin{bmatrix}
            u\\ 
            v\\
            e
            \end{bmatrix} = \mathbf{M}_{c_1} \begin{bmatrix}
             {_{}^{c_1}\mathbf{R}_l} & {_{}^{c_1}\mathbf{t}_l}\\ 
             \mathbf{0} & 1
            \end{bmatrix}\cdot\begin{bmatrix}
            X_l\\ 
            Y_l\\ 
            Z_l\\ 
            1
        \end{bmatrix}
        \\
            \mathbf{u}_l = \vec{u} \odot (\,\vec{e}\,)^{-1}\\
            \mathbf{v}_l = \vec{v} \odot (\,\vec{e}\,)^{-1}\\
            \mathbf{u}_l :=  \left \{ u_l:u_l \in \mathbb{Z}^+ , 0<u_l<W_{img} \right \}\\
            \mathbf{v}_l :=  \left \{ v_l:v_l \in \mathbb{Z}^+ , 0<v_l<H_{img}  \right \}
    \end{matrix}
     \label{eq:lidar_camera_calibration}
\end{equation}

Where $W_{img}$ and $H_{img}$ are the width and height of the image plane. Hence, we generate the depth image $\mathbf{dI}_l$ with the depth modulus of each $\mathbf{\mathbb{R}^3}$ coordinate of the $\mathbf{Pc}_l$ point cloud with $[\mathbf{u}_l ,\mathbf{v}_l]$ coordinates of the image plane as shown in \eqref{eq:depth_lidar_Image}.
\begin{equation}
\label{eq:depth_lidar_Image}
    \mathbf{dI}_{l(u_l,v_l)} = \left \| X_l+Y_l+Z_l \right \|_2
\end{equation}

\begin{figure} [b]
    \centering
  \subfloat[LiDAR point cloud colored with the RGB data from the camera.\label{fig:lidar_camera_fusion_a}]{%
       \includegraphics[width=0.47\linewidth]{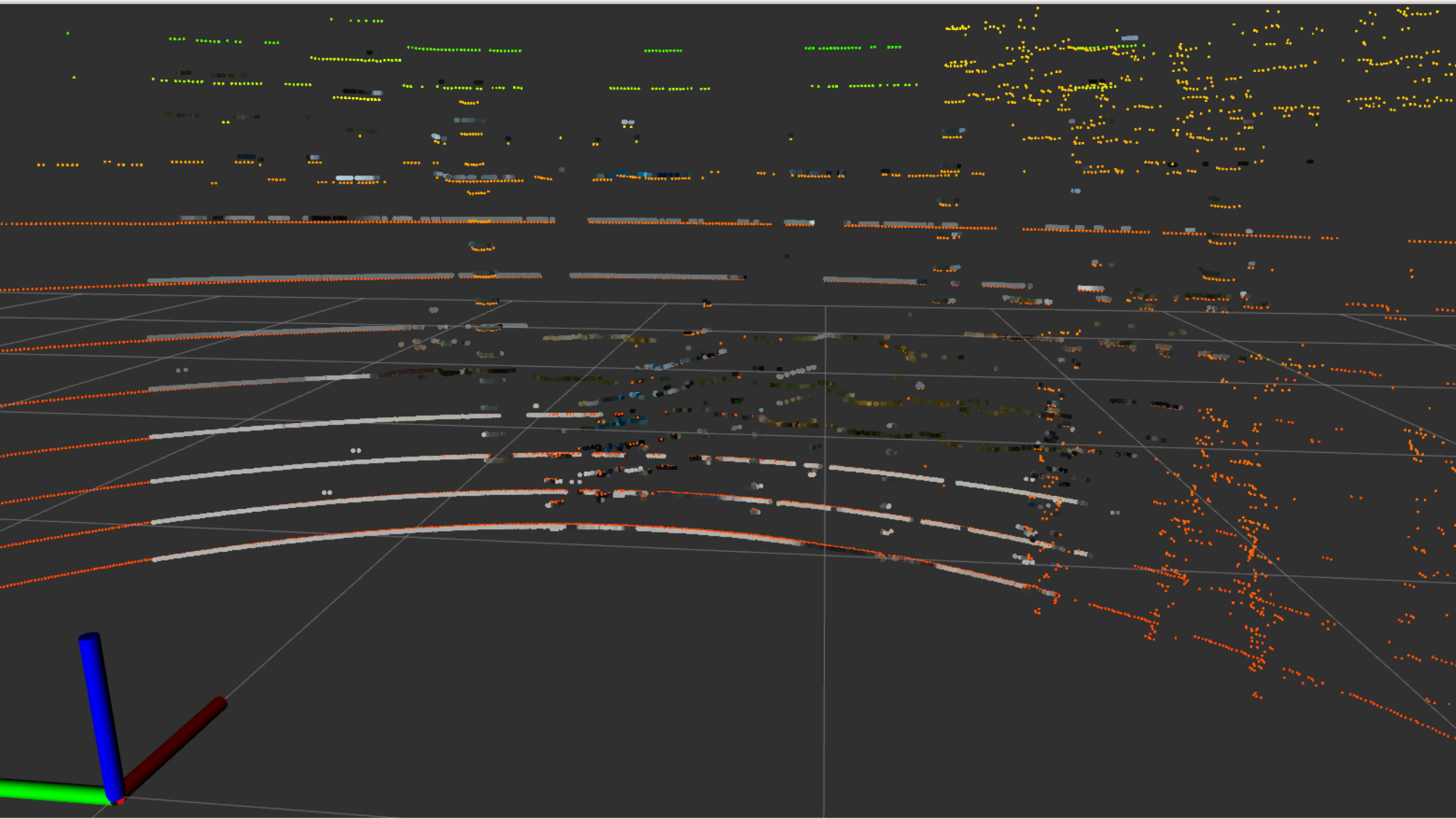}}
    \hfill
  \subfloat[LiDAR point cloud on RGB image.\label{fig:lidar_camera_fusion_b}]{%
        \includegraphics[width=0.47\linewidth]{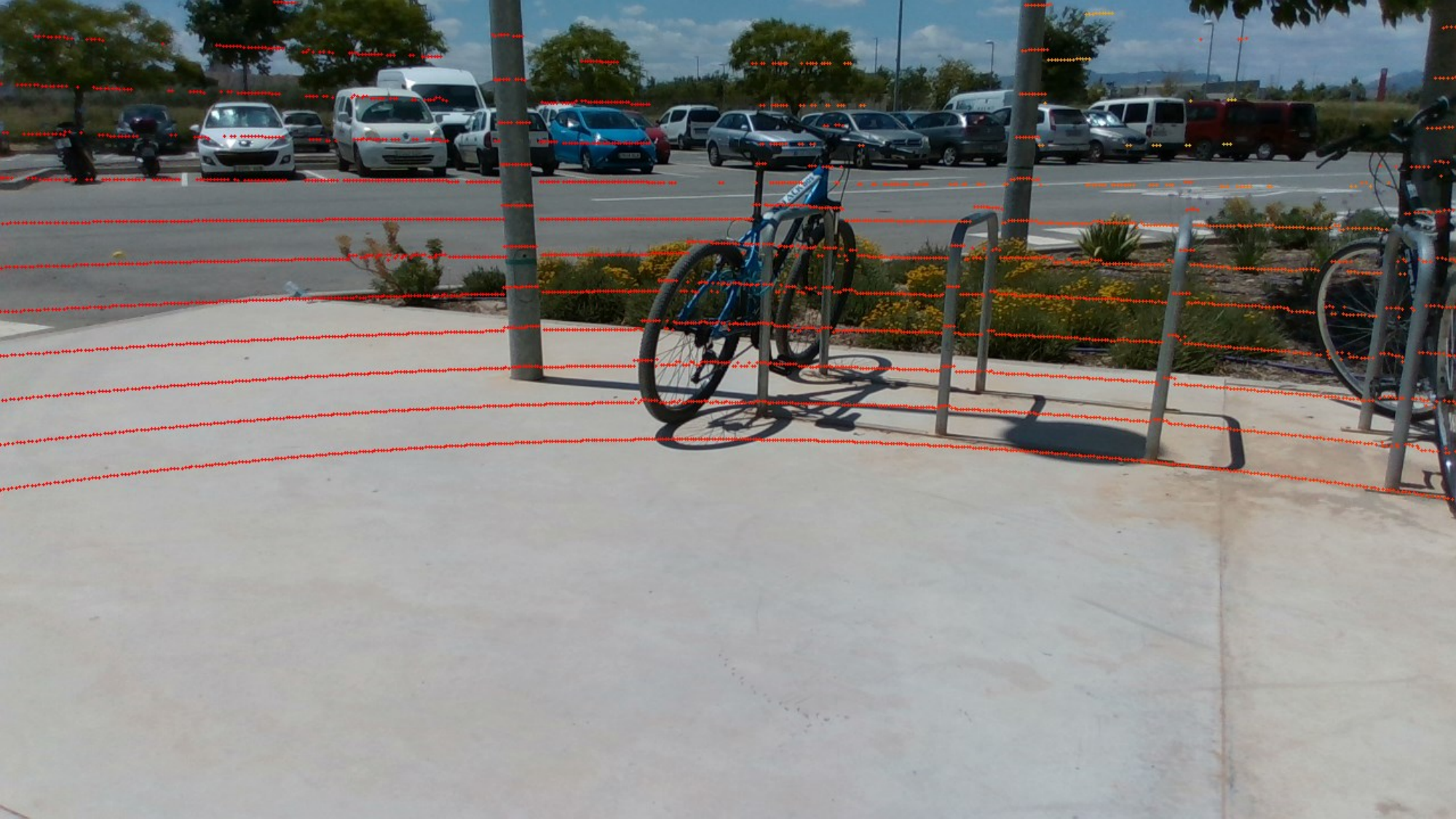}}
    \\
  \subfloat[LiDAR interpolated point cloud colored with camera RGB data.\label{fig:lidar_camera_fusion_c}]{%
        \includegraphics[width=0.47\linewidth]{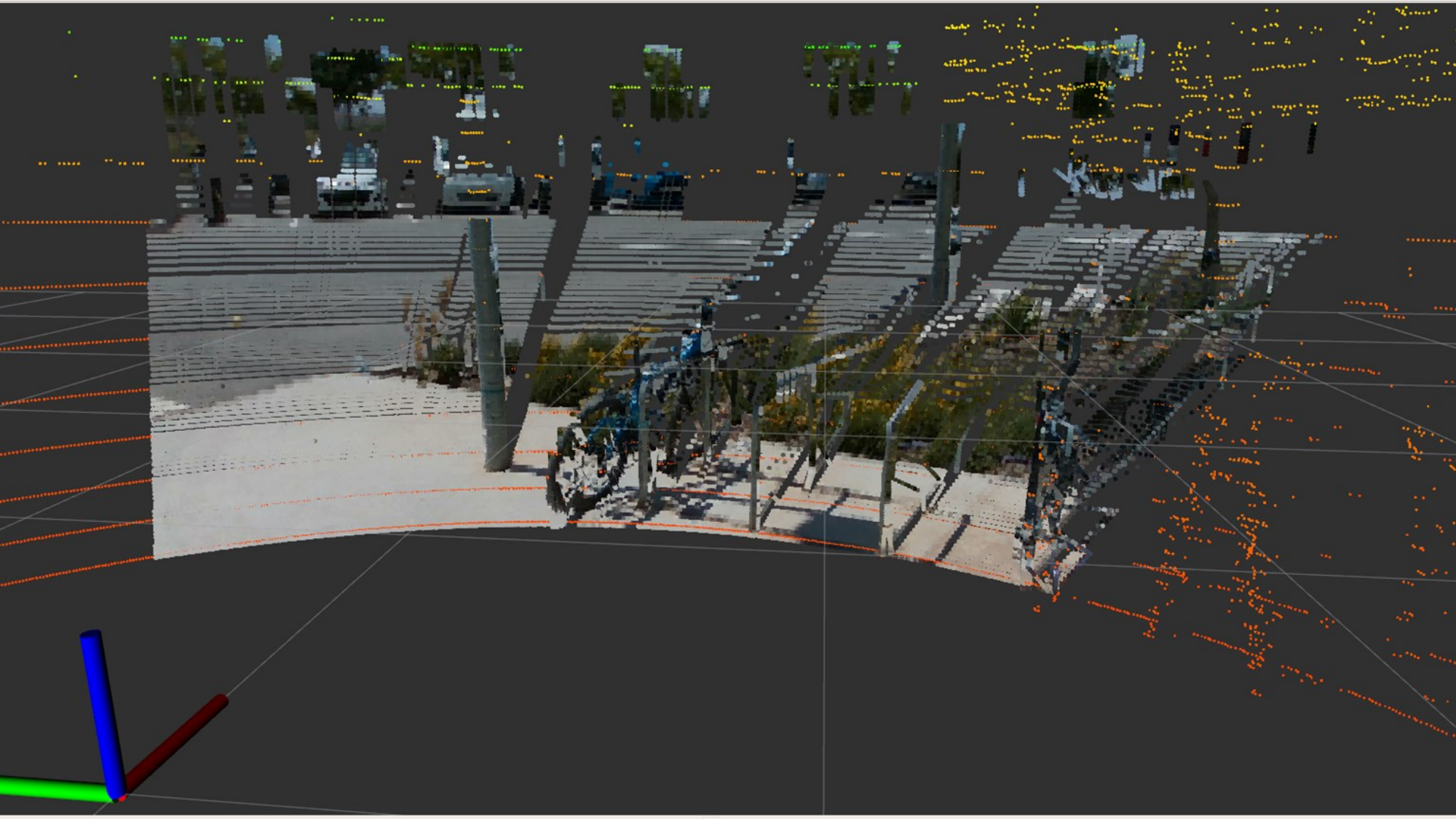}}
    \hfill
  \subfloat[LiDAR interpolated point cloud on RGB image\label{fig:lidar_camera_fusion_d}]{%
        \includegraphics[width=0.47\linewidth]{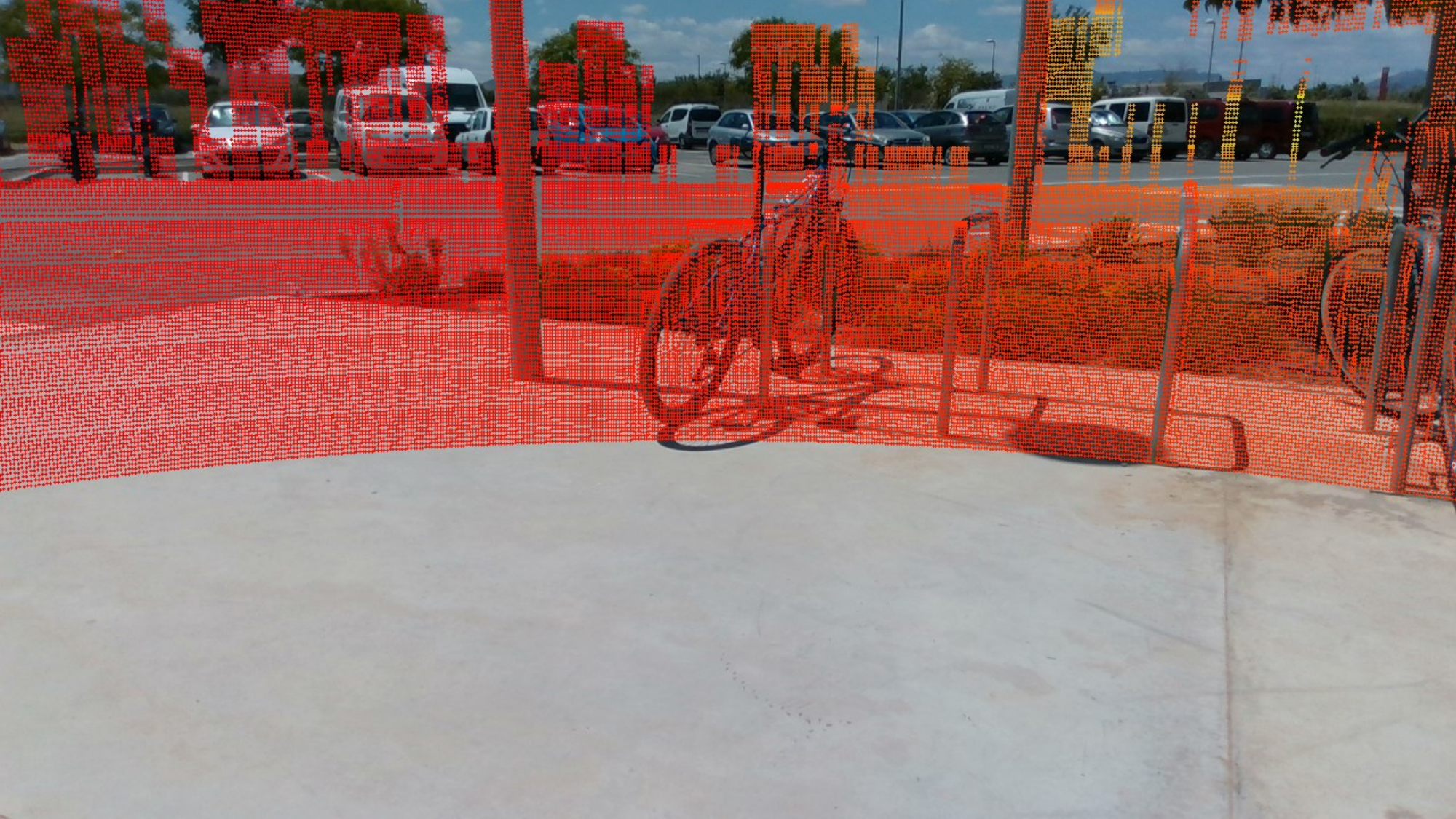}}
  \caption{LiDAR and fusion camera on the point clouds and in the image plane.}
  \label{fig:lidar_camera_fusion} 
\end{figure}

LiDAR sensors are not able to obtain depth information of small objects because they generate a point cloud in layers and this is not as dense as a depth camera. In our case, we use a 16-layer LiDAR sensor (see Fig. \ref{fig:lidar_camera_fusion_a}). Therefore, we augmented the point cloud data with a linear interpolation between each laser beam of the LiDAR sensor. In this way, we generate virtual point cloud data by interpolating the real point cloud  $\mathbf{Pc}_l \in \mathbf{\mathbb{R}^3}$ into a $\mathbf{\mathbb{R}^2}$ range image and using 2D linear interpolation \cite{kirkland2010bilinear}. These data are converted back to $\mathbf{\mathbb{R}^3}$ points, obtaining a new interpolated point cloud as shown in Fig. \ref{fig:lidar_camera_fusion}c.

%\begin{figure}[!t]
%    \centerline{\includegraphics[clip, trim=0cm 0.4cm 4.19cm 0cm, width=8.5cm,height=5.34cm]{figures/lidar_camera_fusion.pdf}}
%    \caption{(a)nube de puntos sin interpolar(b)point cloud on image(c)nube de puntos interpolada(d)point cloud on image} 
%    \label{fig:lidar_camera_fusion}
%\end{figure}

As shown in the Fig. \ref{fig:lidar_camera_fusion}b and Fig. \ref{fig:lidar_camera_fusion}d, the RGB image $\mathbf{cI}_{c1}$ has an area where the LiDAR cannot be projected, this is due to the location of the LiDAR sensor with respect to the front camera, so it is not possible to know the distance information of this area (called LiDAR blind spots). The camera has a depth channel that generates a point cloud working at a maximum distance range of five meters. Therefore, we combine the depth image $\mathbf{dI}_{c_1}$ with the depth data of the LiDAR point cloud already calibrated and interpolated. In this way, the depth channel of the RGB-D camera is used to know the distance of close objects, and the LiDAR-camera fusion is used for distant objects.

To combine both images, we use a mask that removes the depth information from $\mathbf{dI}_{c_1}$ where the LiDAR points are projected. We define the mask as all points with coordinates $X$, $Y$ and with height of the LiDAR sensor to the ground plane $Zhl = -110 \,\text{cm}$ that are within a radius $R = 410.52\,\text{cm}$. In this way, by means of the equation \eqref{eq:lidar_camera_calibration} we generate a 2d projection in the image plane of the LiDAR blind spots on the ground. To generate the pixel coordinates of the mask, the values of $-411\,\text{cm}\leq X \leq 411\,\text{cm}$ and $-411\,\text{cm}\leq Y\leq 411\,\text{cm}$ in \eqref{eq:coordenadaslidar_camera} are sampled at 0.1 cm.
\begin{equation}
\begin{matrix}

\mathbf{u}_m , \mathbf{v}_m := 
    &\begin{Bmatrix}
    u_m, v_m : (u, v) \in \mathbb{Z}^+, \\ \forall (X,Y) \!:\! X^2\!+\!Y^2\!+\!Zhl\!=\!R^2
    \end{Bmatrix}
\end{matrix}
    \label{eq:coordenadaslidar_camera}
\end{equation}

\begin{figure}
    \centering
  \subfloat[LiDAR interpolated on the RGB image.\label{fig:lidar_blind_spots_a}]{%
       \includegraphics[width=0.4\linewidth]{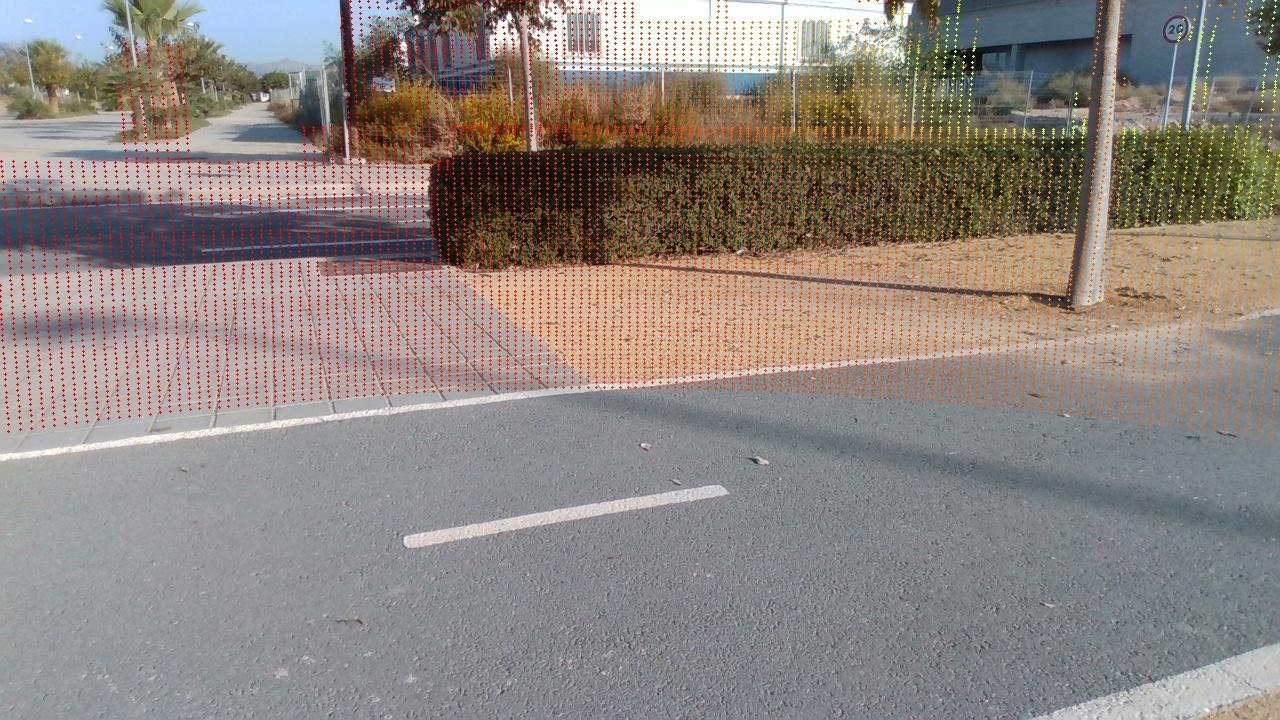}}
    \hfill
  \subfloat[LiDAR blind spot mask.\label{fig:lidar_blind_spots_b}]{%
        \includegraphics[width=0.4\linewidth]{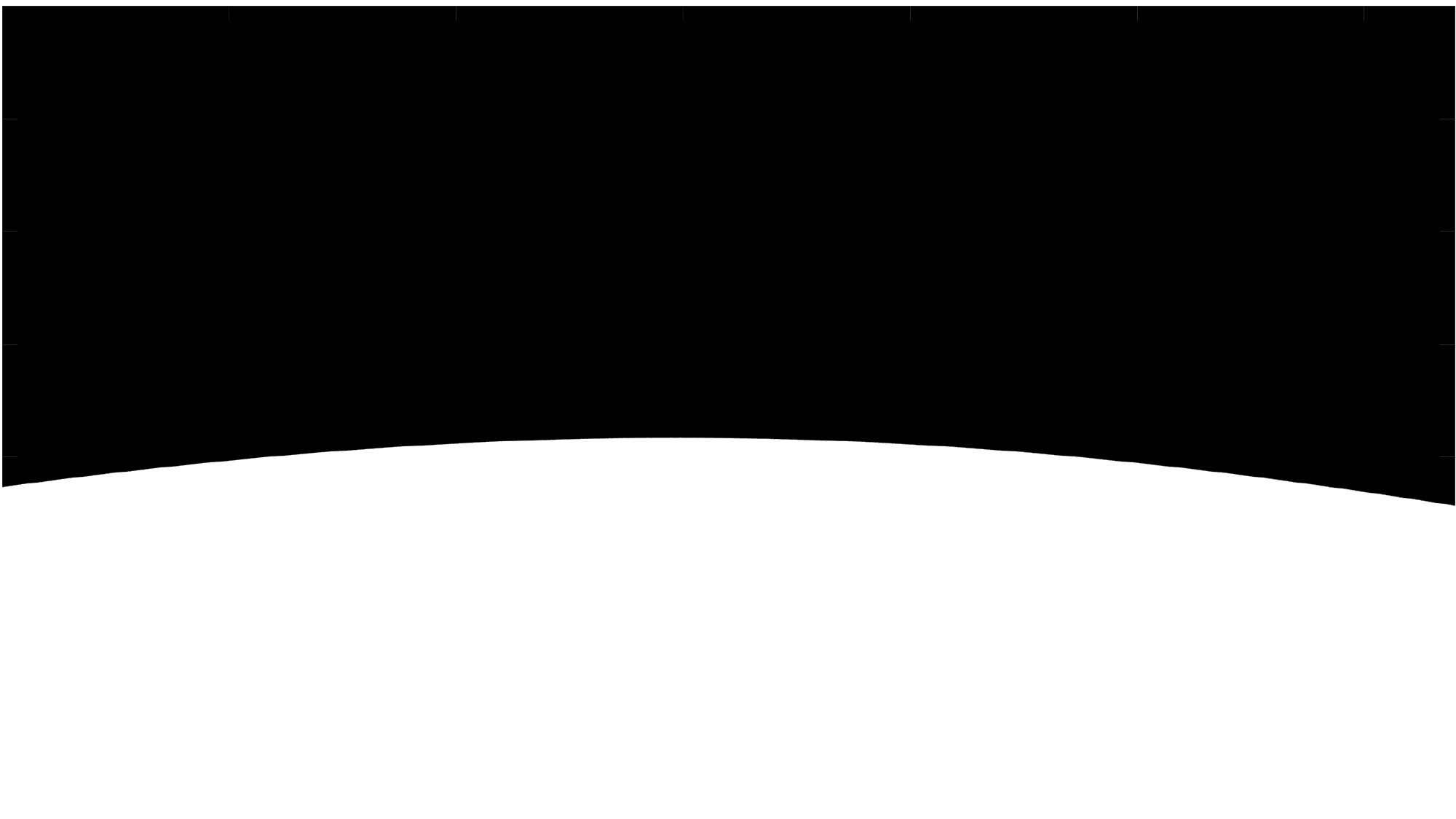}}
    \\
  \subfloat[LiDAR depth points projected on the image plane. \label{fig:lidar_blind_spots_c}]{%
        \includegraphics[width=0.4\linewidth]{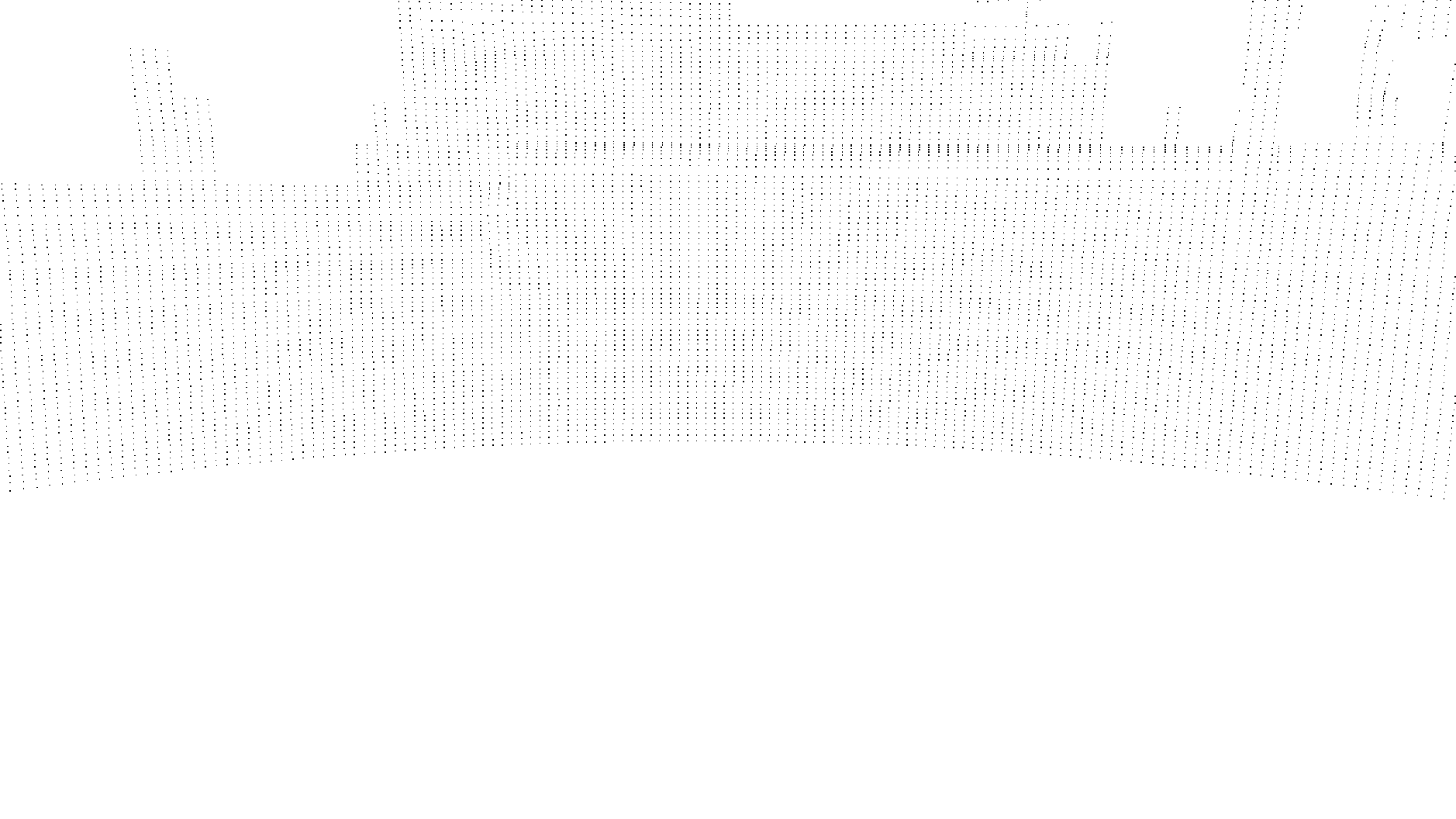}}
    \hfill
  \subfloat[Front camera depth channel image.\label{fig:lidar_blind_spots_d}]{%
        \includegraphics[width=0.4\linewidth]{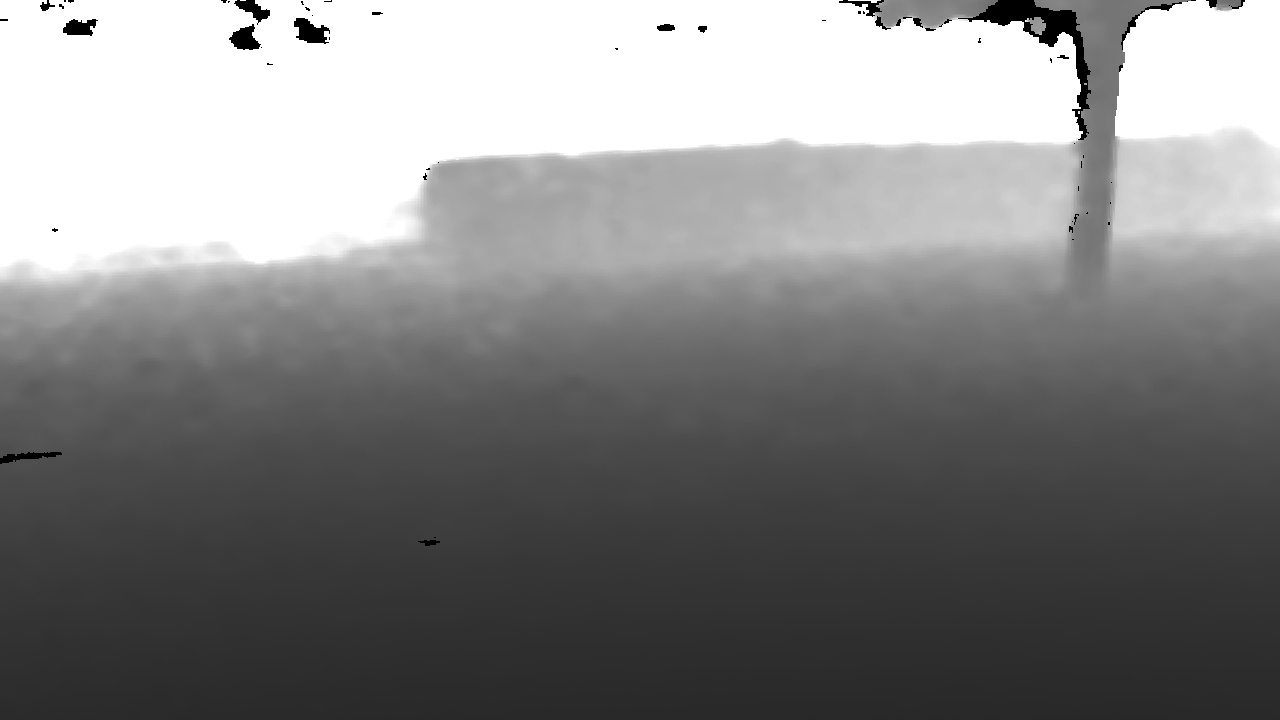}}
        \\
     \subfloat[ Depth fusion of \textbf{(c)} and \textbf{(d)} depths applied to the mask of \textbf{(b)}\label{fig:lidar_blind_spots_e}]{%
        \includegraphics[width=.9\linewidth]{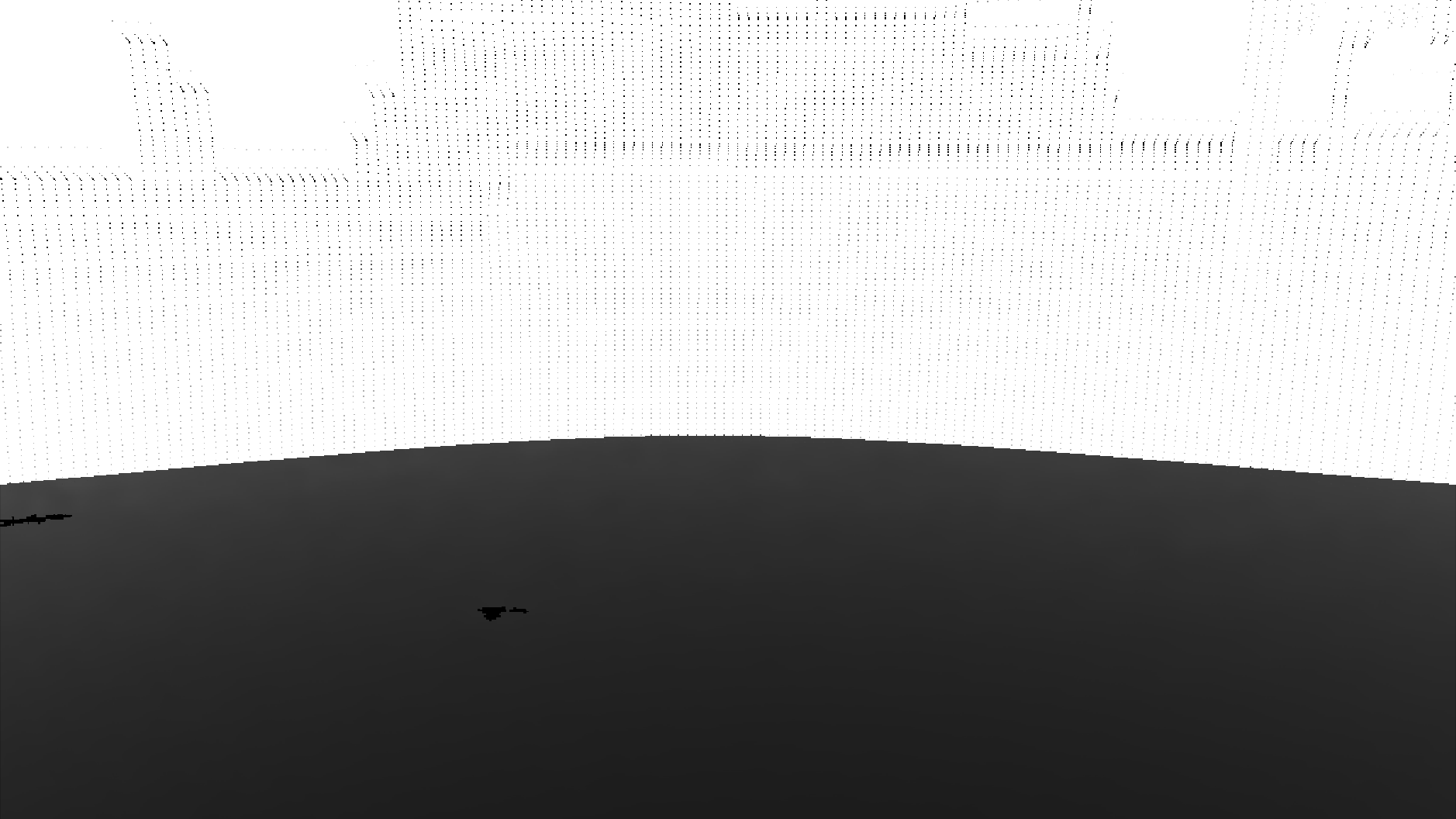}}    
  \caption{Fusion of LiDAR at depths in the image plane and the camera depth channel. \textbf{Note}. Figures (c) and (e) show the colors inverted in the projection area of the LiDAR points for a better visual representation.}

  \label{fig:lidar_blind_spots} 
\end{figure}

We define the mask $\mathbf{M}$ \eqref{eq:mask_lidar_camera} (Fig. \ref{fig:lidar_blind_spots_b}) as a matrix of size $W_{img} \times H_{img}$, where the pixels with coordinates $(\mathbf{u}_m, \mathbf{v}_m)$ \eqref{eq:coordenadaslidar_camera} are 1 and the otherwise are 0. Finally, we obtain the fused depth image $\mathbf{dfI}$ \eqref{eq:depth_image_mask} combining the LiDAR depth image $\mathbf{dI}_l$ (Fig.\ref{fig:lidar_blind_spots_c}) and the front camera depth image $\mathbf{dI}_{c_1}$ (Fig.\ref{fig:lidar_blind_spots_d}) filtered by the element-to-element product $\odot$ with the mask $\mathbf{M}$.

\begin{equation}
        \mathbf{M}=\left\{\begin{matrix}
             1& \forall({u_m,v_m}) \in \mathbb{Z}^+\\ 
             0& \text{otherwise} 
        \end{matrix}\right.
    \label{eq:mask_lidar_camera}
\end{equation}

\begin{equation}
   \mathbf{dfI} = \mathbf{dI}_l + ( \mathbf{M} \odot \mathbf{dI}_{c_1})
    \label{eq:depth_image_mask}
\end{equation}

%\begin{figure}[h]
%    \centerline{\includegraphics[clip, trim=0cm 0.93cm 19.84cm 0cm, width=8cm,height=10.89cm,angle=0]{figures/lidar_camara_mask.pdf}}
%    \caption{} \label{fig:lidar_blind_spots}
%\end{figure}

The Fig. \ref{fig:lidar_blind_spots}e shows the depth information of the robot's front stage and is used for the forward visual servoing control. For the case of reverse robot positioning, the visual servoing controller uses the depth image $\mathbf{dI}_{c_2}$ from the rear camera, because it has a FOV of four meters in the floor plane, which is sufficient for accurate depth measurements to obtain the object's spatial coordinates. Then, we define $\mathbf{dI}_{in}$ as the depth image selected between $\mathbf{dfI}$ for forward visual servoing controller and $\mathbf{dI}_{c_2}$ for rearward visual servoing control.

\subsection{Object Localization}
\label{sec:Object Localization}
Locating a positioning target point is possible with the depth information of robot's environment. We define as target points the objects detected with a YOLOv5 object detection NN. In particular, we start from the work developed in \cite{tornero2022detection}, where objects are detected and localized with respect to our research platform BLUE with the camera depth information and YOLOv5 objects detection in outdoor environments. The results of objects detection were a mAP@.5 around 0.99 and for a mAP@.95 over 0.84 with an average error smaller than 0.25 m.

On the contrary to method in \cite{tornero2022detection}, where the camera-object distance is the value of the center of the bounding box detected in a depth image, on the basis of research \cite{paez2023detection}, we define the camera-object distance with the average value in the bounding box detected in the depth image $\mathbf{dI}_{in}$. 

\begin{figure}
\centerline{\includegraphics[ width=0.9\linewidth]{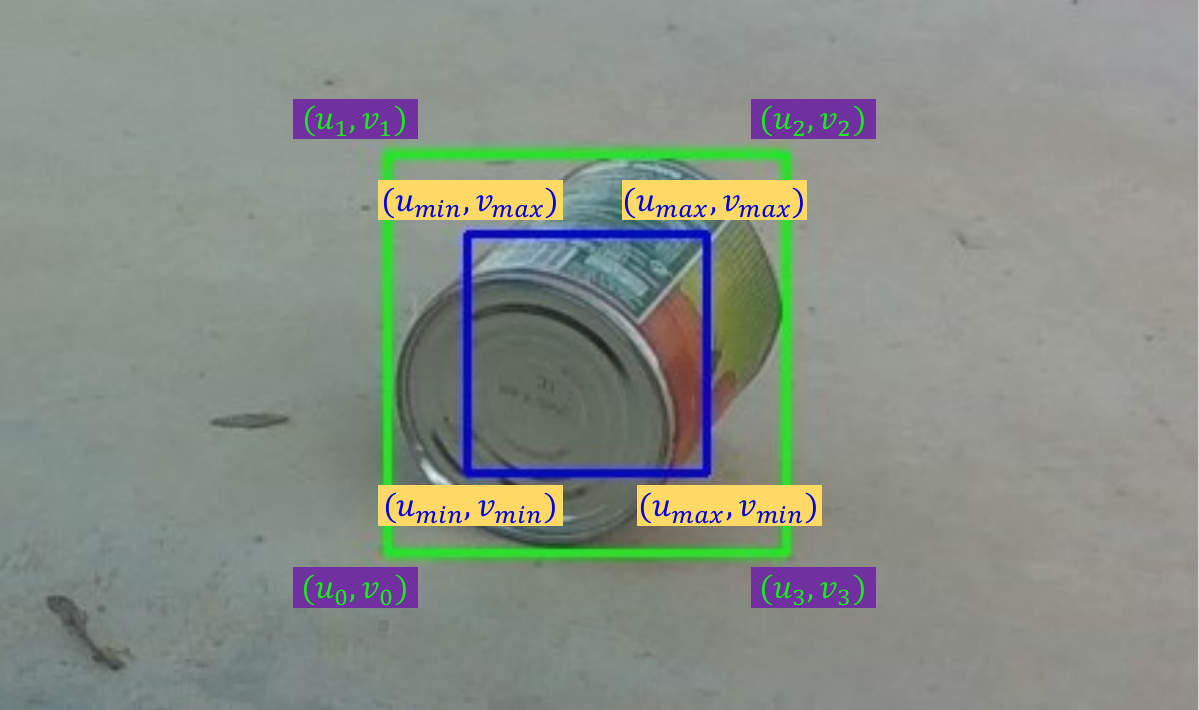}}
\caption{The green box is the bounding box of the object detected with the YOLOv5 NN. The blue box is the 40\% reduction of the original bounding box.} \label{fig:bb_limits}
\end{figure}

Due to irregularities of the detected object some points inside the bounding box may not belong to the object (see Fig. \ref{fig:bb_limits}). For this reason, to avoid measurement errors, a new bounding box $\mathbf{bb}'$ is generated that is 40\% smaller than the original bounding box $\mathbf{bb}$, where the new bounding box coordinates are defined in the equation \eqref{eq:new_bb} . Thus, the camera-object distance $d_o$ is defined as the average of the values inside the new bounding box that are different from 0 \eqref{eq:distance_object}. 

\begin{equation}
    \begin{aligned}
     \mathbf{bb} &=
        \begin{Bmatrix}
             (u_{1},v_{1})& (u_{2},v_{2})\\ 
             (u_{0},v_{0})& (u_{3},v_{3})
        \end{Bmatrix} \\
     u_{min} &= \left \lfloor0.4\left ( 0.5 \times (u_2-u_0) \right )+ u_0\right \rceil\\
     u_{max} &= \left \lfloor0.4\left ( 0.5 \times (u_0-u_2) \right )+ u_2\right \rceil\\
     v_{min} &= \left \lfloor0.4\left ( 0.5 \times (v_2-v_0) \right )+ v_0\right \rceil\\
     v_{max} &= \left \lfloor0.4\left ( 0.5 \times (v_0-v_2) \right )+ v_2\right \rceil\\
     \mathbf{bb}' &=
        \begin{Bmatrix}
             (u_{min},v_{max})& (u_{max},v_{max})\\ 
             (u_{min},v_{min})& (u_{max},v_{min})
        \end{Bmatrix} \\
    \end{aligned}\\
    \label{eq:new_bb}
\end{equation}

\begin{equation}
    d_o = \frac{1}{n} \sum_{v=v_{min}}^{v_{max}} \sum_{u=u_{min}}^{u_{max}} [ \mathbf{dI}_{in}(u,v)\neq0] 
    \label{eq:distance_object}
\end{equation}

Therefore, the position of an object is defined as $\mathbf{P}_o \in \mathbf{\mathbb{R}^3}$ \eqref{eq:objetct_localization}, where $O_{o_u}$ and $O_{o_v}$ are the coordinates in the image plane of the center of the detected object's bounding box, and the values of $(u_c,v_c)$ and $(l_u,l_v)$ are the camera's optical center and focal lengths respectively.

\begin{equation}
   \mathbf{P_o} = \left\{\begin{aligned}
   & Z_o = d_o \\
   & X_o = \frac{(O_{o_u}-u_c)d_o}{l_u} \\
   & Y_o = \frac{(O_{o_v}-v_c)d_o}{l_v} 
\end{aligned}\right.
\label{eq:objetct_localization}
\end{equation}

\subsection{Desired Features Estimation}

The control law of a visual servoing controller, shown in section \ref{sec:camera_mikematics} in equation \eqref{eq:law_control_servoVisual}, calculates the camera velocities $\mathbf{V}_c$ to minimize the error $\mathbf{\dot{f}}$ of desired visual characteristics and current characteristics with the pseudo inverse image Jacobian $(\mathbf{L}_{f})^\dagger$. As detailed in \cite{corke2011robotics} in chapter 11.2.3, at least three features are necessary in the image, where these cannot be col-linear, so there are no singularities in the image Jacobian matrix \eqref{eq:jacobian_camera}. We use the four corners of the object bounding box detected as current features $\mathbf{f}$ at each iteration and obtain the desired features $\mathbf{f}_d$ from the desired center coordinates $(O_{d_u}, O_{d_v})$ in the plane image, the width  $w_o$ and height $h_o$ of the current object bounding box, the variable $k$, which is the ratio of the current camera-object depth $Z_o$ \eqref{eq:distance_object} and the value $Z_d$ defined as the depth at each iteration in image $\mathbf{dI}_{in}$ at the coordinates $(O_{d_u}, O_{d_v})$. In this way, we calculate the coordinates $\mathbf{f}_d =\{u_d,v_d \}$ of each desired feature in \eqref{eq:desired_feature}, where the ratio is defined as $k = Z_{o}\times Z_{d}^{-1}$. 

\begin{equation}  
\begin{matrix}
    \begin{matrix}\\
    u_d = O_{d_u} \pm \frac{w_d}{2}  & , & v_d = O_{d_v} \pm \frac{h_d}{2}\\
    \end{matrix}\\
    \begin{matrix}
    w_d = w_o \times k &,& h_d = h_o \times k\\
     \end{matrix}\\
\end{matrix}
\label{eq:desired_feature}
\end{equation}

In this work, the implemented image Jacobian \eqref{eq:jacobian_camera} uses the same distance $Z_o$ shown in \eqref{eq:distance_object} for each feature $\mathbf{f}$. Thus, the velocities $\mathbf{V}_c$ \eqref{eq:law_control_servoVisual} calculated by the visual servoing control law  are the camera velocities to minimize the positioning of $\mathbf{f}$ to $\mathbf{f}_d$.

\subsection{ViKi-HyCo Control Law}

Visual servoing controllers are widely used when working with image features, generally used in structured environments where there are markers on the stage or on the target objects of the robotic system \cite{tian2019fog}. There has been little work in outdoor environments with varied lighting and where object features are not geometric features or visual markers \cite{vivacqua2017self,lagneau2020automatic,liu2022mgbm}. The main problem with visual servoing controllers is that they require the object features in each frame to compute the controller output velocities. This loss of visual features of detected objects may be due to robot maneuvers, camera-object occlusions or the detected object being outside the FOV of the camera.  In this work, we propose a control law that combines the visual servoing controller with the robot kinematic controller, where, the kinematic controller is triggered when the visual servoing loses the object features. This avoids the problems of the visual servo controller when the visual characteristics of the object are lost and allows positioning maneuvers for the robot even if the object is not detected.

For this, we first convert the axis-camera velocities $\mathbf{V}_c$ calculated in section \ref{sec:camera_mikematics}, to ${\mathbf{V}_r}$ robot velocities through equation \eqref{eq:relation_Vc_Vr}.

\begin{equation}
\mathbf{\dot{f}} = \mathbf{L}_f \cdot \mathbf{T}_{r}^{c} \cdot \mathbf{J}_r \cdot \mathbf{{V}}_r 
\label{eq:relation_Vc_Vr}
\end{equation}

To calculate $\mathbf{{V}}_r$, the BLUE robot Jacobian $\mathbf{J}_b$ \eqref{eq:jacobian_blue} must be defined as a matrix of size 6x6. Therefore, we define a new Jacobian $\mathbf{J}_r$ in \eqref{eq:jacobian_blue_6x6}.

\begin{equation}
 \begin{bmatrix}
        \dot{X}\\ 
        \dot{Y}\\ 
        \mathbf{1}\\
    \end{bmatrix} =
 \underset{\mathbf{J}_r}{\underbrace{
    \begin{bmatrix}
         \cos\theta_r &-d\sin\theta_r &\mathbf{0} \\
         \sin\theta_r &d\cos\theta_r &\mathbf{0} \\
        \mathbf{0} & \mathbf{0} & \mathbf{1}
    \end{bmatrix} }}\cdot
    \begin{bmatrix}
        v_r\\ 
        \omega_r\\ 
        \mathbf{1}\\
    \end{bmatrix}
    \label{eq:jacobian_blue_6x6}
\end{equation}

Where $\mathbf{T}_{r}^{c}$ \eqref{eq:rTc_law_control_ViKi-HyCo} is the matrix of transformation of the robot camera. This matrix uses the rotation matrix $_{}^{r}\mathbf{R}_{c}$ and the skew-symmetric matrix of the translation matrix $_{}^{r}\mathbf{t}_{c}$ \eqref{eq:skew-symmetric matrix rtc}.

\begin{equation}
    \mathbf{T}_{r}^{c} = \begin{bmatrix}
     {_{}^{r}\mathbf{R}_c} &   [_{}^{r}\mathbf{t}_c]_\times {_{}^{r}\mathbf{R}_c}\\
    \mathbf{0} & {_{}^{r}\mathbf{R}_c}
    \end{bmatrix}
    \label{eq:rTc_law_control_ViKi-HyCo}
\end{equation}

\begin{equation}
    [_{}^{r}\mathbf{t}_c] = \begin{bmatrix}
     0&  -t_Z& t_Y\\ 
     t_Z&  0& -t_X\\ 
     -t_Y&  t_X& 0 
    \end{bmatrix}
\label{eq:skew-symmetric matrix rtc}
\end{equation}

Therefore, and with $\mathbf{\dot{f}} = (\mathbf{f}-\mathbf{f}_d)$, the visual servoing controller law for a car-like mobile robot is described in equation \eqref{eq:law_control_ViKi-HyCo}.

\begin{equation}
\mathbf{{V}}_r = -\mathbf{\lambda}(\mathbf{L}_f \cdot \mathbf{T}_{r}^{c} \cdot \mathbf{J}_r)^\dagger \cdot (\mathbf{f}-\mathbf{f}_d)
\label{eq:law_control_ViKi-HyCo}
\end{equation}

The matrix transformation $(_{}^{r}\mathbf{R}_{c},  _{}^{r}\mathbf{t}_{c})$ is selected according to the type of positioning, choosing between $(_{}^{r}\mathbf{R}_{c_{1}},  _{}^{r}\mathbf{t}_{c_{1}})$ corresponding to the front camera-robot transform for forward positioning and $(_{}^{r}\mathbf{R}_{c_{2}},  _{}^{r}\mathbf{t}_{c_{2}})$ corresponding to the rear camera-robot transform for backward positioning. 

Thus, we define the control law of the ViKi-HyCo method in equation \eqref{eq:law_control_ViKi-HyCo_global}, where the controller depends on the variable $c$, which is 1 when there is object detection and 0 when there is no object detection, so we can calculate the robot output velocities $\mathbf{{V}} = [\nu_r,\omega_r]^T$. Finally, the velocities for our robot are the linear velocity $\nu_r$ and the steering angle $\psi_r$ defined in section \ref{sec:Robot Kinematic Mode} in equation \eqref{eq:sterring angle}.

% \begin{equation}
%     \mathbf{{V}} = \begin{bmatrix}\nu_r\\\omega_r\end{bmatrix}  = \left\{\begin{matrix}
%      Ob \neq 0 ,& -\mathbf{\lambda}(\mathbf{L}_f \cdot \ \mathbf{T}_{r}^{c} \cdot \mathbf{J}_r)^\dagger \cdot (\mathbf{f}-\mathbf{f}_d)\\ 
%      Ob = 0 ,& \mathbf{J}_b^{-1} \cdot (\mathbf{k}_1 \cdot \tanh(\mathbf{\dot{h}}))
%     \end{matrix}\right.
%     \label{eq:law_control_ViKi-HyCo_global}
% \end{equation}

% \begin{equation}

% \begin{matrix}
%  \mathbf{{V}} = \begin{bmatrix}\nu_r\\\omega_r\end{bmatrix}  = c
% \times\{\mathbf{J}_b^{-1} \cdot (\mathbf{k}_1 \cdot \tanh(\mathbf{\dot{h}}))\}
%  +\\
%  \hspace{2cm} (1-c)\times\{-\mathbf{\lambda}(\mathbf{L}_f \cdot \ \mathbf{T}_{r}^{c} \cdot \mathbf{J}_r)^\dagger \cdot (\mathbf{f}-\mathbf{f}_d)\}
%     \label{eq:law_control_ViKi-HyCo_global}

% \end{matrix}
% \end{equation}

\begin{equation}
\begin{matrix}
 \mathbf{{V}} = (1-c)~\{-\mathbf{\lambda}(\mathbf{L}_f \cdot \ \mathbf{T}_{r}^{c} \cdot \mathbf{J}_r)^\dagger \cdot (\mathbf{f}-\mathbf{f}_d)\}~+ \\c ~
\{\mathbf{J}_b^{-1} \cdot (\mathbf{k}_1 \cdot \tanh(\mathbf{\dot{h}}))\}
    \label{eq:law_control_ViKi-HyCo_global}

\end{matrix}
\end{equation}

In order to reduce abrupt velocity changes with the control law in \eqref{eq:law_control_ViKi-HyCo_global}, the velocity that the robot uses is calculated with the velocities of a previous frame $(n-1)$ and a current frame $n$, as defined in \eqref{eq:law_control_filter}.

\begin{equation}
    \mathbf{{V}}_n = (1-(\mathbf{{V}}_n-\mathbf{{V}}_{n-1})) \odot \mathbf{{V}}_n
    \label{eq:law_control_filter}
\end{equation}

To enable the implemented method to calculate the robot's positioning velocity, at least one detection of the object must be made at the first instant, so that the target point for the kinematic controller can be determined.  Once this first detection has been made, the controller to be used is determined according to the number of detections of the object.

\section{Evaluation and Results}
\label{sec:experiments}
In this section we validated the robustness and accuracy of our ViKi-HyCo method by performing several experiments comparing the advantages of using a hybrid-control combining a visual servoing controller and a kinematic controller for the positioning of our research platform BLUE. We divided the section into four groups of experiments. \textit{Experiments with only visual servoing controller}, where we evaluate the problems of having only visual servoing controller on the mobile robot when it performs positioning maneuvers and loses detection of its target point. \textit{Experiments with the ViKi-HyCo method}, where we evaluated our hybrid controller approach by allowing the visual servoing controller to work in conjunction with a kinematic controller for complex positioning maneuvers of the mobile robot. \textit{Comparative Experiments}, where we evaluate the approach of continuous calculation of a desired bounding box for unknown objects in a visual servoing controller with another controller from the literature where its desired bounding box is given by the physical characteristics of a known object. Finally, in the \textit{Robot Placement Experiments}, using our Viki-Hyco hybrid controller, we detail the positioning maneuvers of the mobile robot that the hybrid-control generates and that are necessary to position the robot towards a detected object in a desired zone of the robot. For this experiment, it is necessary for the positioning of the robot maneuvers where the data input is switched from a front to a rear camera for the visual servoing controller, thus, the use of a visual object tracker is not possible.

\subsection{Experimental Setup}
We evaluated the approach ViKi-HyCo in our research platform BLUE. It has a VLP-16 LiDAR sensor, two {Intel \textregistered} {RealSense \texttrademark} D435i RGB-D cameras, and an MSI onboard computer with a 2.60 GHz 6-core processor with 16 GB of RAM running Ubuntu 18.04 and an NVIDIA GTX 1660 graphics card with 6.0 GB of video memory. In the experiments, we limited the velocities to the mechanical characteristics of the robot BLUE in the ranges of linear velocity $-0.5<\nu_r<0.5$ m/s and a steering angle $-0.44<\psi_r<0.44$ rad. We compare the positioning errors of a forward and backward positioning when using ViKi-HyCo regarding a only the visual servoing controller. Also, we analyze the results of the robot positioning task to an object, where forward and backward positioning are used together. In addition, we compare our approach of continuous calculation of a desired bounding box for unknown objects with the method MGBM-YOLO proposed in \cite{liu2022mgbm}, where these authors use a visual servoing controller with the bounding box features of an object detected by a YOLOv3 NN as well as a Jacobian image matrix, instead of using camera-object depths, to estimate the depth values with the area of the desired object and the area of the current object. For the positioning experiments, we consider the maneuvers of positioning the robot towards a detected object in a desired area of interest of the robot, where we consider this area of interest as the robot's manipulation zone for future object grasping tasks. Finally, we compare the results of position error, translation point evolution, camera-axis velocities, and calculated robot velocities of all proposed experiments. 

The experiments performed consider solved planning and autonomous global and local navigation tasks of our BLUE robot with the method presented in \cite{munoz2022openstreetmap}. Therefore, we evaluated ViKi-HyCo in an unstructured environment. Besides, we use only one object per experiment of the categories: can, plastic, glass or carton. As a matter of fact, we do not evaluate our controller with several objects in scene, because the ViKi-HyCo method actually does not include an object tracker. We neither classify the objects, because we consider them as a target point regardless of the object class. The performance of the NN in detecting objects is not evaluated. This has already been analyzed in \cite{tornero2022detection} and mentioned in the Section \ref{sec:Object Localization}. For this reason, our scenarios are not challenging for object detection by YOLOv5. However, experimentation shows that a NN may eventually lose object detection in a simple scenario. In addition, our hybrid-control system does not have a visual object tracker, because this would not be possible when performing the complete positioning maneuver where it is necessary to change the image input source of the visual servoing controller.

The ground truth was established as the physical measurements taken from the robot's base point $\mathbf{T^r}$ to the object location point. We consider for forward visual servoing positioning the RGB image from the front camera and the interpolated LiDAR to 112 virtual layers (\ref{sec:lidar_camera_fusion}), and for backward positioning we use the depth image and the RGB image from the rear camera. We set the two cameras at  15 fps (frames per second) with 1280x720 resolution ($W_{img}=1280, \, H_{img} = 720$). The gains in \eqref{eq:law_control_ViKi-HyCo_global} were established prior experimentation, where $\lambda= [0.85, 0.3, 1.0, 1.0, 1.0]$ to the front camera and  $\lambda=  [0.85, 1.05, 1.0, 1.0, 1.0]$ to the rear camera. The gains of the forward kinematic controller is $\mathbf{k}_1 = [2.0, 1.0]$ and the backward kinematic controller is $\mathbf{k}_1 = [1.0, 2.0] $. Additionally, according to the metrics for determining the error of a positioning system presented in \cite{munoz2021criterios}, we evaluated the results of our method using the Mean Squared Error (\text{MSE}). Finally, each iteration shown in the result plots represents the run-time of the method and is equivalent to an average of 44 ms each iteration.

%% Figure experiment 1

\begin{figure} [t]
    \centering
  \subfloat[Camera path\label{fig:exp1_camera_path}]{%
        \includegraphics[width=0.7\linewidth]{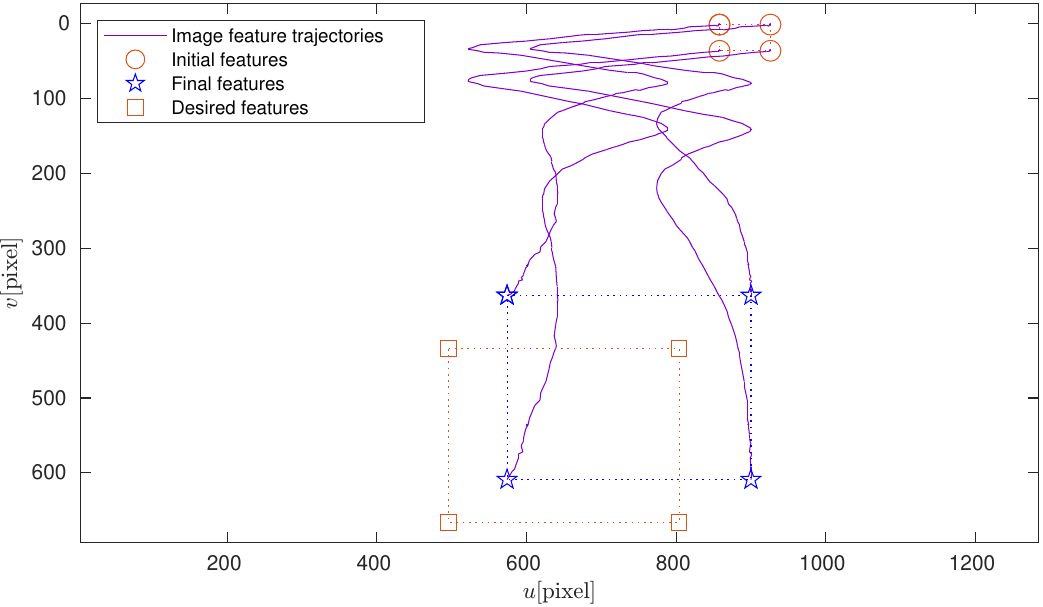}}
  \\
    \subfloat[Camera velocity \label{fig:exp1_velocitycamera}]{%
    \includegraphics[width=0.75\linewidth]{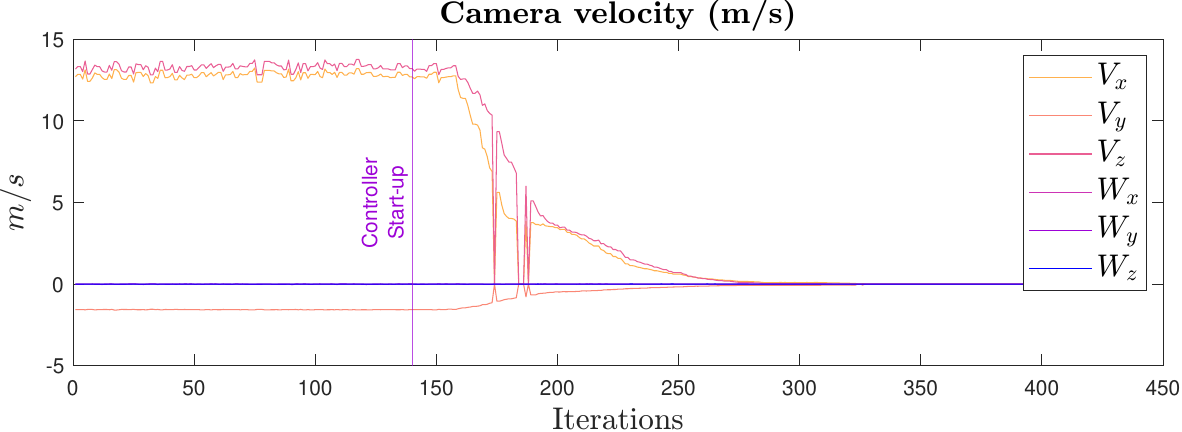}}
     \\  
  \subfloat[Robot path\label{fig:exp1_robot_path}]{%
       \includegraphics[width=0.75\linewidth]{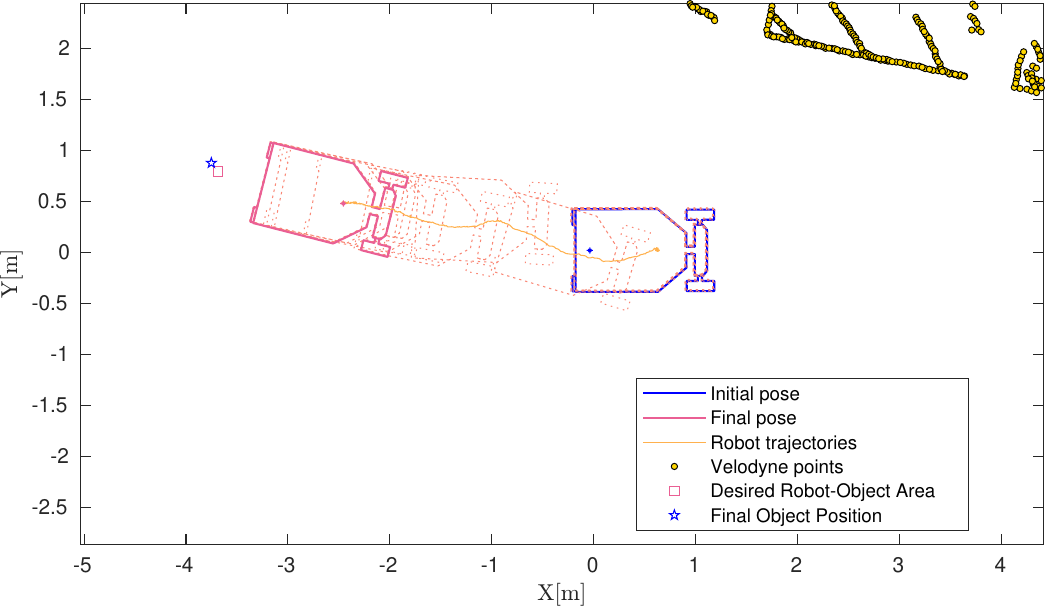}}
    \\
      \subfloat[Lineal velocity and steering angle\label{fig:exp1_velocityrobot}]{%
        \includegraphics[height=2.5cm]{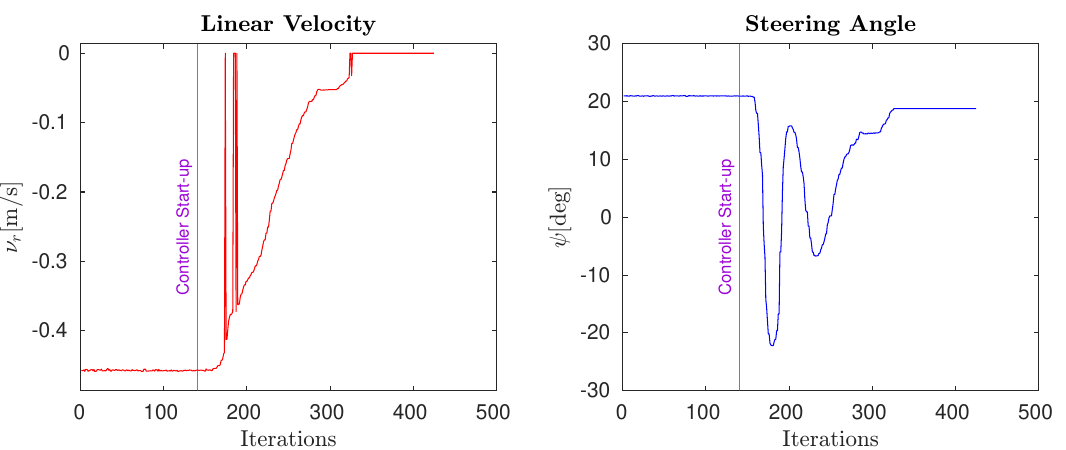}}
    \\
    \subfloat[X and Y robot translation\label{fig:exp1_xyrobot}]{%
        \includegraphics[height=2.5cm]{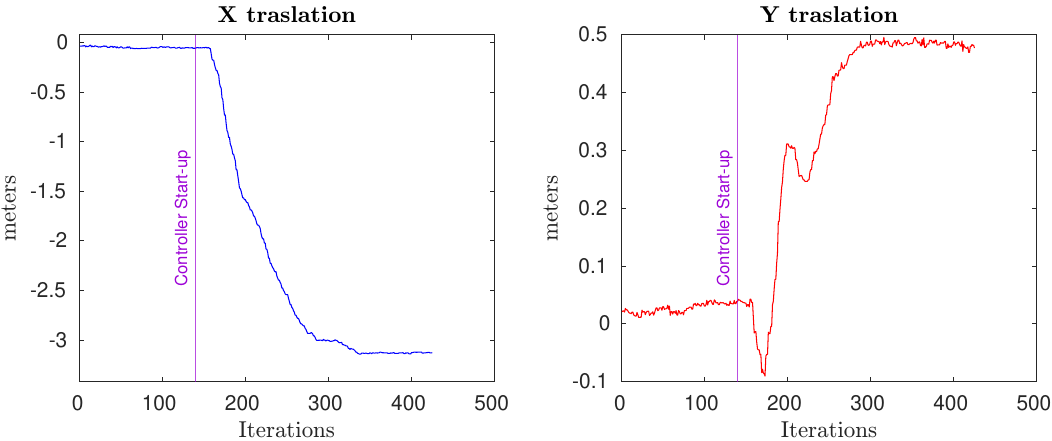}}
  \caption{Backward positioning of BLUE robot towards an object detected using only the visual servoing controller.}
  \label{fig:exp1} 
\end{figure}

\begin{figure} [t]%% Figure experiment 2
    \centering
      \subfloat[Camera path\label{fig:exp2_camera_path}]{%
        \includegraphics[width=0.7\linewidth]{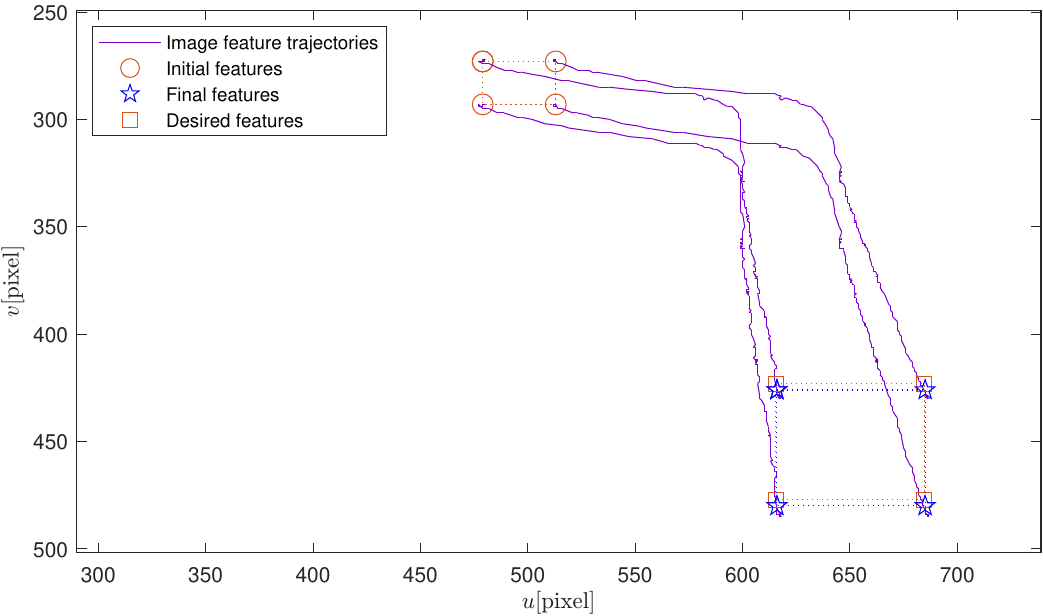}}
        \\
    \subfloat[Camera velocity \label{fig:exp2_velocitycamera}]{%
    \includegraphics[width=0.75\linewidth]{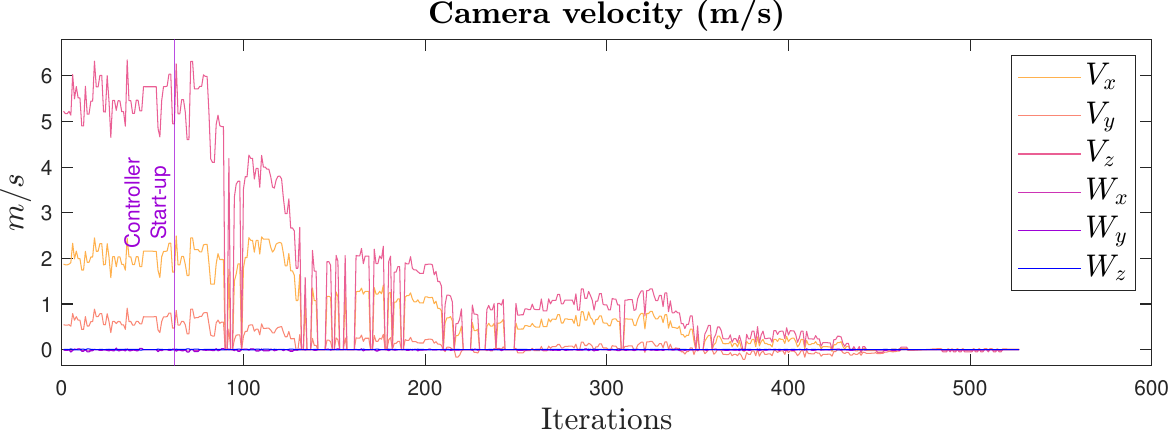}}
     \\
    \subfloat[Robot path\label{fig:exp2_robot_path}]{%
       \includegraphics[width=0.75\linewidth]{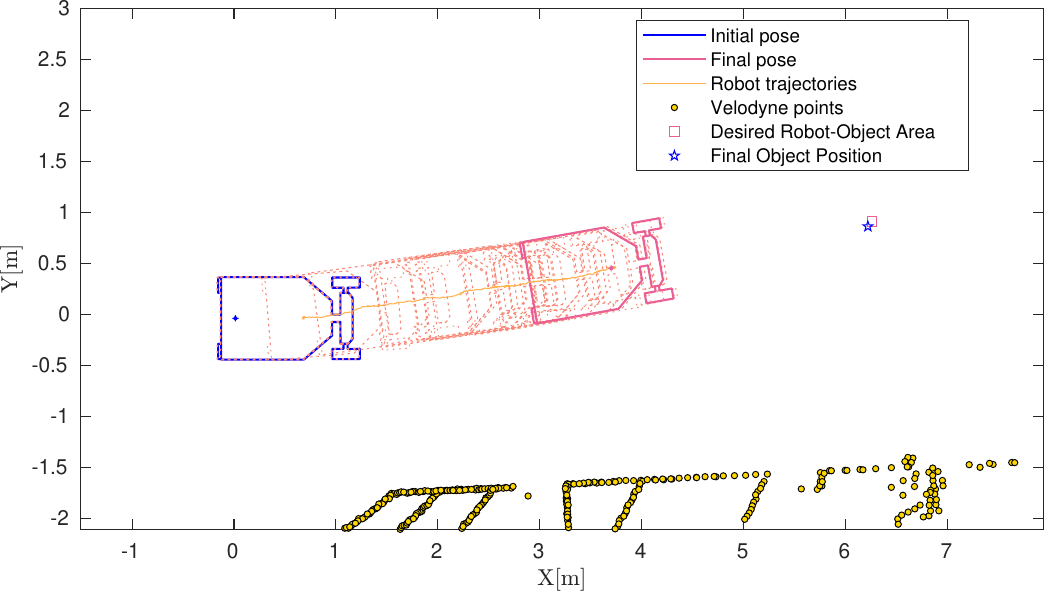}}
    \\    
    \subfloat[Lineal velocity and steering angle\label{fig:exp2_velocityrobot}]{%
        \includegraphics[height=2.5cm]{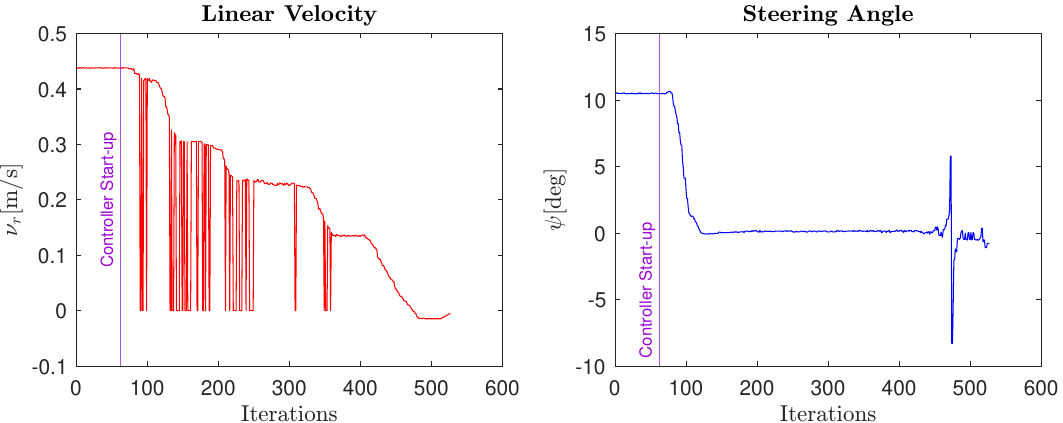}}
    \\
    \subfloat[X and Y robot translation\label{fig:exp2_xyrobot}]{%
        \includegraphics[height=2.5cm]{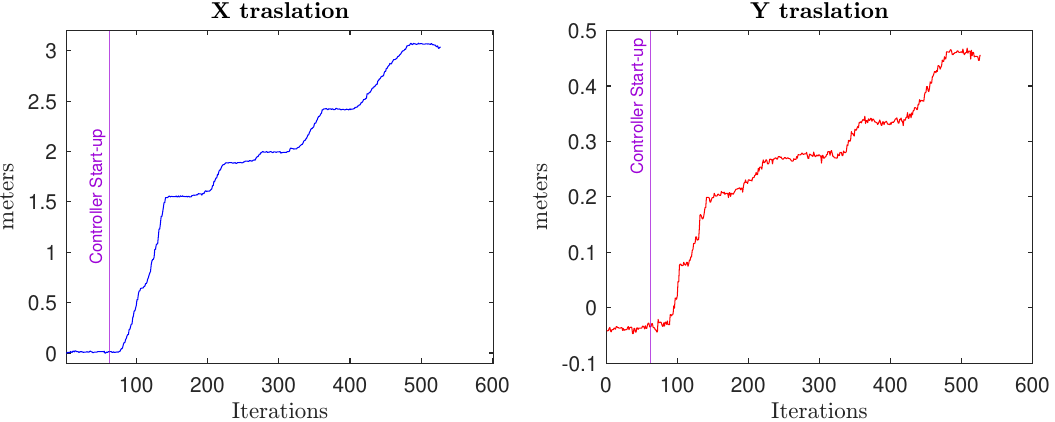}}
  \caption{Forward positioning of BLUE robot towards an object detected using only visual servoing controller. Missed object detections cause the robot to stop abruptly.}
  \label{fig:exp2_forward} 
\end{figure}

\subsection{Experiments with the Classic Visual Servoing Controller} 
\label{sec:exp1}
We performed 40 experiments with four different objects in two outdoor environments. 20 experiments with forward positioning and 20 with backward positioning, applying in both cases the visual servoing controller, calculating the robot velocity $V_r$ by means of the equation \eqref{eq:law_control_ViKi-HyCo}.

In Fig. \ref{fig:exp1_robot_path} shows an experiment of the positioning path of the robot towards the object with the visual servoing controller backward. When using a visual servoing controller we have a positioning error of $0.0645$ m in X-axis and $0.0832$ m in Y-axis (see Fig. \ref{fig:exp1_xyrobot}). In the image plane (Fig. \ref{fig:exp1_camera_path}), the camera motion fails to converge completely to the desired characteristics, this is because the YOLOv5 NN detector loses object detections in some frames, consequently the linear and angular velocities are 0 m/s, stopping the robot's motion. The velocities change to 0 m/s because no object is detected in the image plane. As a result, there are no features $\mathbf{f}$ for the visual servo controller, and through the desired feature $\mathbf{f}_d$ \eqref{eq:desired_feature} and the visual servo control law \eqref{eq:law_control_ViKi-HyCo}, the velocities calculated by the controller become 0 m/s (see Fig.\ref{fig:exp1_velocitycamera}). The above is shown in Fig. \ref{fig:exp1_velocityrobot}, where the calculated linear velocity is 0 m/s in some iterations of the experiment (near iteration number 200).

Fig. \ref{fig:exp2_forward} shows the forward positioning of the robot with the visual servoing controller. In this experiment, it is seen in more detail that the calculated camera velocities (see Fig. \ref{fig:exp2_velocitycamera}) and linear velocities of the mobile robot (see Fig. \ref{fig:exp2_velocityrobot}) are equal to 0 m/s. Likewise, as in the backward positioning with the visual servoing controller, the abrupt velocity changes are caused when the object detections are missed. Despite the fact that the robot converges to the desired position in the image plane (see Fig. \ref{fig:exp2_camera_path}) and approaches the desired position in the navigation plane (see Fig. \ref{fig:exp2_robot_path}), the missed object detections cause the robot stopped, generating a discontinuous trajectory, as seen in the trajectory in X and Y (see Fig. \ref{fig:exp2_xyrobot}). This experiment has an error of $0.0416$ in X-axis and $0.0466$ in Y-axis.

\begin{figure} %% Figure experiment 1-2 results
    \centering
  \subfloat[Initial front camera image\label{fig:initial_front_image}]{%
       \includegraphics[width=0.48\linewidth]{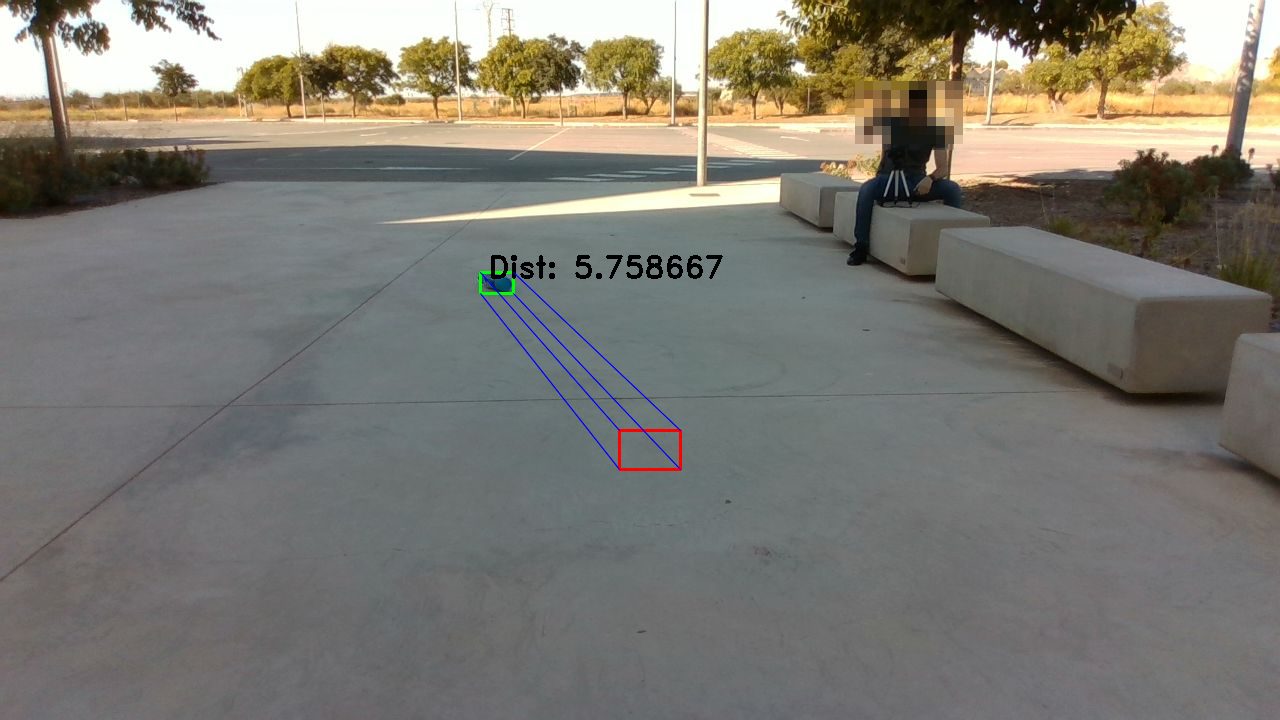}}
    \hfill
  \subfloat[Final front camera image\label{fig:final_front_image}]{%
        \includegraphics[width=0.48\linewidth]{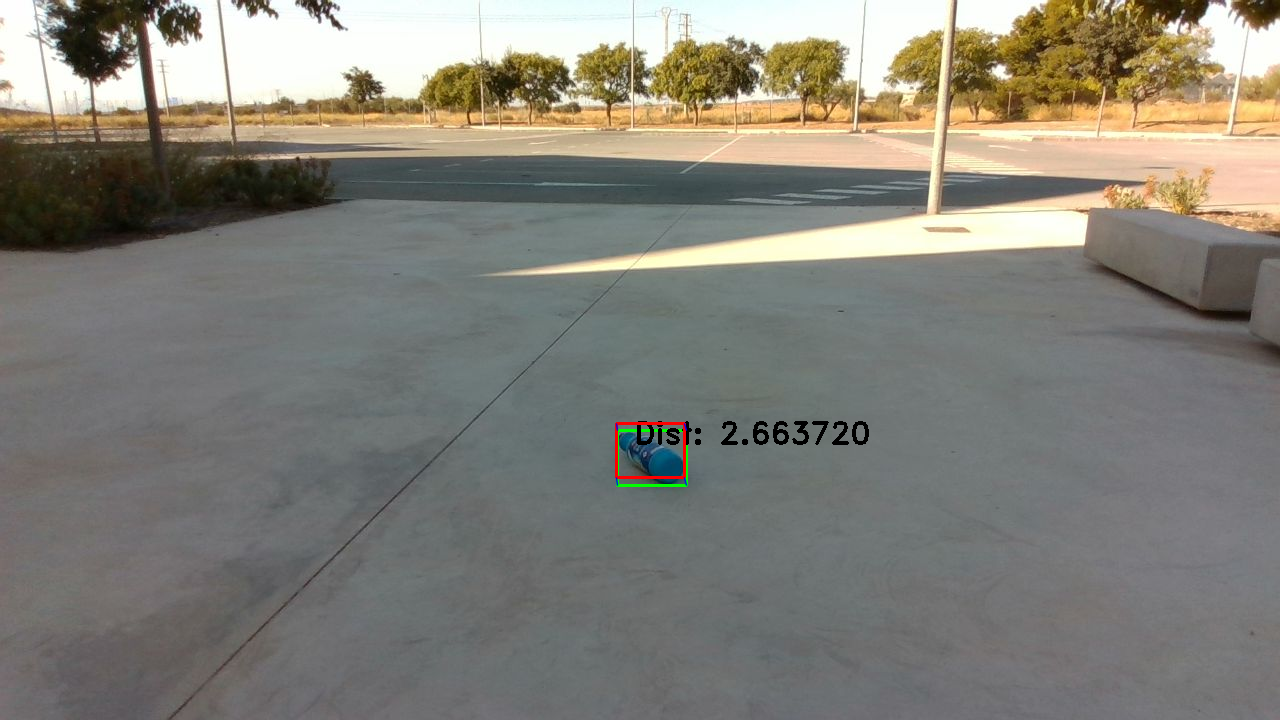}}
\\
  \subfloat[Final rear camera image\label{fig:initial_rear_image}]{%
        \includegraphics[width=0.48\linewidth]{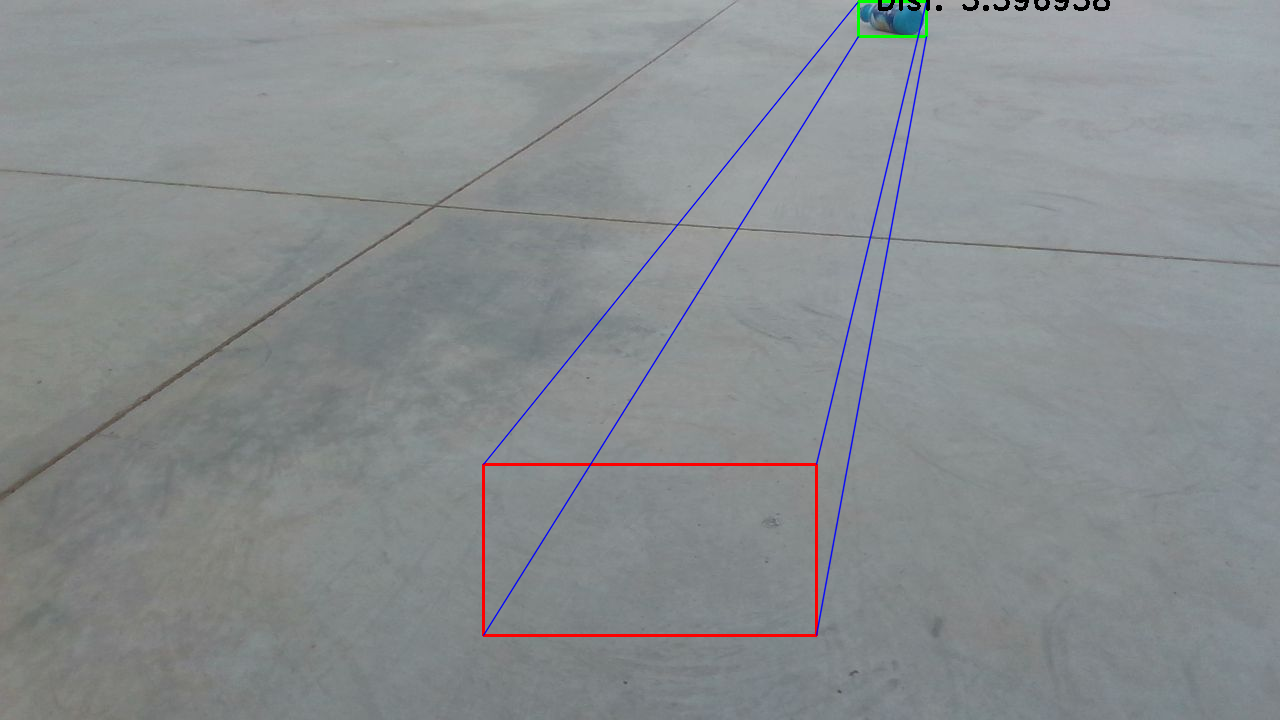}}
    \hfill
  \subfloat[Final rear camera image\label{fig:final_rear_image}]{%
        \includegraphics[width=0.48\linewidth]{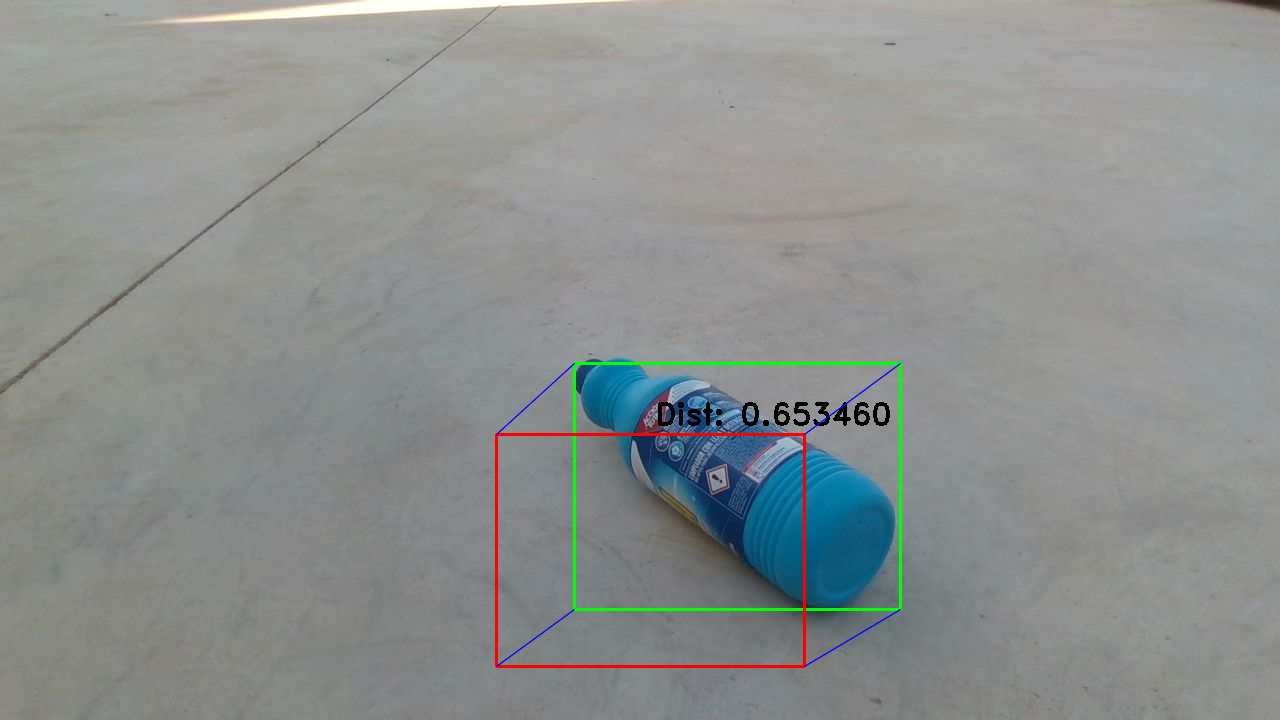}}
  \caption{Initial and final images from the cameras when using a classic visual servoing controller in the forward and backward positioning of the BLUE robot.}
  \label{fig:exp1-2_results} 
\end{figure}

The initial and final image of the front and rear cameras when using only the visual servoing controller are shown in Fig. \ref{fig:exp1-2_results}. 

\begin{figure} %% Figure experiment 2
    \centering
      \subfloat[Camera path\label{fig:error_vs_camera_path}]{%
        \includegraphics[width=0.8\linewidth]{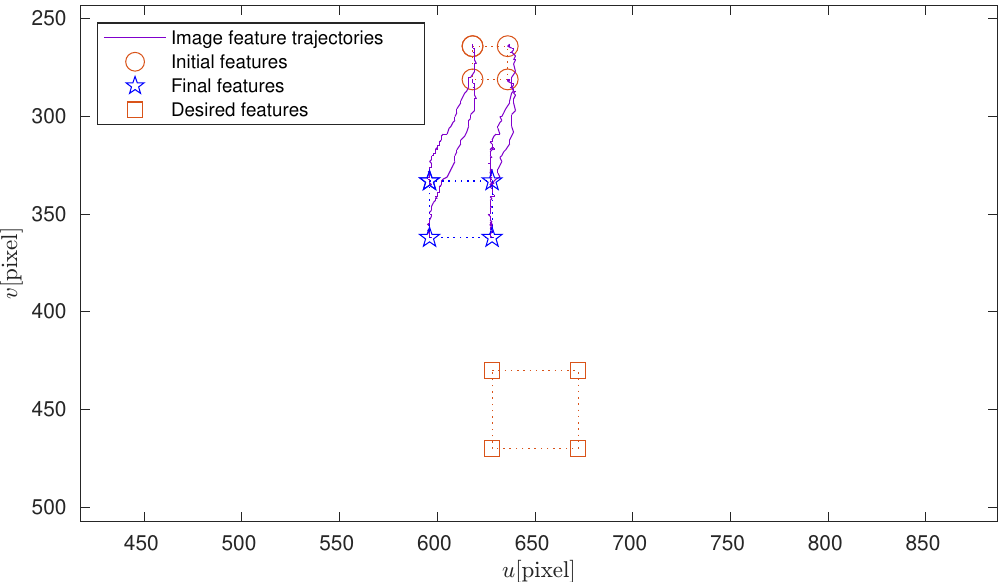}}
     \\
    \subfloat[Robot path\label{fig:error_vs_robot_path}]{%
       \includegraphics[width=0.8\linewidth]{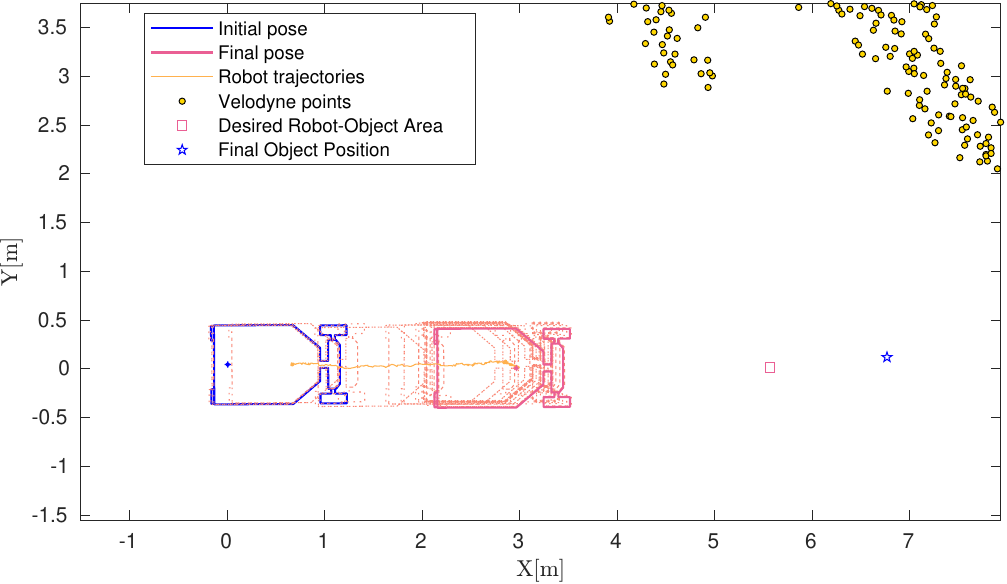}}
    \\
    \subfloat[Lineal velocity and steering angle\label{fig:error_vs_velocityrobot}]{%
        \includegraphics[height=3cm]{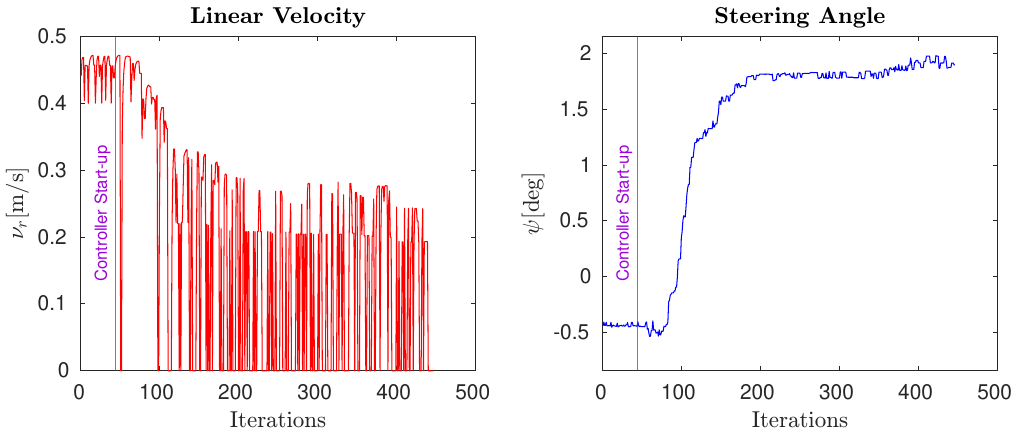}}
   % \\
   %\subfloat[X, Y robot translation\label{fig:error_vs_xyrobot}]{%
    %    \includegraphics[height=3.5cm]{figures/Results/visual_servoing_forward_error/xy_traslation.pdf}}
  \caption{Errors on forward positioning of BLUE robot towards an object detected by YOLOv5 using the visual servoing controller. Missed object detections cause the robot to stop abruptly.}
  \label{fig:error_vs_forward} 
\end{figure}

During several experiments performed in forward positioning with only the visual servoing controller, object detections by YOLOv5 were lost and the robot stopped before the desired position. When the robot stops, the desired and actual bounding boxes do not converge, as shown in Fig. \ref{fig:error_vs_camera_path}. These lost detections are caused by the robot's positioning movements, causing the object to move in the image plane. The robot does not move to the desired position due to the abrupt changes of the linear velocities that stop it (see Fig. \ref{fig:error_vs_velocityrobot}). Thus there are positioning errors of $1.202$ m in the X-axis and $0.105$ in the Y-axis (see Fig. \ref{fig:error_vs_robot_path}). 

\subsection{Experiments with ViKi-HyCo Method} 
\label{sec:exp2}
\begin{figure}
    \centering
   \subfloat[Camera path\label{fig:exp3_camera_path}]{%
        \includegraphics[width=0.7\linewidth]{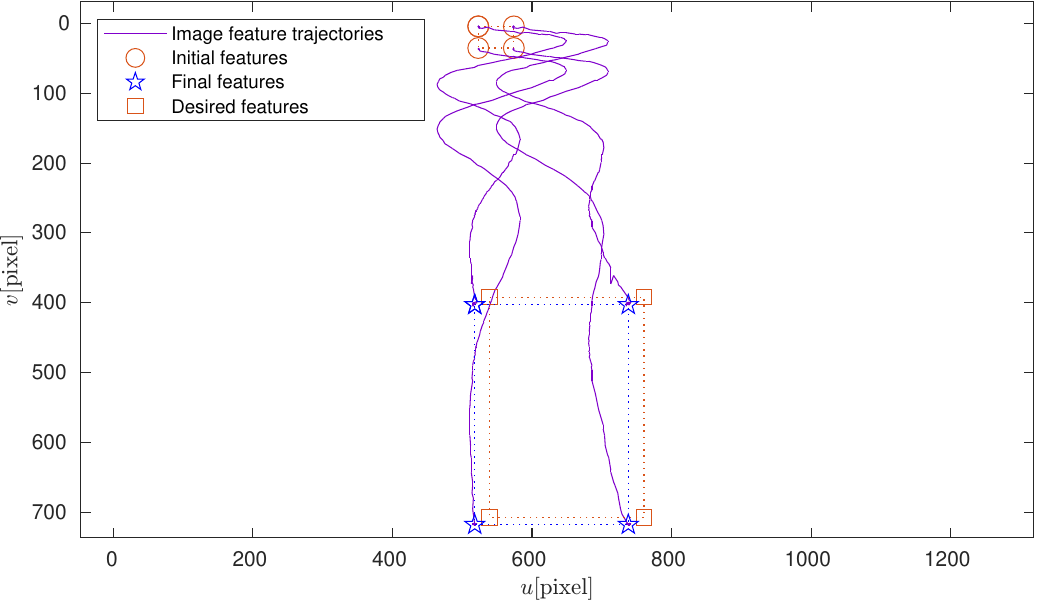}}
    \\
    \subfloat[Camera velocity \label{fig:exp3_velocitycamera}]{%
    \includegraphics[width=0.75\linewidth]{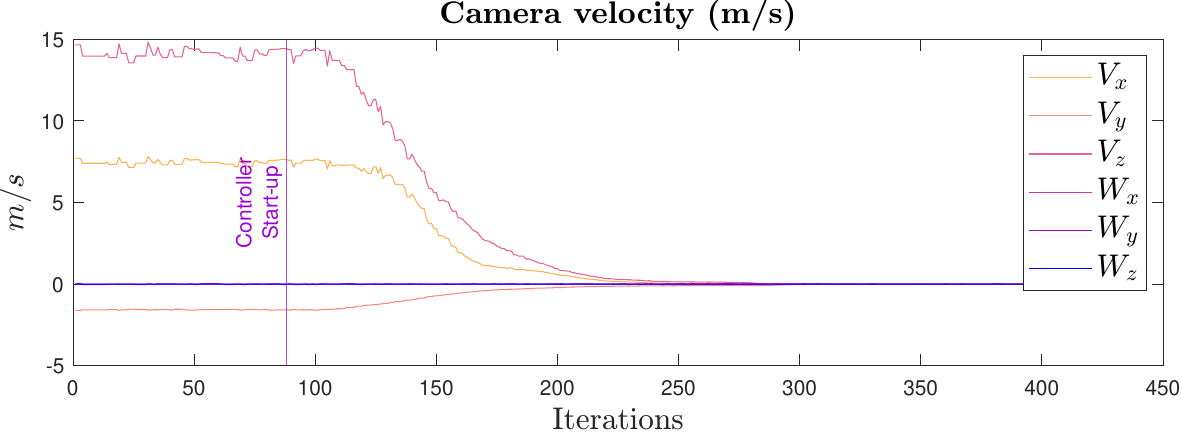}}
     \\    
    \subfloat[Robot path\label{fig:exp3_robot_path}]{%
       \includegraphics[width=0.75\linewidth]{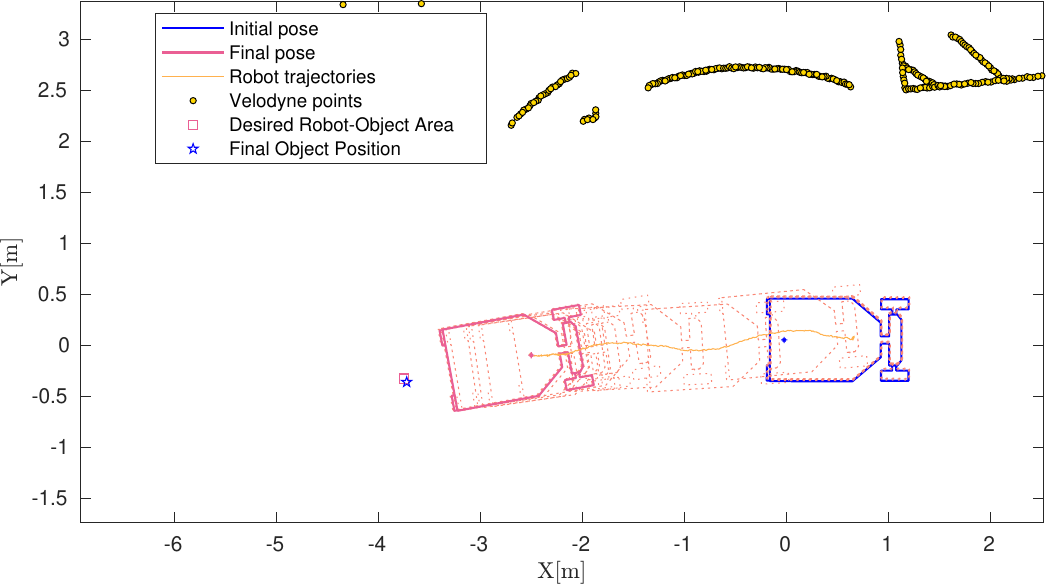}}
    \\
  \subfloat[Lineal velocity and steering angle\label{fig:exp3_velocityrobot}]{%
        \includegraphics[height=2.5cm]{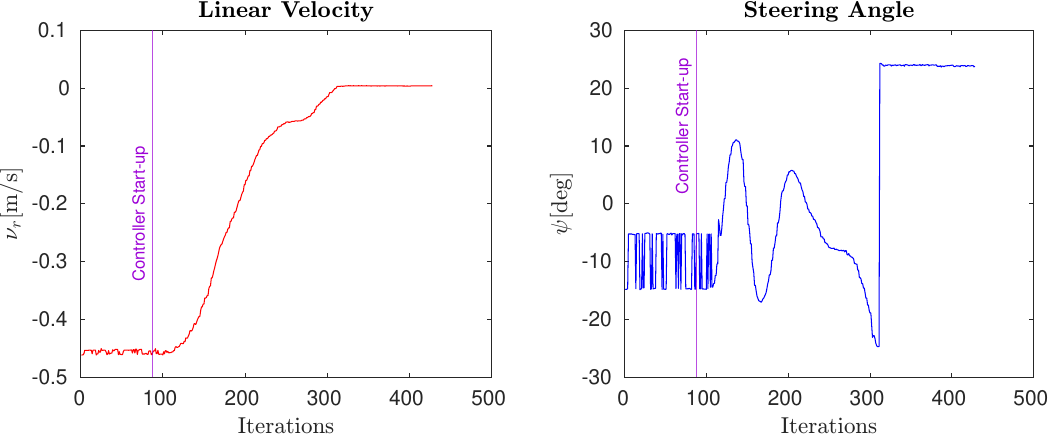}}
    \\
    \subfloat[X and Y robot translation\label{fig:exp3_xyrobot}]{%
        \includegraphics[height=2.5cm]{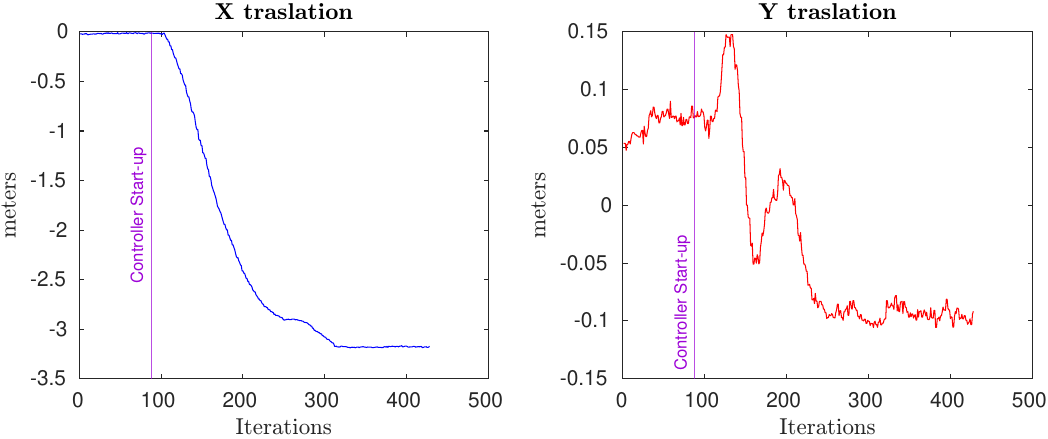}}
  \caption{Backward positioning of BLUE robot towards an object detected by YOLOv5 using the ViKi-HyCo method.}
  \label{fig:exp3} 
\end{figure}

Like in the previous experimentation, we performed 40 tests of our ViKi-HyCo algorithm, which are: 20 for forward positioning and 20 for backward positioning with four different objects. As shown in Fig. \ref{fig:exp3_camera_path}, our ViKi-HyCo method approaches the desired bounding box to the current one in the image plane in backward positioning. Moreover, unlike the visual servoing controller (see Section. \ref{sec:exp1}), ViKi-HyCo has no discontinuities when it misses the object detections. Since, immediately after missing a detection, the kinematic controller of the robot is activated. In this way, although object detections are lost, the hybrid controller allows the robot to perform smooth and uninterrupted maneuvers. Consequently, as shown in Fig. \ref{fig:exp3_velocitycamera} and Fig. \ref{fig:exp3_velocityrobot}, the camera and mobile robot velocities does not decrease to 0 m/s and is continuous with lower perturbations. Thus, the Fig. \ref{fig:exp3_robot_path} shows the robot's path of backward positioning towards the desired object with positioning errors of $0.0271$ m in the X-axis and $0.0336$ m in the Y-axis. Additionally, Fig. \ref{fig:exp3_xyrobot} shows that the trajectory has no discontinuities.
%% Figure experiment 3

\begin{figure}[!ht] 
    \centering
    \subfloat[Camera path\label{fig:exp4_camera_path}]{%
        \includegraphics[width=0.7\linewidth]{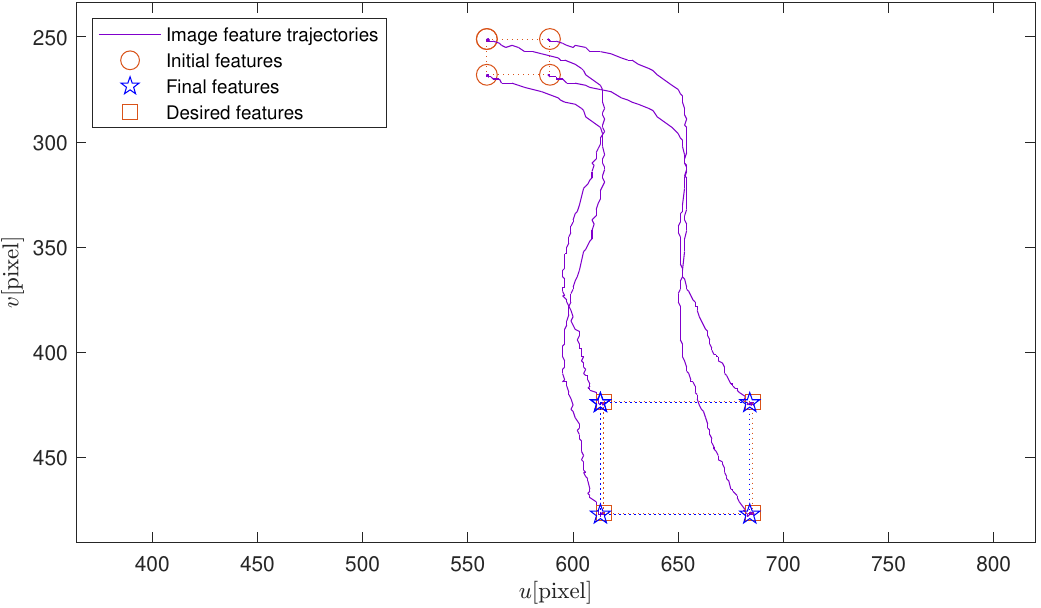}}
    \\
    \subfloat[Camera velocity \label{fig:exp4_velocitycamera}]{%
    \includegraphics[width=0.75\linewidth]{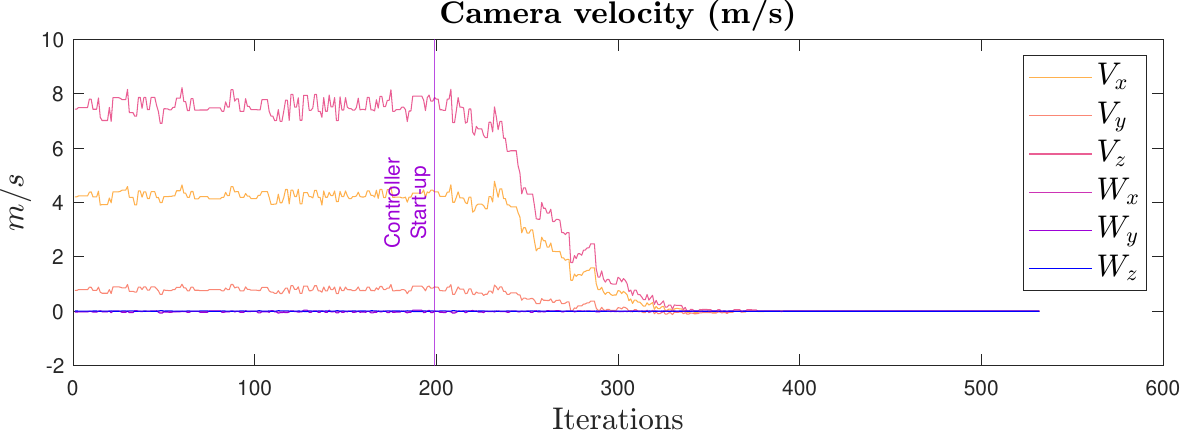}}
     \\    
  \subfloat[Robot path\label{fig:exp4_robot_path}]{%
       \includegraphics[width=0.75\linewidth]{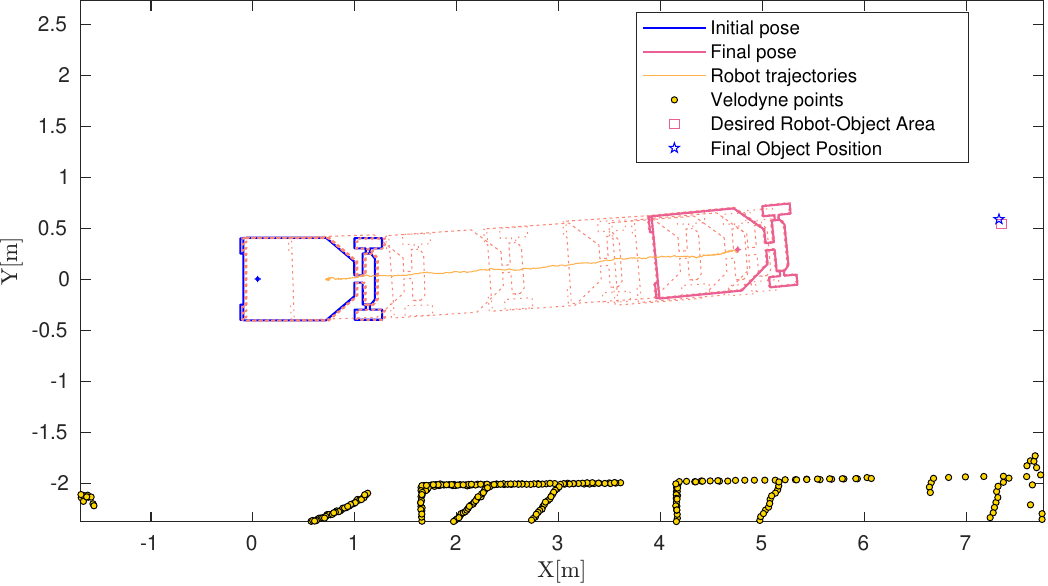}}
    \\
  \subfloat[Lineal velocity and steering angle\label{fig:exp4_velocityrobot}]{%
        \includegraphics[height=2.5cm]{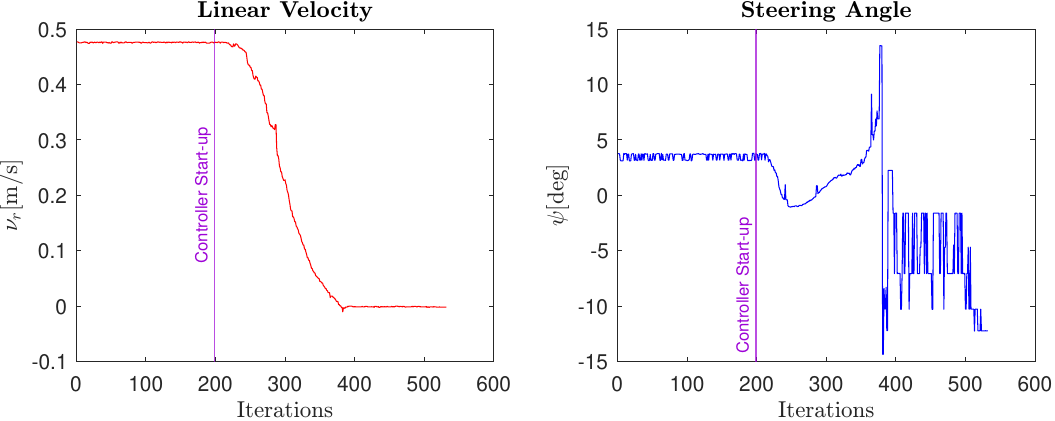}}
    \\
     \subfloat[X and Y robot translation\label{fig:exp4_xyrobot}]{%
        \includegraphics[height=2.5cm]{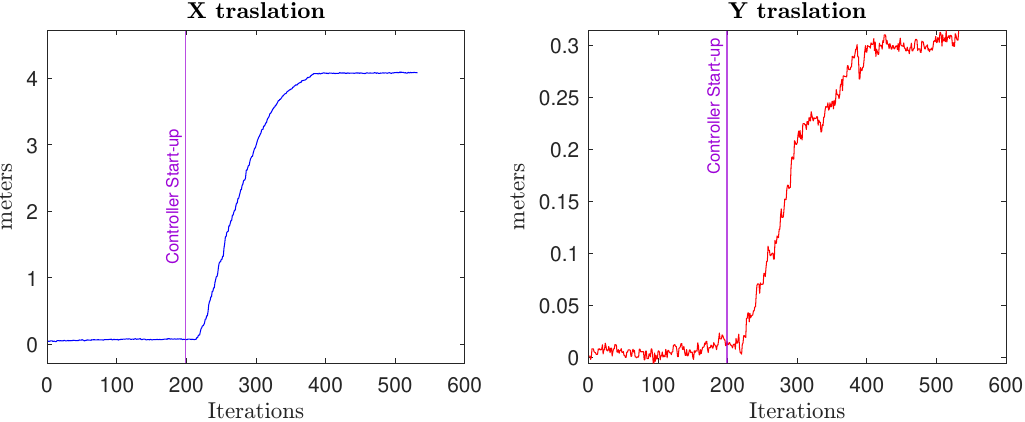}}
  \caption{Forward positioning of BLUE robot towards an object detected by YOLOv5 using the ViKi-HyCo method.}
  \label{fig:exp4} 
\end{figure}

Also, in forward positioning with the ViKi-HyCo algorithm, Fig. \ref{fig:exp4_camera_path} shows that the controller converges to the desired bounding box in the plane image, and the Fig \ref{fig:exp4_robot_path} shows that the robot moves to the desired point with positioning errors of $0.0256$ m in the X-axis and $0.0469$ m in the Y-axis. Similarly to the previous experiment, the camera and mobile robot velocities does not decrease to 0 m/s (see Fig. \ref{fig:exp4_velocitycamera} and Fig. \ref{fig:exp4_velocityrobot}). The X and Y translations (see Fig.\ref{fig:exp4_xyrobot}) shows that the robot had a smooth positioning without stopping. 

%% Figure experiment 4

\begin{figure} [!ht]%% Figure experiment 1-2 results
    \centering
  \subfloat[Initial front camera image\label{fig:exp_34_initial_front_image}]{%
       \includegraphics[width=0.48\linewidth]{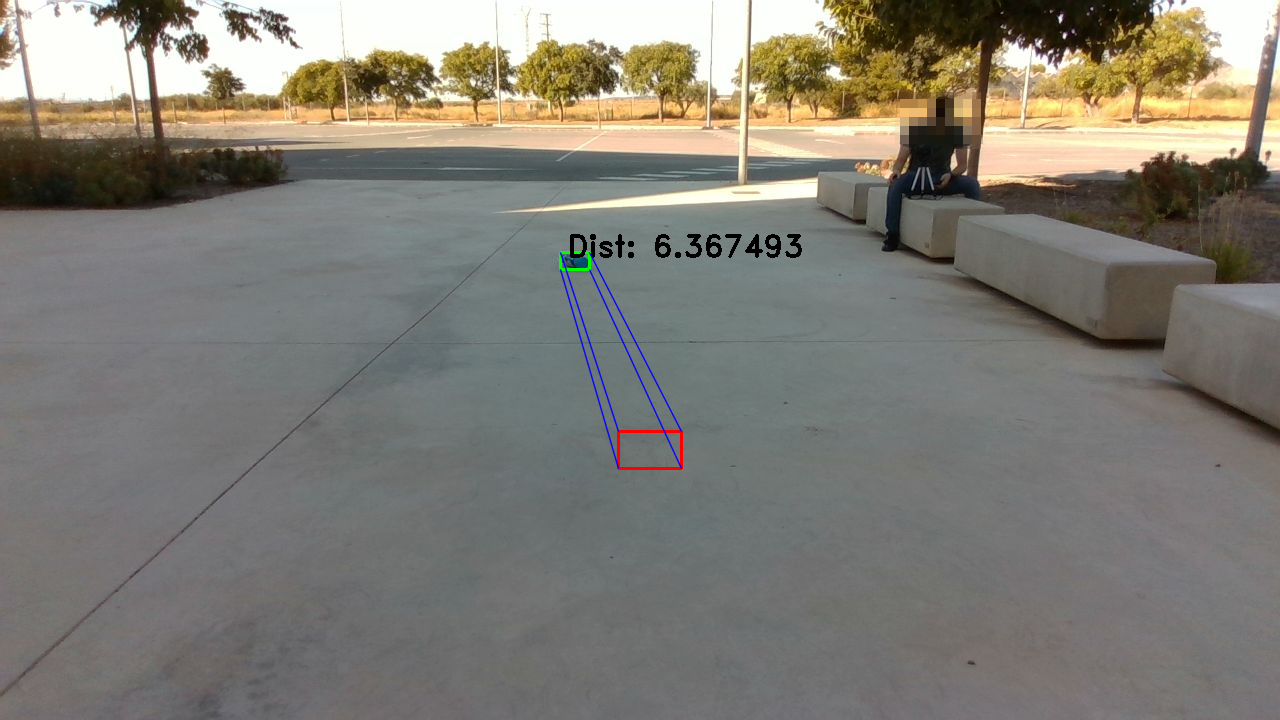}}
    \hfill
  \subfloat[Final front camera image\label{fig:exp_34_final_front_image}]{%
        \includegraphics[width=0.48\linewidth]{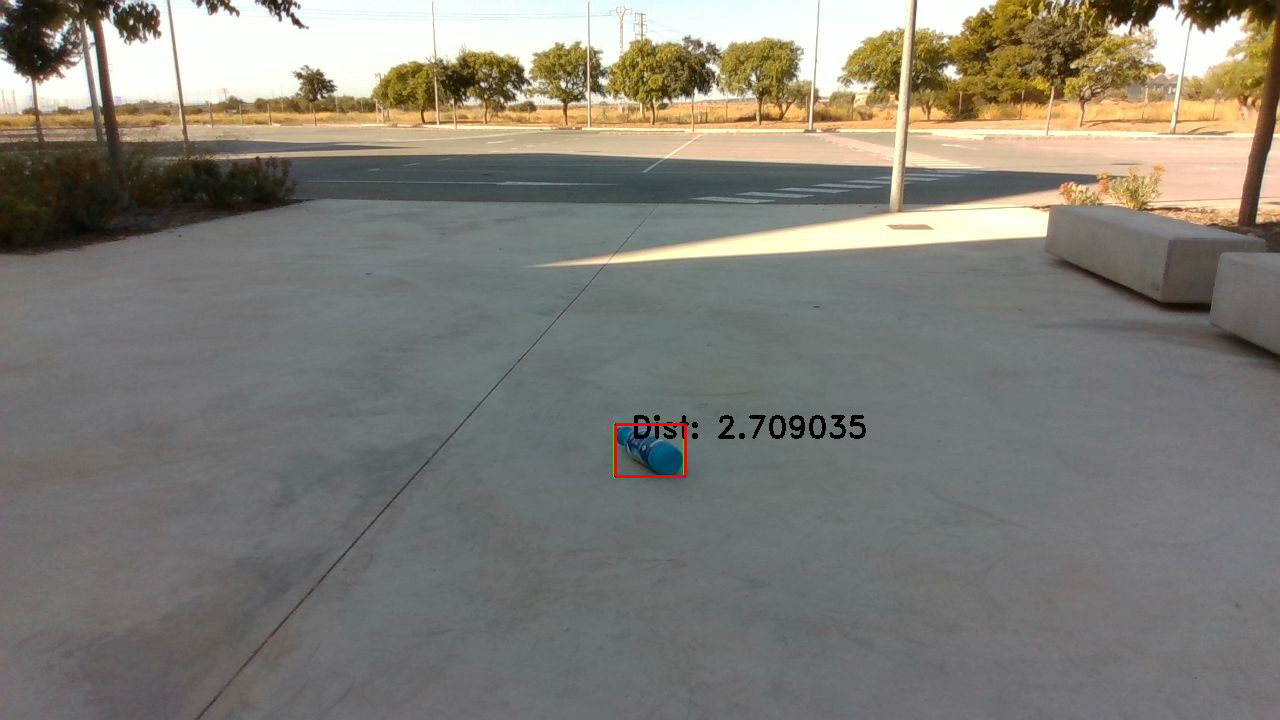}}
\\
  \subfloat[Final rear camera image\label{fig:exp_34_initial_rear_image}]{%
        \includegraphics[width=0.48\linewidth]{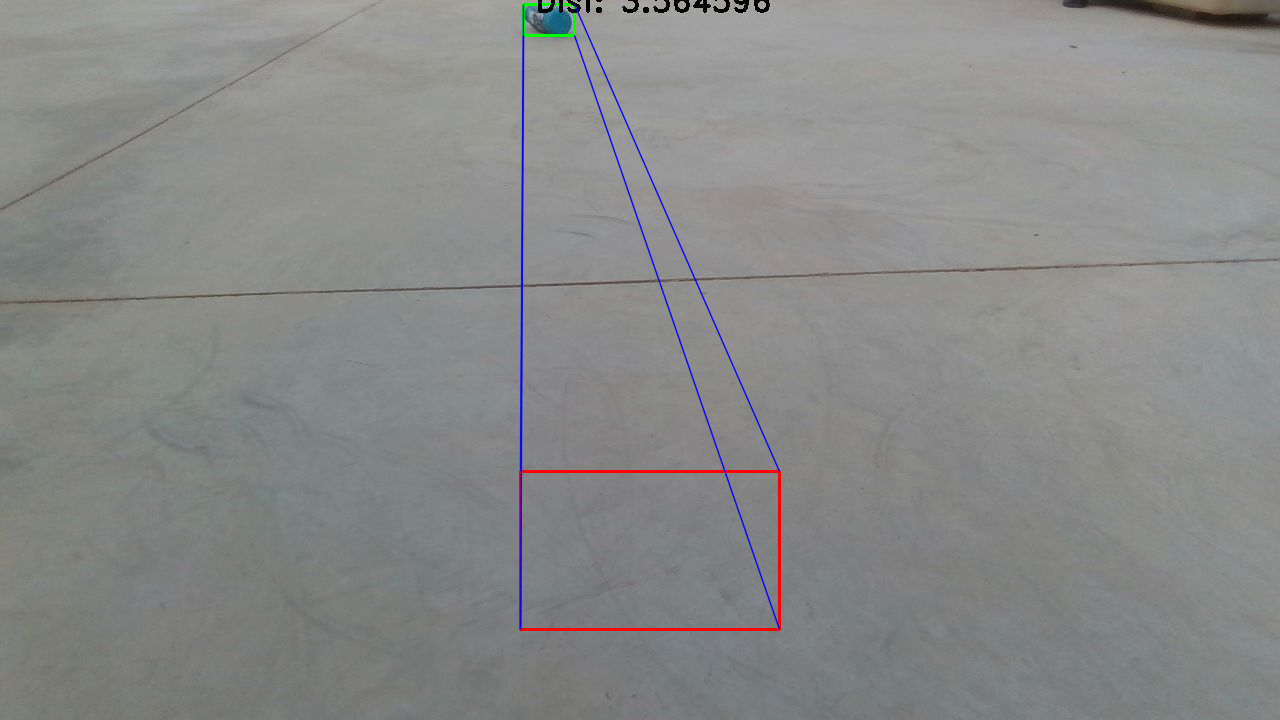}}
    \hfill
  \subfloat[Final rear camera image\label{fig:exp_34_final_rear_image}]{%
        \includegraphics[width=0.48\linewidth]{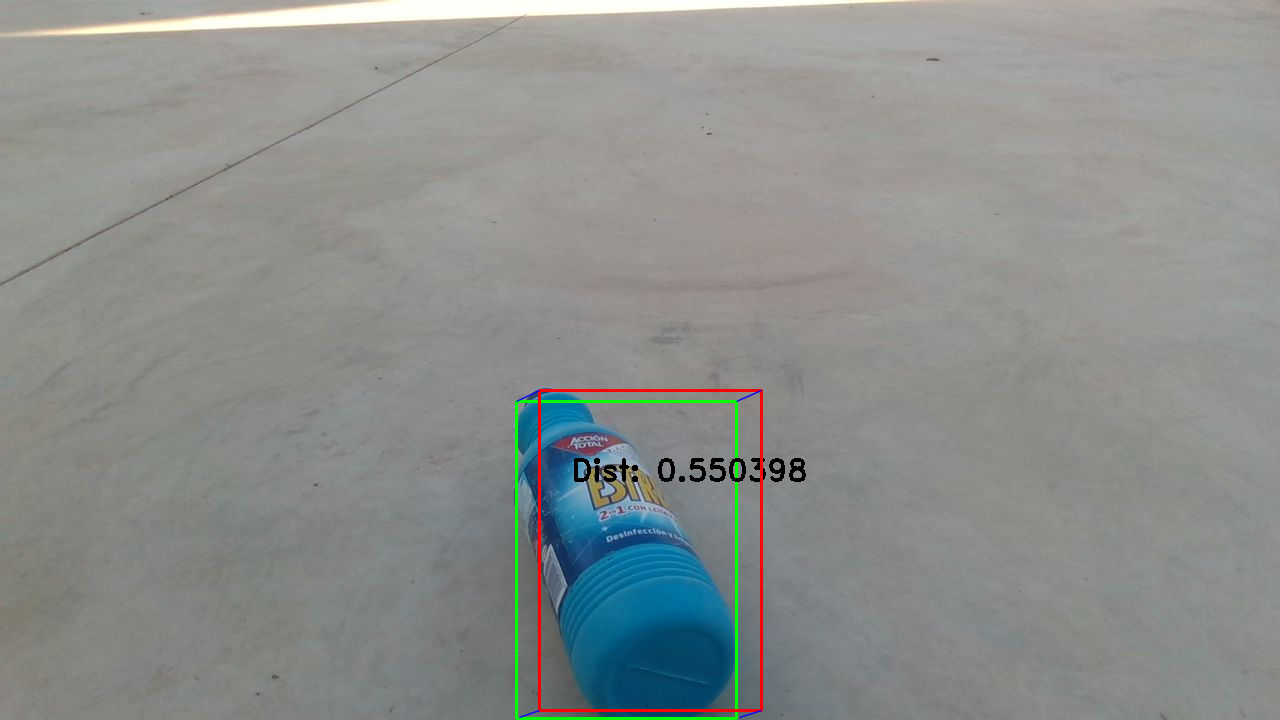}}
  \caption{Initial and final images from the cameras when using ViKi-HyCo in the forward and backward positioning of the BLUE robot.}
  \label{fig:exp3-4_results} 
\end{figure}

The initial and final image of the front and rear cameras with The ViKi-HyCo method are shown in Fig. \ref{fig:exp3-4_results}. 

\subsection{Comparative Experiments}
\label{sec:exp3}
The MGBM-YOLO method proposed in \cite{liu2022mgbm}, uses a visual servoing controller with YOLOv3 detections and modifies the Jacobian's image. Instead of using the camera-object depth, it works with the desired and current bounding box of the detected object. MGBM-YOLO method has the final bounding box because it knows the dimensions and shape of the object  projected on the image plane, in this case springs.  In this way, in their application, the velocities calculated by the controller converge and reduce the positioning errors. It is worth mentioning that, although MGBM-YOLO is developed for spring grasping with an image-based robotic arm, it uses a methodology similar to ours. The visual servoing controller law is similar to ours, since it uses the vertices of the bounding box of the YOLOv3 NN. The main difference in our proposal is that in the visual servoing controller, we apply the desired bounding box update to facilitate the match the final shape. That is why, in this experimentation, we compare the behavior of the visual servoing controller proposed in MGBM-YOLO and part of our proposal which is the updating of the bounding box in the visual servoing controller.  
To perform this experiment, we generated a desired object's bounding box with the dimensions of a bottle as if the robot's camera placed it in its final position. The disadvantage of the MGBM-YOLO method is that the shape and dimensions of the detected object must be known in advance to generate a desired bounding box. Therefore, if the shape of the actual bounding box of the object does not match the final shape, the velocities of the controller will not converge. 

\begin{figure*}
    \centering
    \subfloat[Camera path\label{fig:exp5_camera_path}]{%
        \includegraphics[width=0.25\textwidth]{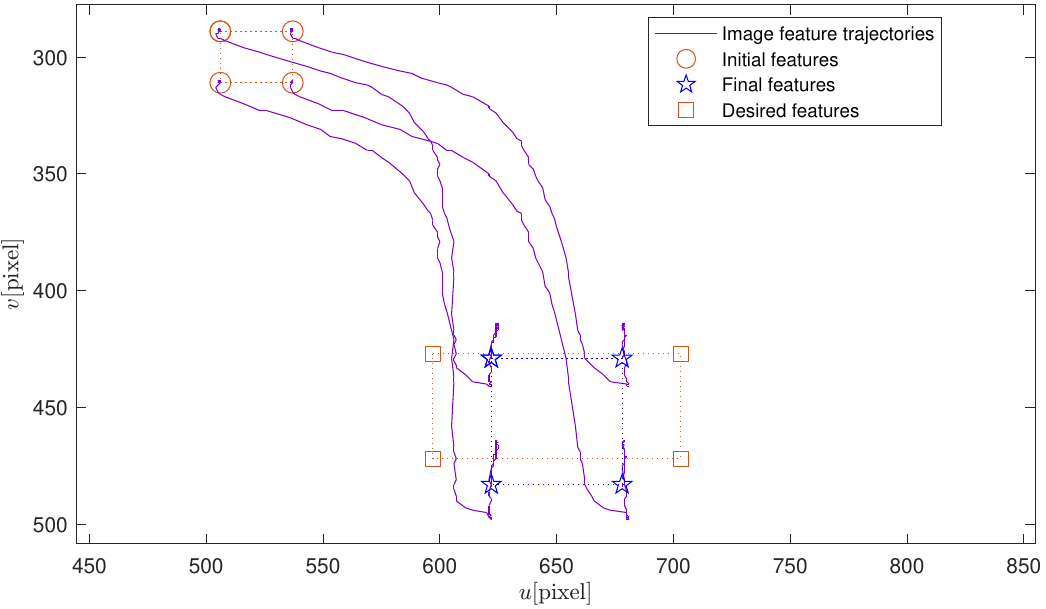}}
    \hfill 
    \subfloat[Robot path\label{fig:exp5_robot_path}]{%
       \includegraphics[width=0.25\textwidth]{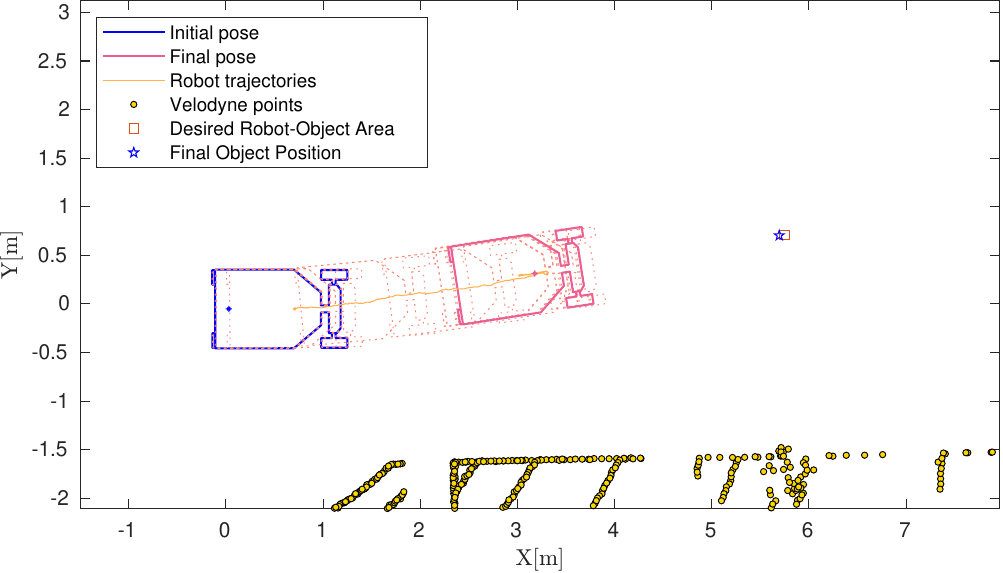}}
    \hfill 
    \subfloat[Camera path\label{fig:exp5_camera_path_viki}]{%
        \includegraphics[width=0.25\textwidth]{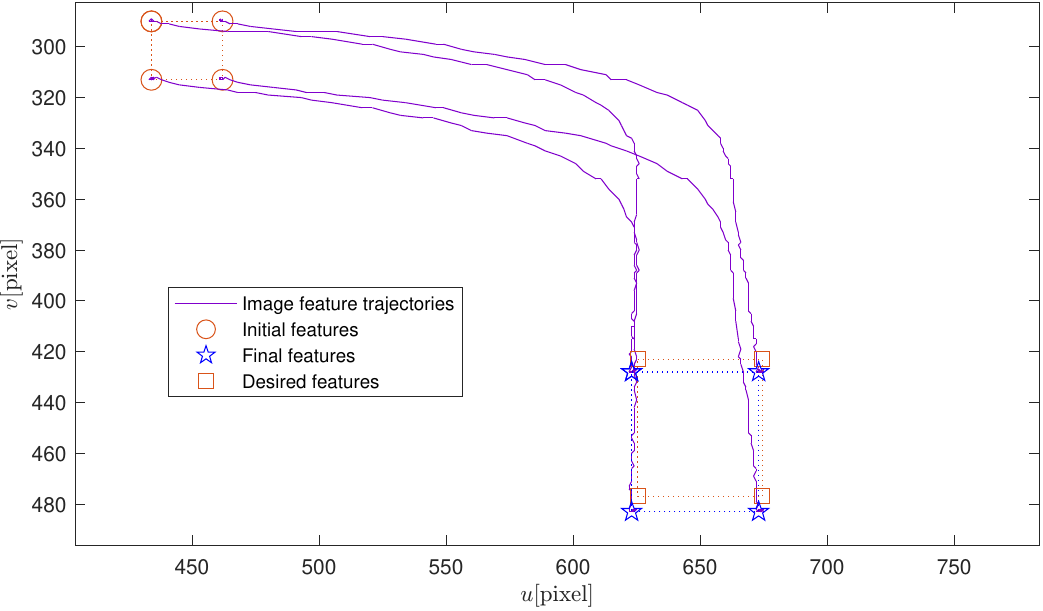}}
    \hfill
    \subfloat[Robot path\label{fig:exp5_robot_path_viki}]{%
       \includegraphics[width=0.25\textwidth]{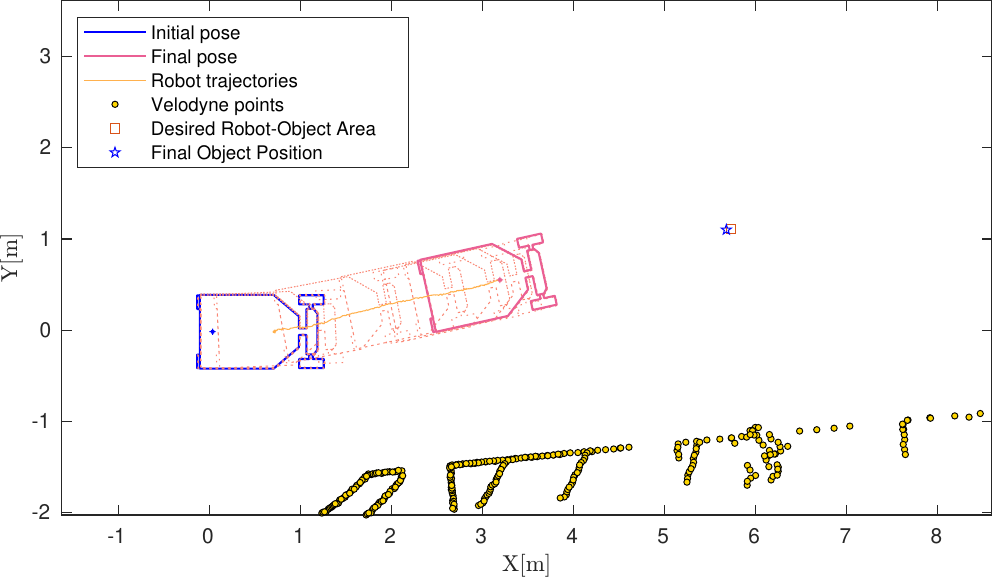}}
    \\
    \subfloat[Lineal velocity and steering angle\label{fig:exp5_velocityrobot}]{%
        \includegraphics[height=3.5cm]{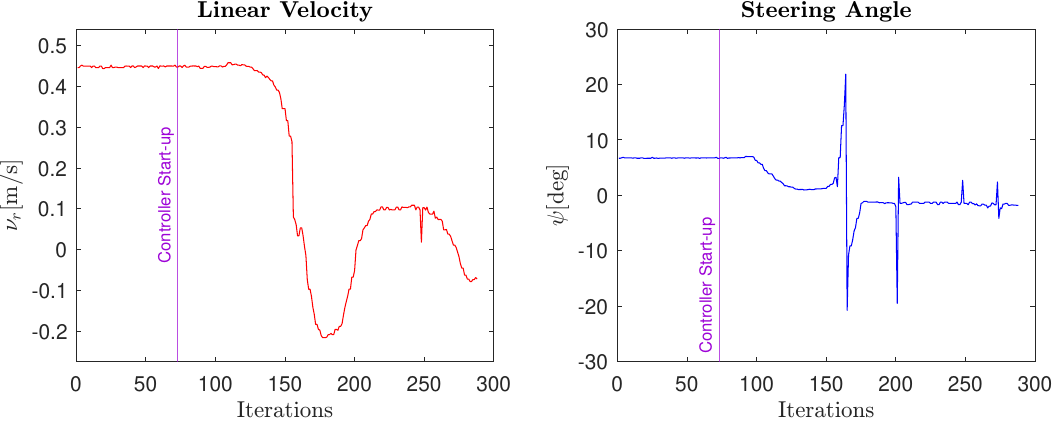}}
    \hfill 
    \subfloat[Lineal velocity and steering angle\label{fig:exp5_velocityrobot_viki}]{%
        \includegraphics[height=3.5cm]{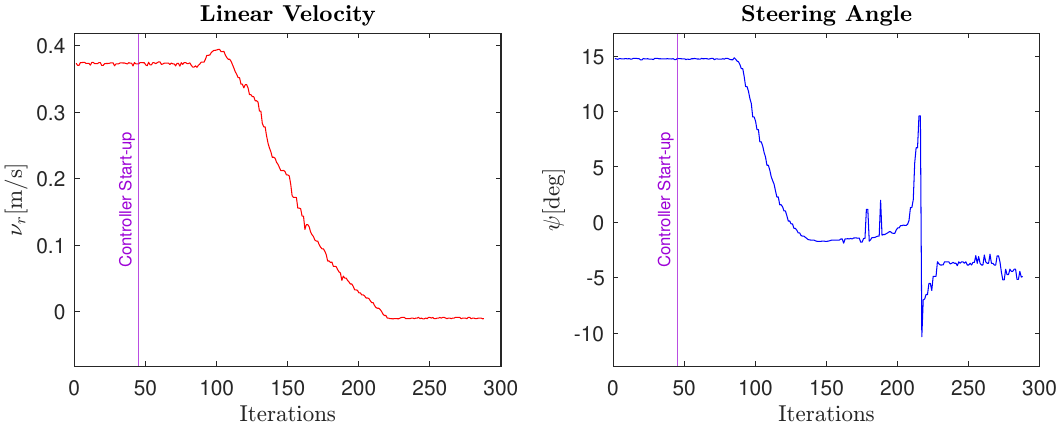}}
    \\
    \subfloat[X and Y robot translation\label{fig:exp5_xyrobot}]{%
        \includegraphics[height=3.5cm]{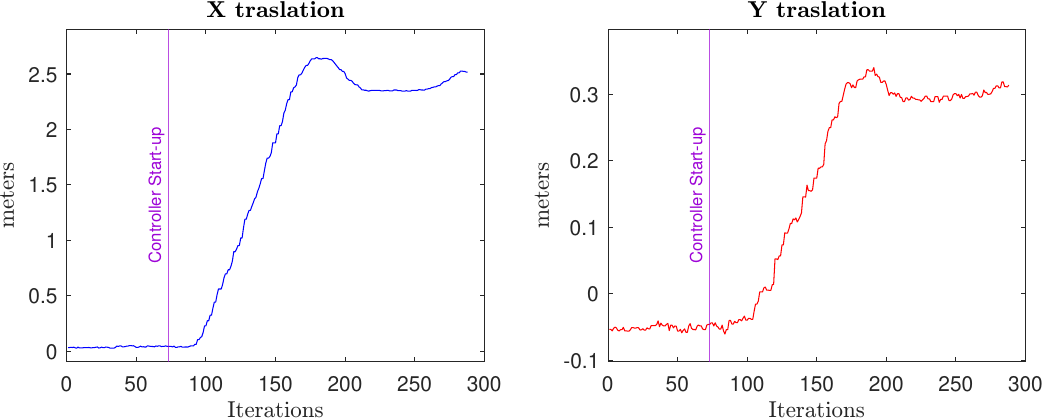}}
    \hfill
    \subfloat[X and Y robot translation\label{fig:exp5_xyrobot_viki}]{%
        \includegraphics[height=3.5cm]{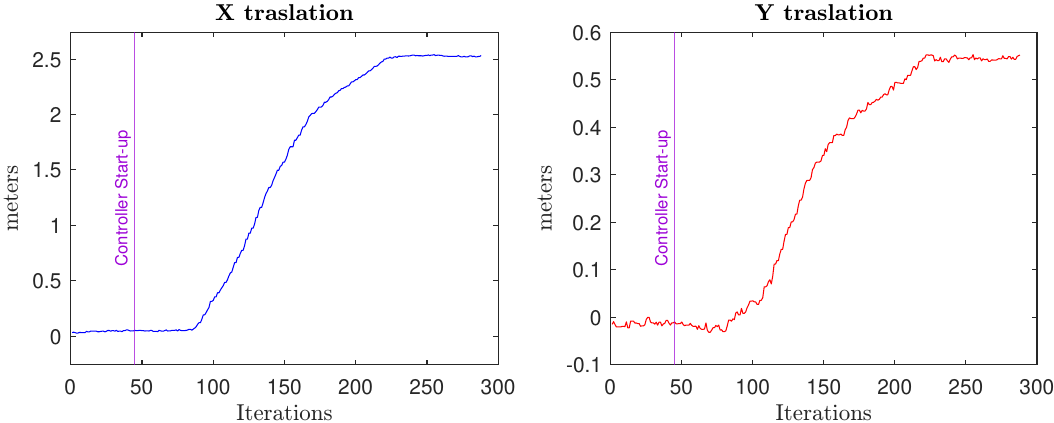}}
        
  \caption{\textbf{(a), (b), (e)} and \textbf{(g)} results of forward positioning of the BLUE robot implemented the MGBM-YOLO method, where the Jacobian's image is calculated with the areas of the current and desired object's bounding box. \textbf{(c), (d), (f)} and \textbf{(h)} results of forward positioning of the BLUE robot implemented the ViKi-HyCo method. In the MGBM-YOLO method, the robot does not arrive at a static position because the actual and desired features in the image plane do not match. On the other hand, with the ViKi-HyCo method, the robot does arrive at a static position, since the actual and desired features eventually coincide.}
  \label{fig:exp5} 
\end{figure*}

Fig. \ref{fig:exp5_camera_path} shows that the visual servoing controller based on the MGBM-YOLO method, fails when navigate to the desired point, because the desired object's bounding box is static and is not constantly updated to the shape of the current object's bounding box. Hence, as shown in Fig. \ref{fig:exp5_velocityrobot}, the calculated linear velocity and steering angle of the controller oscillate as they approach the desired point and do not converge to static value. Then, as shown in Fig. \ref{fig:exp5_robot_path} and \ref{fig:exp5_xyrobot}, the robot does not find a stabilization point. The stabilization point is only reached if the controller velocities converge to zero by making the bounding boxes match in size and shape. 

%% Figure experiment 5

\begin{figure} %% Figure experiment 5 camera_velocities
    \centering
  \subfloat[Camera lineal velocity with MGBM-YOLO\label{fig:camera_lineal_velocity_MGBM}]{%
       \includegraphics[width=0.48\linewidth]{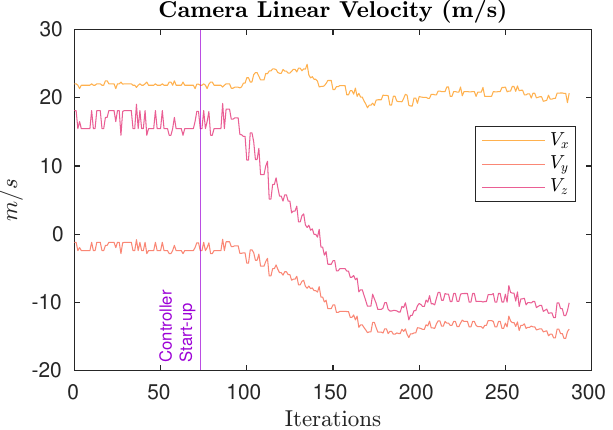}}
    \hfill
   \subfloat[Camera angular velocity with MGBM-YOLO\label{fig:camera_ang_velocity_MGBM}]{%
       \includegraphics[width=0.48\linewidth]{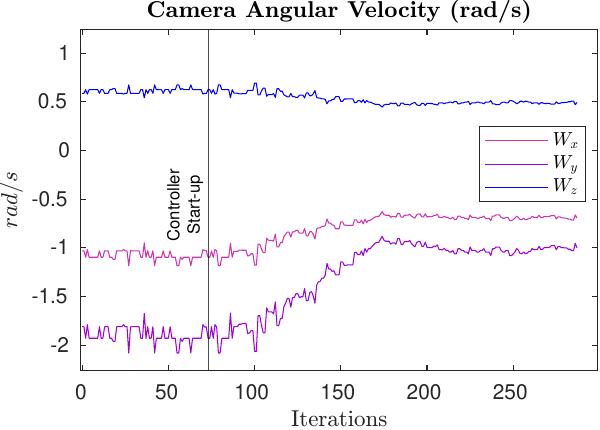}}
    \\
     \subfloat[Camera lineal velocity with ViKi-HyCo\label{fig:camera_lineal_velocity_ViKi}]{%
       \includegraphics[width=0.48\linewidth]{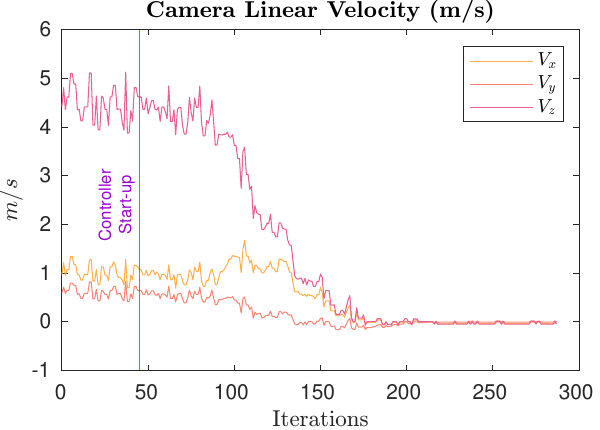}}
    \hfill
   \subfloat[Camera angular velocity with ViKi-HyCo\label{fig:camera_ang_velocity_ViKi}]{%
       \includegraphics[width=0.48\linewidth]{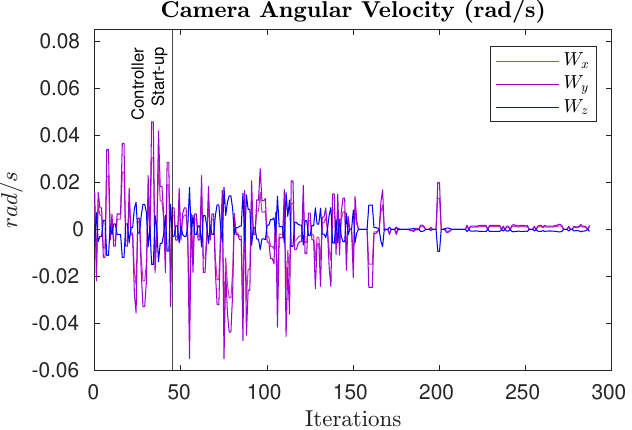}}

  \caption{Camera velocities calculated by visual servoing controllers using the MGBM-YOLO and ViKi-HyCo methods.}
\label{fig:exp5_camera_velocites} 
\end{figure}

We performed the same experiment with the ViKi-HyCo method, as shown in Fig.  \ref{fig:exp5_camera_path_viki}, the desired and detected bounding boxes converge, thus, the robot positioning velocities (see Fig. \ref{fig:exp5_velocityrobot_viki}) have a stabilization point to finish the trajectory, so the robot does not present oscillations to reach the target point, as shown in Fig. \ref{fig:exp5_robot_path_viki} and \ref{fig:exp5_xyrobot_viki}. The visual servoing controller of the MGBM-YOLO method does not converge in the image plane, with the result that the calculated velocities for the camera never reach 0m/s. On the other hand, in the visual servoing controller of ViKi-HyCo, the velocities are stabilized during robot positioning, as the actual and desired features converge on the image plane. Fig. \ref{fig:exp5_camera_velocites} shows the camera velocities provided by the visual servoing controller of the MGBM-YOLO and ViKi-HyCo methods in the comparative experiment.

%\begin{figure} %% Figure experiment 5 camera_velocities
 %   \centering
 % \subfloat[Camera velocity with MGBM-YOLO\label{fig:camera_velocity_MGBM}]{%
 %      \includegraphics[width=0.48\linewidth]{figures/velas51.pdf}}
 %   \hfill
 %  \subfloat[Camera velocity with ViKi-HyCo\label{fig:camera_velocity_ViKi}]{%
  %     \includegraphics[width=0.48\linewidth]{figures/velas52.pdf}}
  
  %\caption{Camera velocities calculated by visual servoing controllers using the MGBM-YOLO and ViKi-HyCo methods.}
%\label{fig:exp5_camera_velocites} 
%\end{figure}

\subsection{Robot Placement Experiments}
\label{sec:exp4}
In this section, we evaluate the results of the complete placement task of the BLUE robot, which consists of implementing experiments in \ref{sec:exp2} in a combined positioning process. In this way, we analyze the complete placement task of the robot to reverse towards a target object. We consider this as the manipulation zone, because the robot must position itself in this way to be able to manipulate objects with the onboard robotic arm. To analyze the results of these experiments, we performed a total of 12 experiments with different objects. It is important to mention that the experiment is initialized when there is at least one detection of the object by the YOLOv5 NN with the front camera image. In this way, the object is located with respect to the robot. In these experiments, the full ViKi-HyCo method implemented is divided into 3 states, which are the following: 

\textbf{State 1. ViKi-HyCo forward positioning} The method used in experimentation \ref{sec:exp2}. (see Fig.\ref{fig:exp4}). 

\textbf{State 2. Kinematic control} Since no camera focuses on the desired object while the robot rotates, a kinematic controller is used for robot rotation. This is used after positioning the object at the desired point in front of the robot with the ViKi-HyCo forward positioning. 

\textbf{State 3. ViKi-HyCo backward positioning} The method used in experimentation \ref{sec:exp2}. (see Fig.\ref{fig:exp3}).

The robot carries out the transition from one state to another of the experiment automatically. To do this, the transition are validated with the value of the features error of the visual servoing controller or the position error of the kinematic control. In the case of ViKi-HyCo controllers for forward and backward positioning, the $\mathbf{\dot{f}}$ feature error is evaluated, which must be in range $-2<\mathbf{\dot{f}}< 2$ pixels to establish that the positioning has been successful. For kinematic control, the change of state is given with a positioning error within the range $-0.01<\mathbf{\dot{h}}<0.01$ m. Finally, our method prioritizes the detections with the front camera, in this way, the process begins with the front camera detection of the object.

\begin{figure}%% Figure experiment 6
    \centering
  \subfloat[Robot path positioning\label{fig:exp6_robot_path}]{%
       \includegraphics[width=0.75\linewidth ]{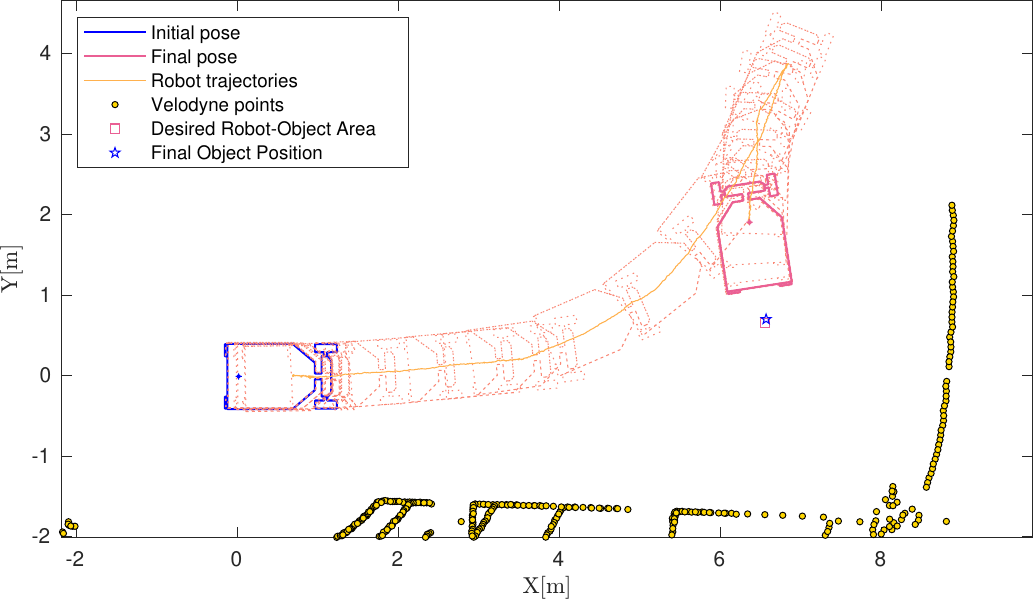}}
    \\
    \subfloat[Camera velocity \label{fig:exp6_velocitycamera}]{%
    \includegraphics[width=0.75\linewidth]{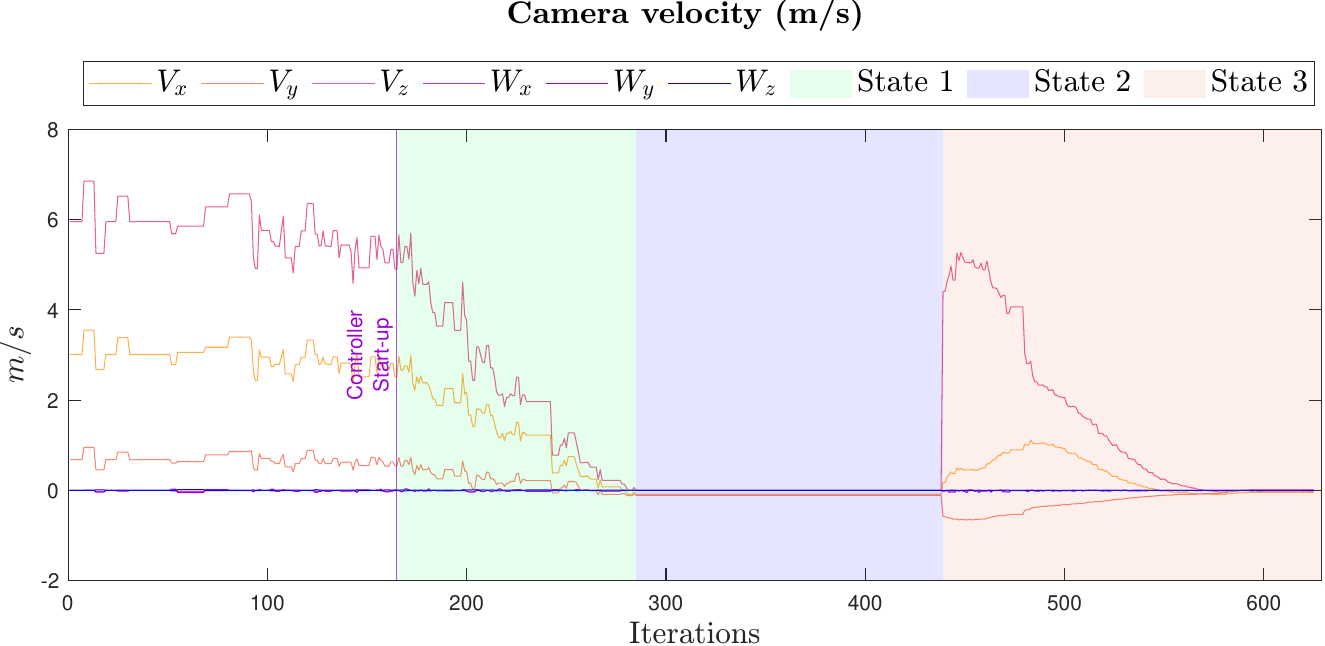}}
     \\   
  \subfloat[Lineal velocity\label{fig:exp6_velocityrobot}]{%
        \includegraphics[height=3.5cm]{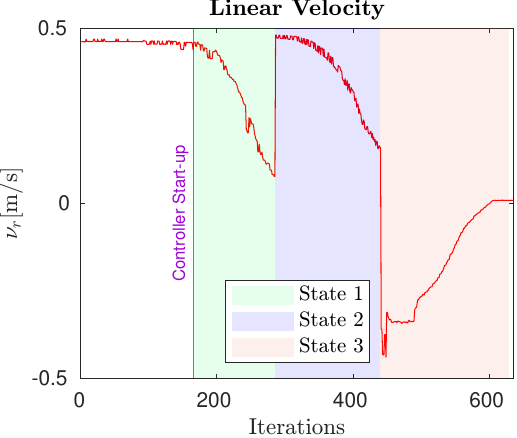}}
    \hfill
   \subfloat[Steering angle\label{fig:exp6_steeringrobot}]{%
        \includegraphics[height=3.5cm]{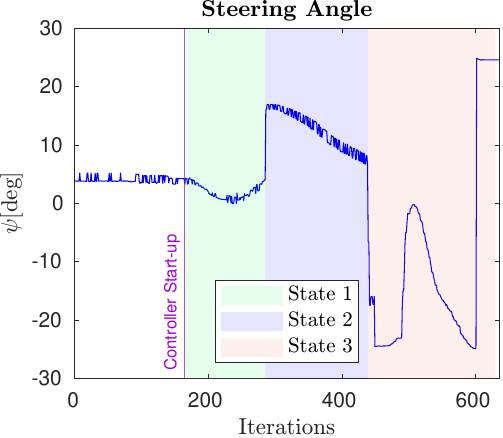}}
 \\
  \subfloat[X robot translation\label{fig:exp6_x_translation}]{%
        \includegraphics[height=3.5cm]{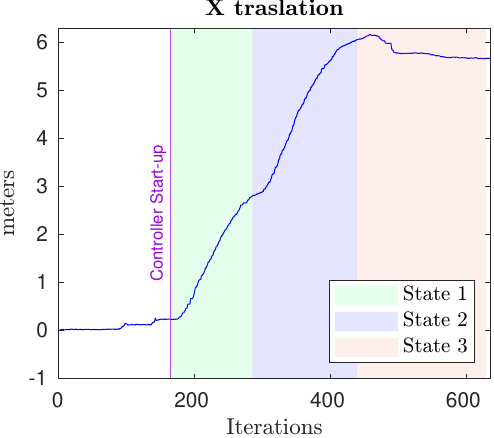}}
    \hfill
  \subfloat[Y robot translation\label{fig:exp6_y_translation}]{%
        \includegraphics[height=3.5cm]{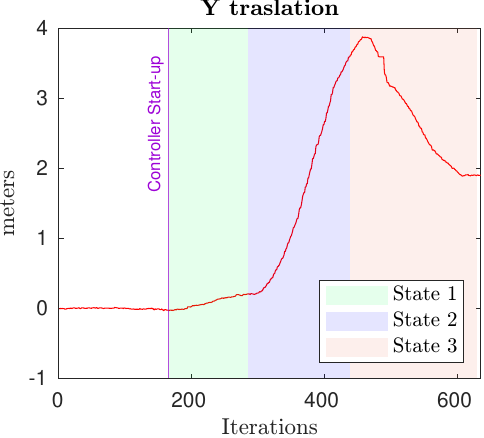}}
  \caption{Results of the complete positioning experiment using the ViKi-HyCo method. The colored zones represent the state of the experiment. State 1 (green) forward positioning. State 2 (blue) Robot rotation with kinematic controller. State 3 (red) backward positioning.}
  \label{fig:exp6} 
\end{figure}

Fig. \ref{fig:exp6} shows the results of the complete positioning experiment using the ViKi-HyCo controller. The path in Fig. \ref{fig:exp6_robot_path} shows that the robot is positioned in reverse towards the object, locating it in the desired area of the experiment. The velocities of the front and rear camera are shown in Fig.\ref{fig:exp6_velocitycamera}. In Step 1, the velocities of the front camera are calculated with the ViKi-HyCo controller. In Step 2, the velocities of both cameras are 0 m/s, since only the kinematic controller is used and the object is outside the camera FOV. Finally, in Step 3, the velocities shown correspond to the rear camera, calculated by the ViKi-HyCo controller.

\begin{figure}
    \centering
    \subfloat[Robot initial position\label{fig:exp6_initial_pos_robot}]{%
        \includegraphics[width=0.47\linewidth]{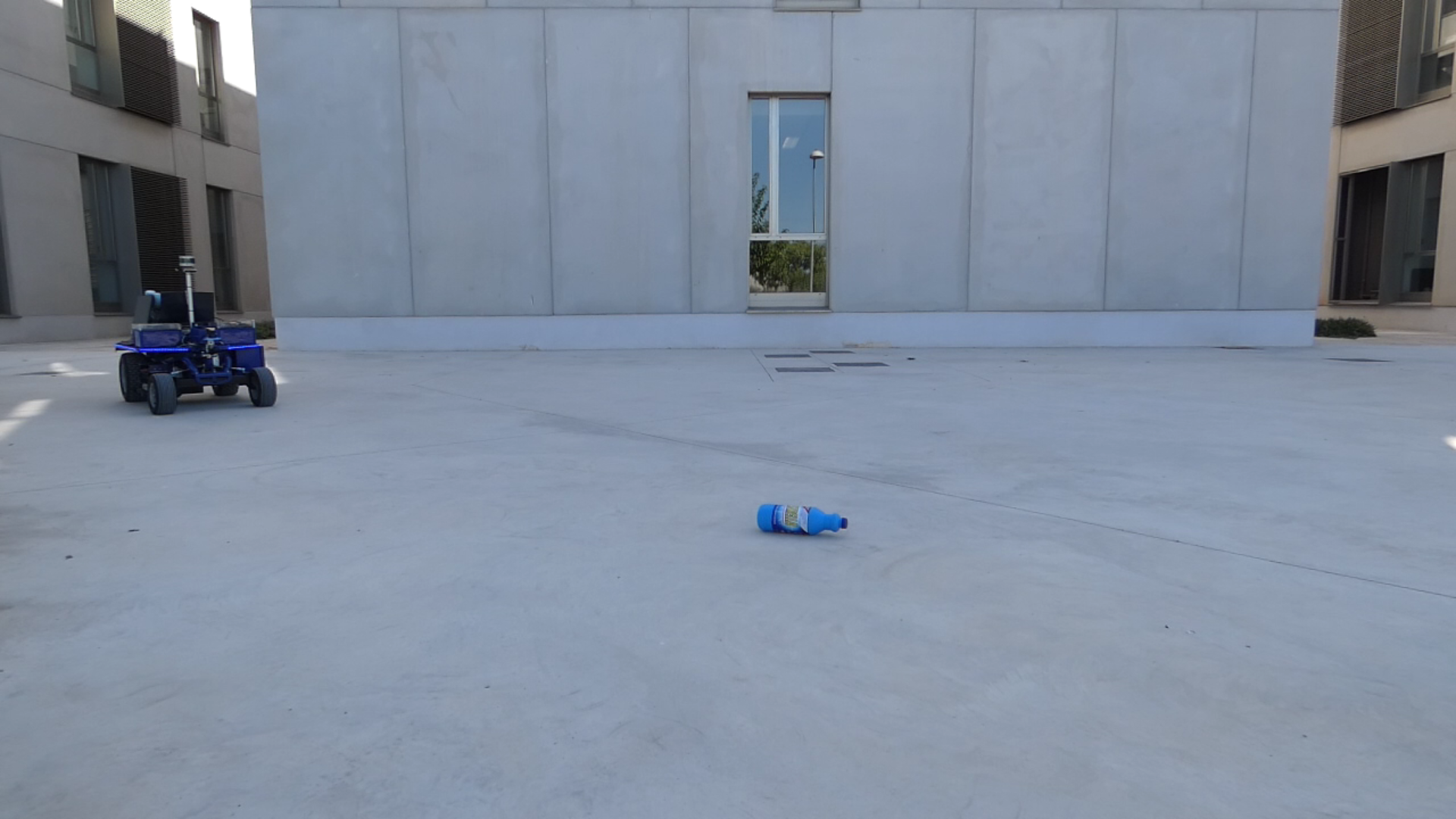}}
    \hfill
    \subfloat[Robot final position\label{fig:exp6_final_pos_robot}]{%
        \includegraphics[width=0.47\linewidth]{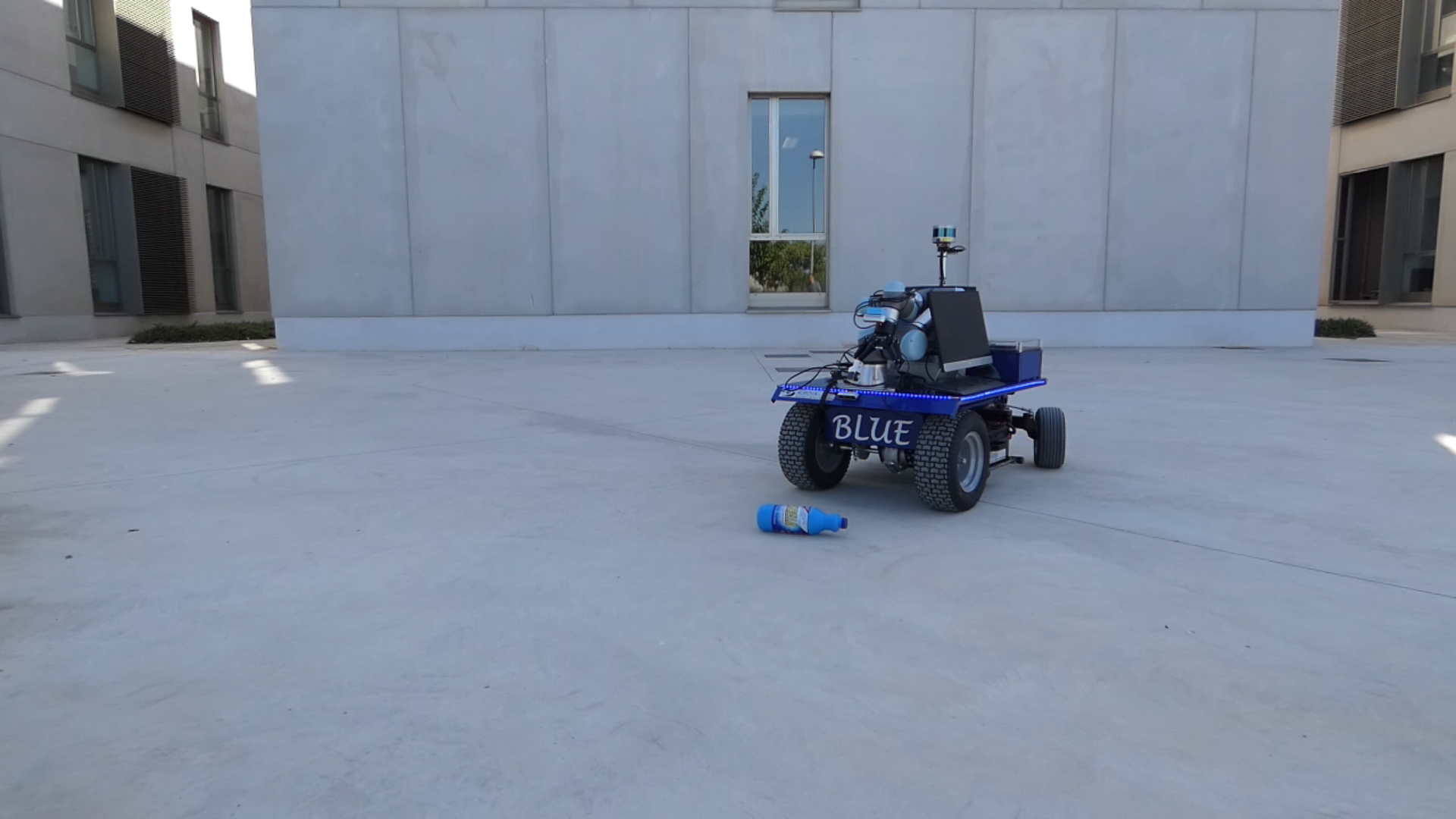}}
    \\
    \subfloat[Initial front camera image\label{fig:exp6_initial_camera}]{%
        \includegraphics[width=0.47\linewidth]{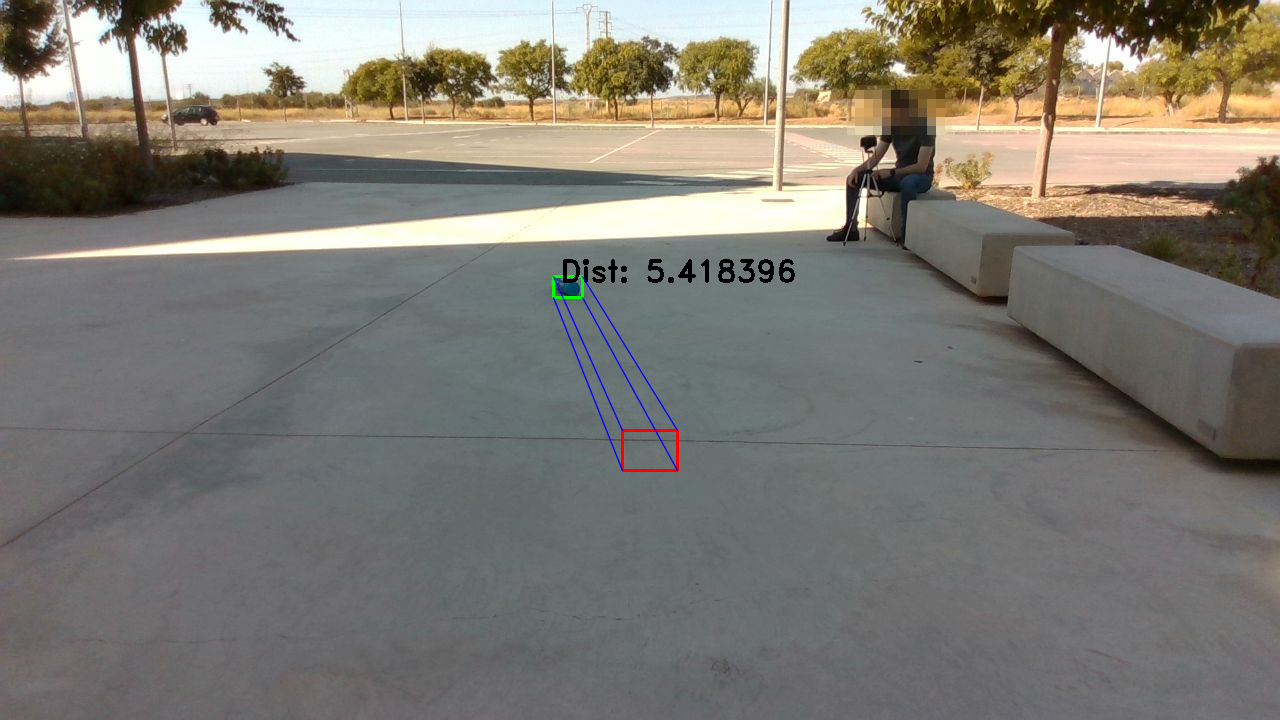}}
    \hfill
    \subfloat[Final rear camera image\label{fig:exp6_final_camera}]{%
        \includegraphics[width=0.47\linewidth]{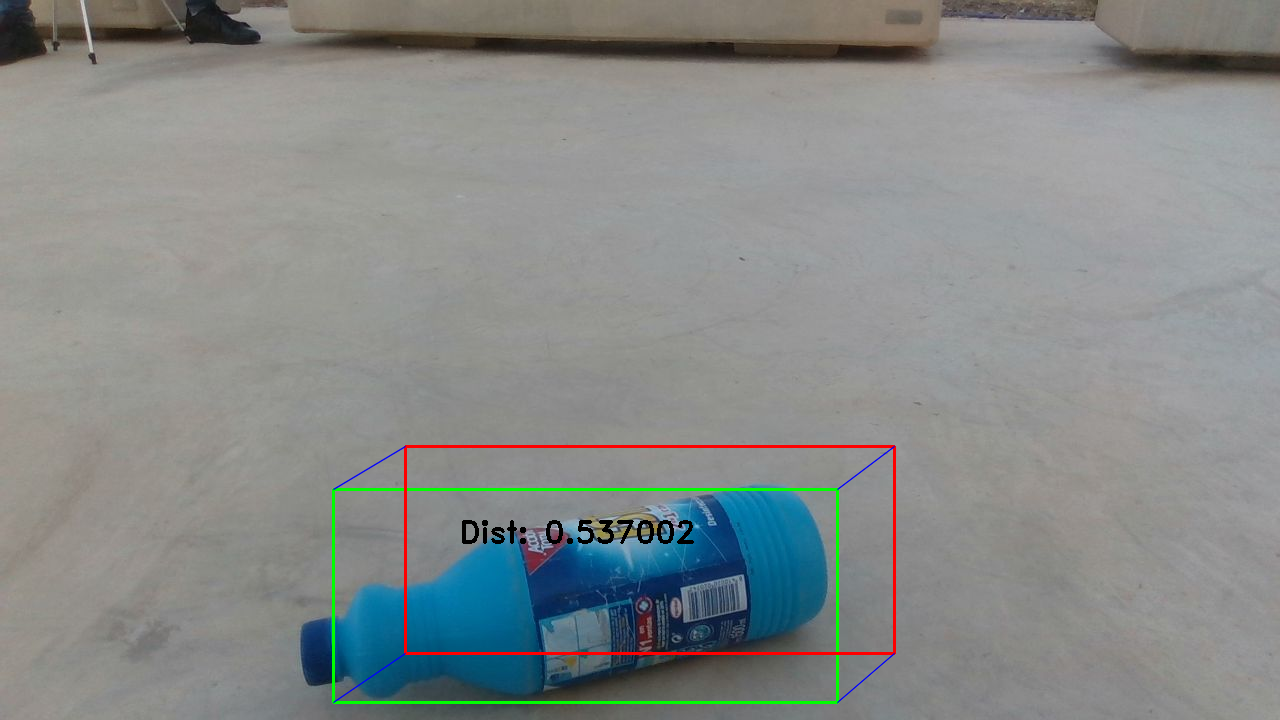}}
  \caption{Initial and final position of the robot in the complete positioning experiment and images from the front and rear cameras.}
  \label{fig:exp6_images} 
\end{figure}

Fig. \ref{fig:exp6_x_translation}, \ref{fig:exp6_y_translation}, \ref{fig:exp6_velocityrobot} and \ref{fig:exp6_steeringrobot} shows respectively the X translation, Y translations, the linear velocity and the steering angle in each of the three states of the robot's trajectory from the initial point to the reverse positioning towards the desired object. Finally, Fig. \ref{fig:exp6_initial_pos_robot} and \ref{fig:exp6_final_pos_robot} show the initial and final pose of the robot's trajectory, and the Fig. \ref{fig:exp6_initial_camera} with Fig. \ref{fig:exp6_final_camera} show the initial and final frame of frontal and rear camera respectively.

\subsection{Results}
\begin{figure}
    \centering
    \subfloat[Results of backward positioning\label{fig:results_backward}]{%
        \includegraphics[width=0.47\linewidth]{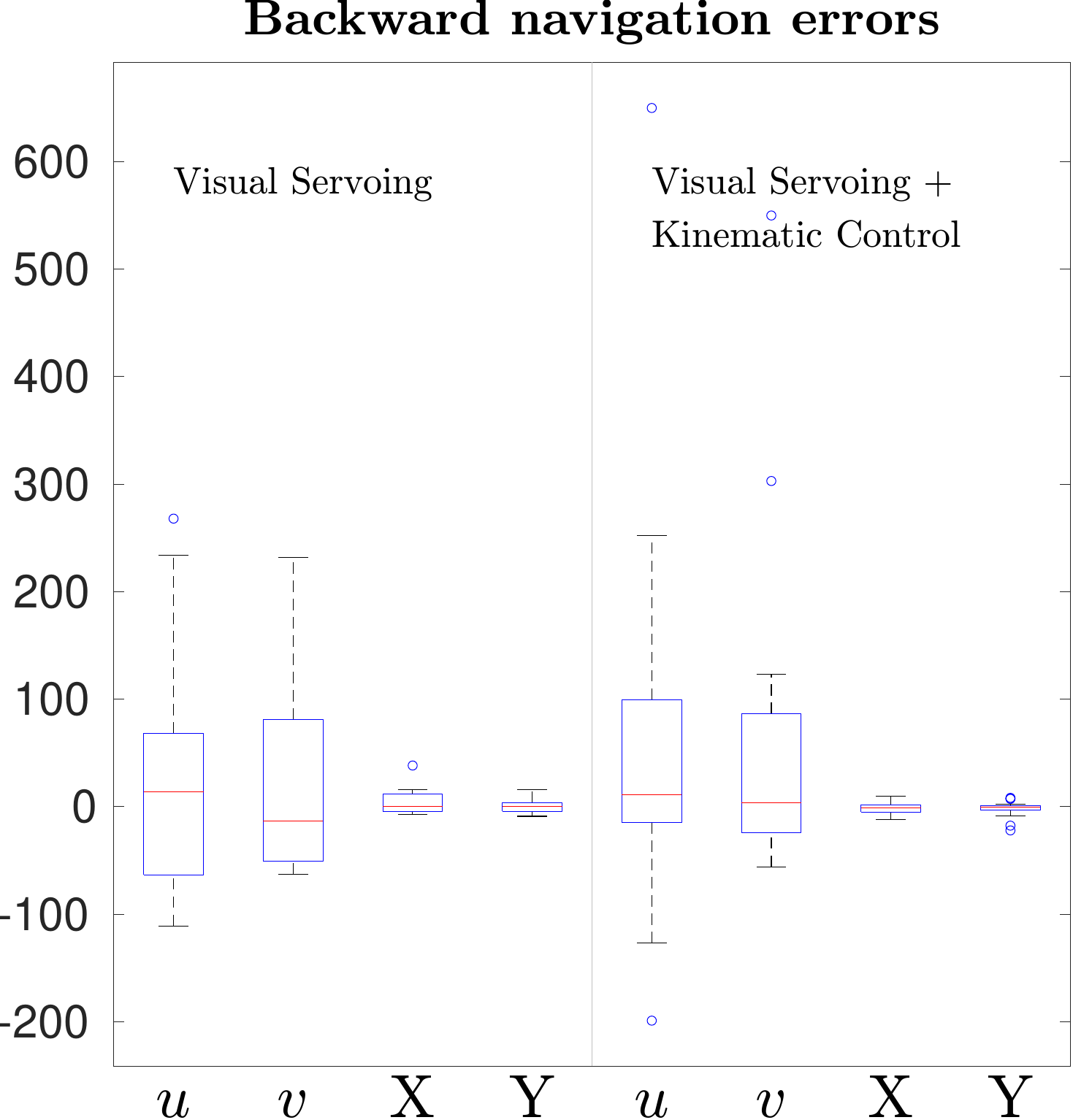}}
    \hfill
    \centering
    \subfloat[Results of forward positioning\label{fig:results_forward}]{%
       \includegraphics[width=0.47\linewidth]{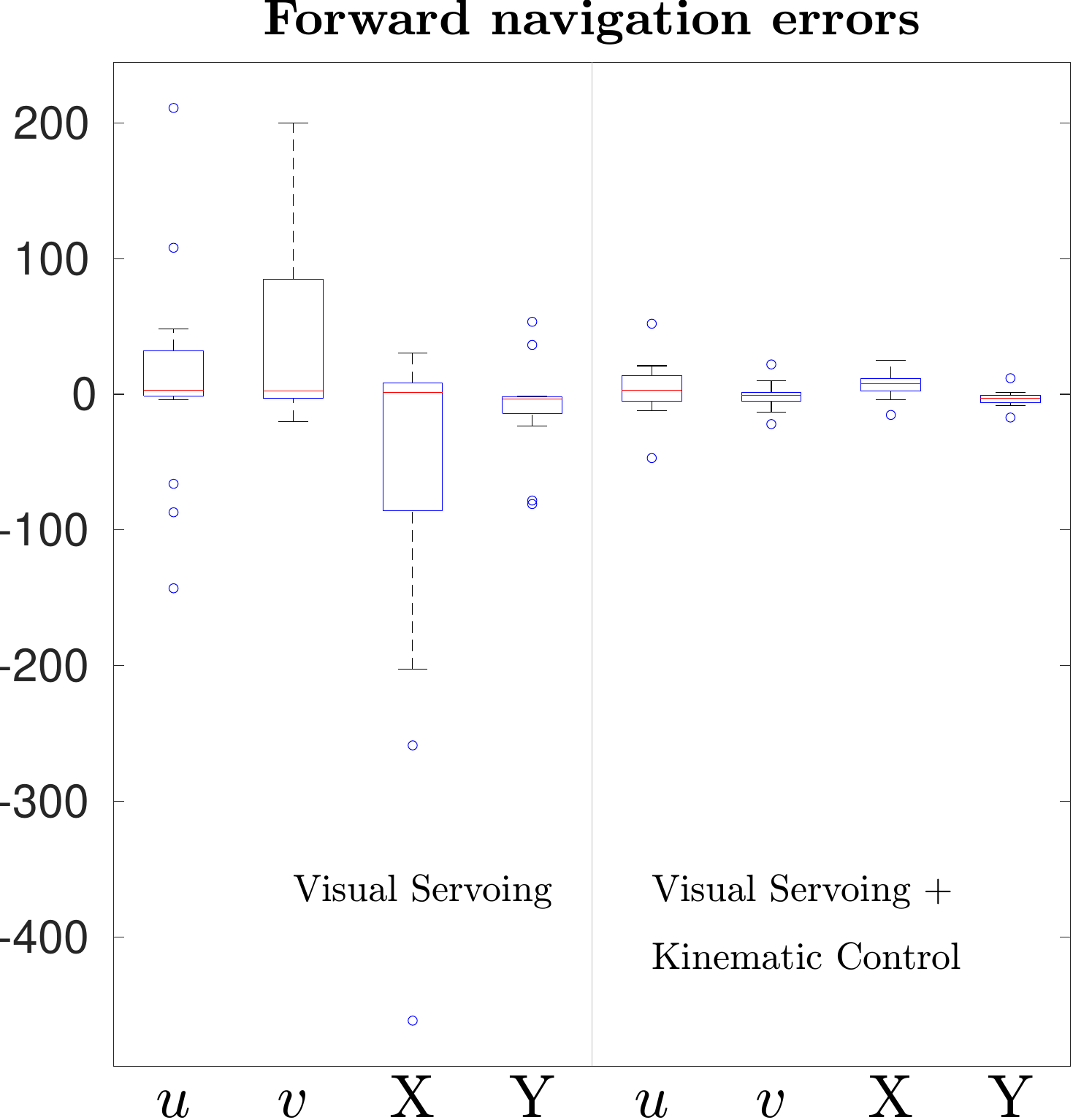}}
    \hfill
  \caption{Comparative results of forward and backward positioning experiments using the classic visual servoing controller versus the ViKi-HyCo method. The plane image errors $u$ and $v$ are in pixels and positioning errors $\text{X}$ and $\text{Y}$ are in centimeters. In each box, the red center line indicates the median, and the lower and upper ends of the box indicate the 25th and 75th percentiles, respectively. The dashed line extends to the most extreme data points, and outliers are individually represented by the blue symbol ''o''.} 
  %\label{fig:final_results_boxplot} 
\end{figure}

In the following section, we show the statistical results of 94 experiments of sections \ref{sec:exp1}, \ref{sec:exp2},\ref{sec:exp3} and \ref{sec:exp4}. Fig.\ref{fig:results_backward}  shows the reverse positioning errors when the robot arrives at a desired position. It can be seen that, when using a kinematic controller combined with a visual servoing controller (ViKi-HyCo), the translations on the x-axis show a minor improvement as opposed to when only a visual servoing controller is used. This small difference is due to the fact that in this positioning some object detection frames were lost, since the camera focuses on the ground, having the object in direct line of sight and at close range. Our backward positioning results have an error of $0.0375\pm0.0334$ m for the X axis and $0.0452\pm0.0601$ m for the Y axis.

For the case of the forward positioning experiments with the visual servoing and ViKi-HyCo controllers, Fig. \ref{fig:results_forward} shows that our method has a considerable decrease of the error in the X-axis, this is due to the fact that when the visual servoing controller does not have feature points, obtained from the object detections, the kinematic controller is used. Additionally, switching from the visual servoing controller to a kinematic controller helps to recover object detections with YOLOv5. Hence, the robot stops are avoided when it performs position maneuvers. In this way, the $u$,$v$ pixel errors also decrease. The positioning errors in the forward positioning of our method are $0.0911\pm0.0631$ m in the X-axis and $0.0457\pm0.0437$ m in the Y-axis.

When comparing ViKi-HyCo against MBGB-YOLO, the results in section \ref{sec:exp3} show that our method does not depend on the physical characteristics of the target object of the controller, since it is not necessary to know in advance the dimensions of the desired bounding box. Our system constantly updates the characteristics of the bounding box depending on the current distance to the object and the position in the image plane where we want to place the object. Although MBGB-YOLO positioning errors approach zero, they do not converge because the calculated camera velocities do not decrease due to errors in the image plane. ViKi-HyCo converges to the desired point by constantly updating the desired bounding box.

Finally, by using both positioning experiments in a combined task, the robot positioning error with reference to the target point resulting in an error of $0.0428\pm0.0467$ m in the X-axis and $0.0515\pm0.0313$ m in the Y-axis at the end of the task. In this way, we demonstrate that the proposed method generates the velocities required for the BLUE robot's positioning, in order to position it towards an object that is in the ground plane. These positioning errors are adequate for our experimentation since the task has successfully positioned the BLUE robot with respect to the object in the manipulation zone. 

%\begin{figure}[h]
%\centerline{\includegraphics[width=0.9\linewidth]{figures/Results/ViKi_error_complete_trajectory%.pdf}}
%\caption{} \label{fig:results_combined_task}
%\end{figure}

\subsection{ViKi-HyCo Run-time}
To validate the execution time of each iteration of the complete process, the times of each of the sub-processes has been measured, and they are indicated in Table \ref{tab:run-time}. The total time of Object detection, Interpolated point cloud, LiDAR-Camera fusion, Visual Servoing and Kinematic controller, result in a time of less than 45 ms. This  is guaranteed for real-time in our application, because the camera works at 15 fps and the odometry has a frequency of 20 Hz. It is worth mentioning that the Visual Servoing and Kinematic Controller execution time of 2 ms is the control law calculation time of equation \eqref{eq:law_control_ViKi-HyCo_global}, together with equation \eqref{eq:law_control_filter}, which prevents abrupt velocity changes in the controller output. In addition, the transition between controllers is despicable, since it is incorporated in the same control law.  

\begin{table} 
\caption{Total run-time of the process in each iteration }
%\extrarowheight = 0.5ex
\renewcommand{\arraystretch}{0.8}
\centering
\begin{tabular}{ll} 
\hline
\textbf{Process~}                                                                           & \begin{tabular}[c]{@{}l@{}}\textbf{Average }\\\textbf{time (ms)}\end{tabular}  \\ 
\hline
\\
Object detection (YOLOv5):                                                                  & 12                                                                             \\ \\
\begin{tabular}[c]{@{}l@{}}Interpolated Point cloud \\and LiDAR-Camera fusion:\end{tabular} & 30                                                                             \\ \\
\begin{tabular}[c]{@{}l@{}}Visual Servoing and \\Kinematic controller:~\end{tabular}         & 2                                                                              \\ \\
  \textbf{Total}    & 44                                                                             \\ 
\hline
\end{tabular}
\label{tab:run-time}

\end{table}

\section{Conclusion}
\label{sec:cloncusion}
In this paper, we present a hybrid-control system combining a visual servoing and a kinematic controller for the positioning maneuvers of a non-holonomic robot. As features of the visual servoing controller, we use the detections of a YOLOv5 NN, which detects and generates the current and desired bounding box of domestic waste in outdoor environments. We demonstrate that the union of both controllers is possible by performing several experiments of forward and backward positioning and a combined task in outdoor environments with depth images that can come from RGB-D cameras or from point clouds generated by a LiDAR sensor. Furthermore, our approach demonstrates that it does not need a visual tracker algorithm for object tracking when the camera is in motion or when the image input source is switched, this is achieved with the kinematic controller that locates the detected object in a desired position. Our method achieves a run-time of less than 45 ms, which we consider real-time for our robotic applications. In addition, the method has a small positioning error considered for the applications for which the BLUE robot was designed. ViKi-HyCo allows positioning control for any type of robot as long as the kinematic model of the robot is known and the appropriate sensing sensors are available. 

In future projects, we will use a cost mapping system to generate a safe route to the object if there is an obstacle in the scenario that prevents proper navigation \cite{Mobile_Robot_Mapping}. In addition, having solved the positioning to domestic waste in our BLUE robot, we will design a method for handling this domestic waste in outdoor environments with a robotic arm on board the BLUE robot. 

\bibliographystyle{IEEEtran}
\bibliography{references.bib}
\begin{IEEEbiography}[{\includegraphics[width=1in,height=1.25in,clip,keepaspectratio]{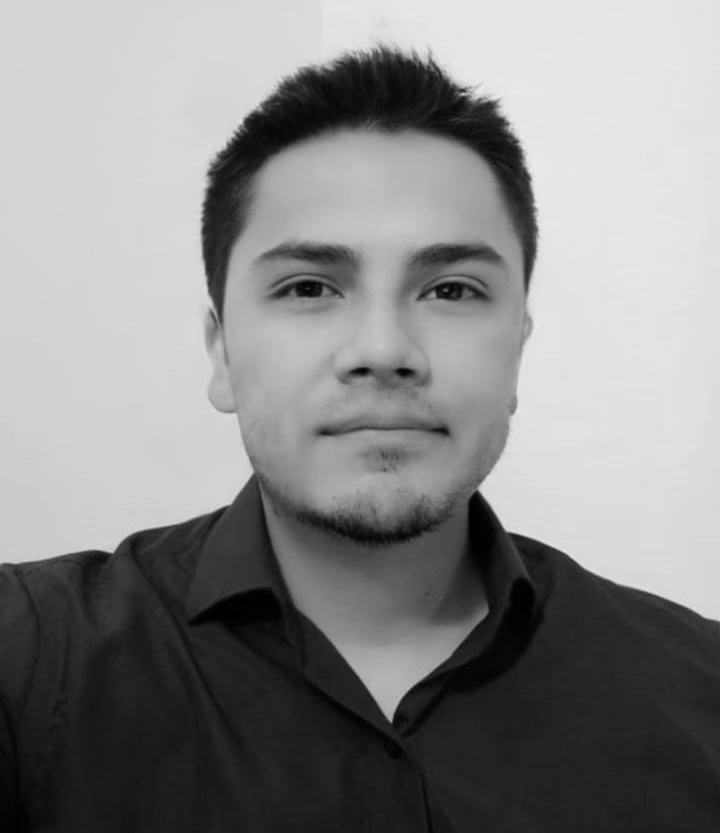}}]{Edison P. Velasco Sánchez} received the degree in Electronic Engineering and Instrumentation from the University of the Armed Forces ESPE (Ecuador) in 2015 and a Master's Degree in Automation and Robotics from the University of Alicante (Spain) in 2018. He was a research technician in the ARSI research group at the ESPE University and in the Automation, Robotics and Computer Vision Group (AUROVA) of the University of Alicante. He is currently pursuing a Ph.D. degree in AUROVA at the University of Alicante funding by the Regional Valencian Community Government and the Ministry of Science, Innovation and Universities  through the grant PRE2019-088069. His research interests include navigation and autonomous localization in UGVs with cameras and LIDAR sensors.
\end{IEEEbiography}
\begin{IEEEbiography}[{\includegraphics[width=1in,height=1.25in,clip,keepaspectratio]{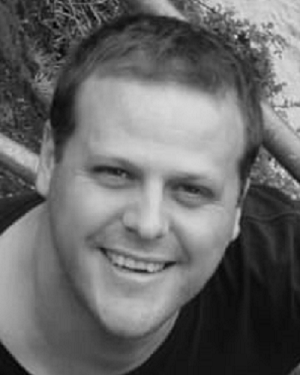}}]{Miguel Ángel Muñoz-Bañón} received his B.S. degree in telecommunications engineering from the University of Alicante in 2016, the M.S. degree in artificial intelligence with UNED in 2017, and his PhD from the University of Alicante in the robotics research line in 2022. He completed his dissertation as part of the AUROVA research group. In 2021, he visited, as a researcher, the Karlsruhe Institute of Technology (KIT) in Germany. He received the award of the best dissertation in robotics for 2022 in Spain, granted by the Spanish Automatics Committee (CEA). His research interests include geo-localization for autonomous robots and environment reconstruction by neural radiance fields models.
\end{IEEEbiography}
\begin{IEEEbiography}[{\includegraphics[width=1in,height=1.25in,clip,keepaspectratio]{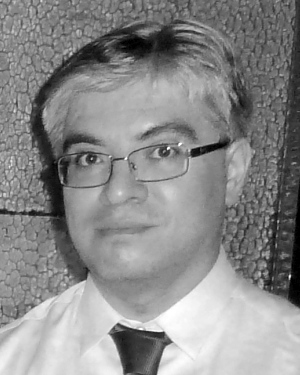}}]{Francisco A. Candelas} received the Computer Science Engineer and the Ph.D. degrees in the University of Alicante (Spain), in 1996 and 2001 respectively. He is Associate Professor in the University of Alicante since 2003, where he teaches currently courses about Automation and Robotics Sensors in the Degree in Robotic Engineering. Previously, he was in tenure track from 1999 to 2003. Dr. Candelas also researches in the Automation, Robotics and Computer Vision Group (AUROVA) of the University of Alicante since 1998, and he has involved in several research projects and networks supported by the Spanish Government, as well as development projects in collaboration with regional industry. His main research topics are autonomous robots, robot development, and virtual/remote laboratories for teaching.
\end{IEEEbiography}
\begin{IEEEbiography}[{\includegraphics[width=1in,height=1.25in,clip,keepaspectratio]{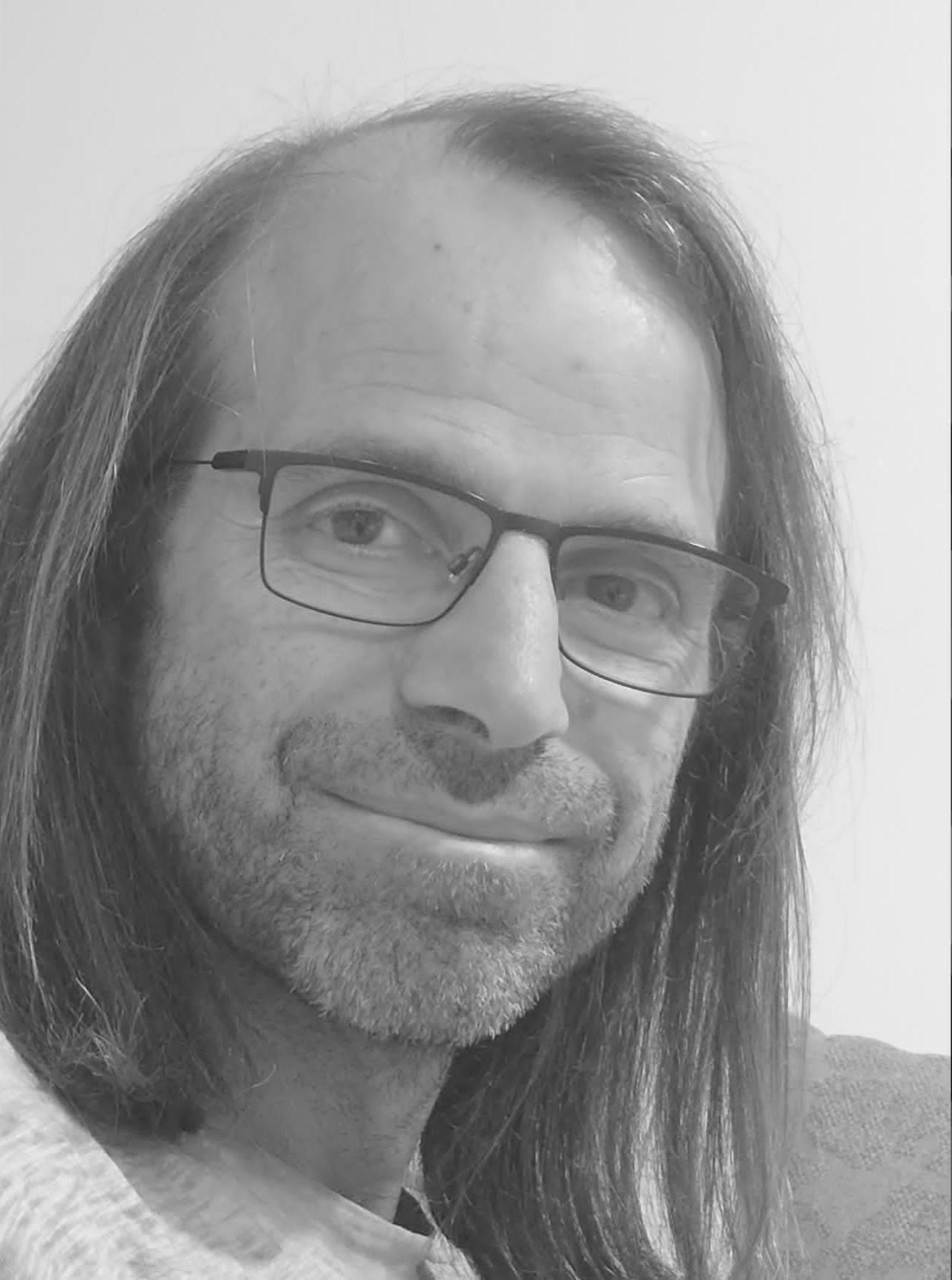}}]{Santiago Puente} received the Computer Science Engineer and the PhD degrees in the University of Alicante (Spain), in 1998 and 2003 respectively. He is a full-time lecture and researcher at the University of Alicante since 2003, where he teaches currently in the Degree in Robotic Engineering. From 2013 to 2019, Dr. Puente has been deputy director of Infrastructures and facilities Polytechnic School of the University of Alicante, Furthermore, from 2019 to 2021 he has been Academic Coordinator  of BEng Robotics Engineering. Dr. Puente also researches in the Automation, Robotics and Computer Vision Group (AUROVA) of the University of Alicante since 1999, and he has involved in several research projects and networks supported by the Spanish Government, as well as development projects in collaboration with regional industry. His research interests include automation and robotics (intelligent robotic manipulation, robot perception systems, robot imitation learning, field mobile robots), and e-learning.
\end{IEEEbiography}
\begin{IEEEbiography}[{\includegraphics[width=1in,height=1.25in,clip,keepaspectratio]{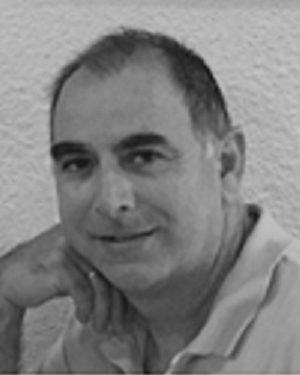}}]{Fernando Torres} was born in Granada, where he attended primary and high school. He moved to Madrid to undertake a degree in Industrial Engineering at the Polytechnic University of Madrid, where he also carried out his PhD thesis. The last year of his PhD thesis he became a full-time lecturer and researcher at the University of Alicante, and he has worked there ever since. He directs the research group ``Automatics, Robotics and Computer Vision'' founded in 1996 at the University of Alicante. He is a member of TC 5.1 and TC 9.4 of the IFAC, a Senior Member of the IEEE and a member of CEA. Since july 2018 he is coordinator of the area of Electrical, Electronic and Automatic (IEA) of the Spanish Agency of Statal Research (AEI). His research interests include automation and robotics (intelligent robotic manipulation, visual control of robots, robot perception systems, field mobile robots, advanced automation for industry 4.0, artificial vision engineering), and e-learning. Currently, his research focuses on automation, robotics, and e-learning. In these lines, it currently has more than fifty publications in JCR-ISI journals and more than a hundred papers in international congresses. He was Leader Research in several research projects and he has supervised several PhD in these lines of research.
\end{IEEEbiography}

\EOD

\end{document}